\documentclass[runningheads]{llncs}

\usepackage{eccv}

\usepackage{eccvabbrv}

\usepackage{graphicx}
\usepackage{booktabs}
\definecolor{gray}{gray}{0.9}
\definecolor{tgray}{gray}{0.5}
 
\usepackage{colortbl}

\usepackage{xcolor} 
\usepackage{wrapfig}
\usepackage{algorithm}
\usepackage{algpseudocode}

\definecolor{color1}{RGB}{102,194,165}
\definecolor{color2}{RGB}{252,141,98}
\definecolor{color3}{RGB}{141,160,203}
\definecolor{color4}{RGB}{231,138,195}
\definecolor{color5}{RGB}{166,216,84}

\usepackage[accsupp]{axessibility}  

\usepackage{pifont}
\newcommand{\cmark}{\ding{51}}%
\newcommand{\xmark}{\ding{55}}%

\usepackage{hyperref}

\usepackage{orcidlink}

\begin{document}

\title{Object-Conditioned Energy-Based  Attention Map Alignment in Text-to-Image Diffusion Models} 

\titlerunning{Object-Conditioned Energy-Based  Attention Map Alignment}

\author{Yasi Zhang\and
Peiyu Yu \and
Ying Nian Wu }

\authorrunning{Y. Zhang et al.}

\institute{Department of Statistics and Data Science \\ University of California, Los Angeles \\
\email{yasminzhang@ucla.edu, yupeiyu98@g.ucla.edu, ywu@stat.ucla.edu} }

\maketitle

\begin{abstract}
 Text-to-image diffusion models have shown great success in generating high-quality text-guided images. Yet, these models may still fail to semantically align generated images with the provided text prompts, leading to problems like incorrect attribute binding and/or catastrophic object neglect. Given the pervasive object-oriented structure underlying text prompts, we introduce a novel object-conditioned Energy-Based Attention Map Alignment (EBAMA) method to  address the aforementioned problems. We show that an object-centric attribute binding loss naturally emerges by approximately maximizing the log-likelihood of a $z$-parameterized energy-based model with the help of the negative sampling technique. We further propose an object-centric intensity regularizer to prevent excessive shifts of objects attention towards their attributes. Extensive qualitative and quantitative experiments, including human evaluation, on several challenging benchmarks demonstrate the superior performance of our method over previous strong counterparts. With better aligned attention maps, our approach shows great promise in further enhancing the text-controlled image editing ability of diffusion models. Code is available at \url{https://github.com/YasminZhang/EBAMA}.
\end{abstract}

\section{Introduction}
\label{sec:intro}

Recently, large-scale  text-to-image diffusion models  \cite{Rombach_2022_CVPR, nichol2021glide, ramesh2022hierarchical, saharia2022photorealistic, balaji2022ediffi, hoogeboom2023simple}  have showcased remarkable capabilities in producing diverse, imaginative, high-resolution visual content based on free-form text prompts.  
Despite their revolutionary progress, however, these models may not consistently capture and convey the full semantic meaning of the provided text prompts \cite{conwell2022testing, rassin2022dalle}. 
Some well-known issues include   omission, hallucination, or duplication of details \cite{yu2022scaling}, semantic leakage of attributes between entities \cite{rassin2022dalle}, and miscomprehension of intricate textual descriptions \cite{saharia2022photorealistic}.\footnotetext{Accepted to European Conference on Computer Vision (ECCV) 2024}



\begin{figure}[t]
    \centering
\includegraphics[width=\textwidth]{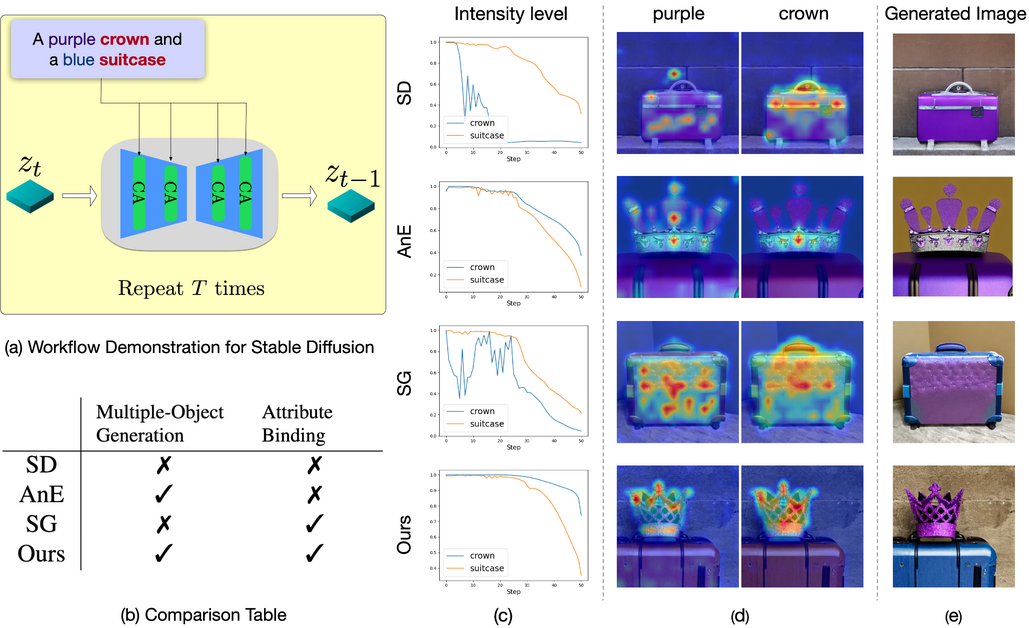}
     \caption{\textbf{Key observations of the generation process of diffusion models}. The given prompt is ``a \textcolor{Orchid}{purple} \underline{crown} and a  \textcolor{blue}{blue} \underline{suitcase}''. In panel (c), we hypothesize that if the intensity level of any object in the prompt does not remain high during the first half of the denoising process, e.g. the crown in SD and SG,  the model would fail to generate the object in the final image. The panel (d) suggests that if the attention map distributions of any attribute-object pair are not aligned, the model would struggle to correctly bind attributes to their respective objects, e.g. `purple' and `crown' in SD and AnE. The generated images are  displayed in the panel (e). All methods share the same random seed.  } \label{key}
\end{figure}

Many previous works have focused on addressing the semantic misalignment issues, particularly concerning multiple-object generation and attribute binding. Composable Diffusion (CD) \cite{liu2022compositional} composes multiple output noises guided by different objects in a text prompt during the generation process. However, this approach often results in a blend of objects, failing to distinctly separate them. Prompt-to-Prompt (PtP) \cite{hertz2023prompttoprompt} observes a strong correlation between cross-attention maps and the layout of an image. Building on this, Structured Diffusion (StrD) \cite{feng2023trainingfree} experiments with averaging attention maps generated by different noun phrases for the same queried image latent representation. 
Yet, a simple average is inadequate for consistently generating images with multiple objects possessing complex attributes. Attend-and-Excite (AnE) \cite{chefer2023attend} proposes a novel approach of maximizing the attention map scores of object tokens by updating the latent at each sampling step. However, we note that artifacts and incorrect attribute binding are likely when AnE maximizes the attention weights of object tokens without any concerns on attributes. Similarly, A-star (A$^*$) \cite{agarwal2023star} aims to minimize the intersection of different objects' attention maps. In response, SynGen (SG) \cite{rassin2023linguistic} proposes an attribute-object pair-centric objective, aiming to minimize the distribution distance within the pair while maximizing it from other tokens, based on the assumption that normlized attention maps follow a multinomial distribution. Our findings (see Figs. \ref{key} and \ref{ane_image_comparison}-\ref{abc_image_comparison}) indicate that this approach still struggles with object neglect due to its pair-centric nature. We argue that multiple-object generation is more critical than attribute binding, as attributes cannot manifest without the presence of objects. Furthermore, in scenarios with multiple objects and no explicit attributes in a prompt, SG is degraded to  standard Stable Diffusion Models (SD) \cite{Rombach_2022_CVPR}. Diverging from these methods, Energy-Based Cross Attention (EBCA) \cite{park2023energy} introduces an Energy-Based Model (EBM) framework \cite{xie2016theory,yu2021unsupervised,yu2022latent,yu2024latent,yu2024learning} for queries and keys within cross-attention mechanisms, proposing updates to text embeddings instead of latent noise representations.

A closer look at both the fluctuations of attention intensities and the attention distributions of attribute-object pairs in these methods shed light on the root cause of the misalignment issues. As illustrated in Fig. \ref{key}, alignment in attribute-object attention maps (e.g., `purple crown' in SG) encourages attribute binding. However, attention map alignment alone does not guarantee complete semantic alignment, as the intensity levels of object attention maps are crucial in determining the presence of an object in the final image. For example, in the image generated by SG, the crown is notably absent. Conversely, despite successful generation of both objects with strong intensities, AnE binds the attribute `purple' incorrectly to the suitcase resulting from misaligned attention distributions of attribute-object pairs. 
Motivated by these key observations, we introduce a novel \textit{object-conditioned} Energy-Based Attention Map Alignment (EBAMA) method to hopefully address both the incorrect attribute binding and the catastrophic object neglect problems in a unified framework. Notably, we show that approximately maximizing the log-likelihood for a $z$-parameterized EBM effectively leads to optimizing an object-centric binding loss, which emphasizes both the object attention map intensity levels and the attribute-object attention map alignment.
We further develop  an object-centric intensity regularizer to prevent excessive shifts of objects towards their attributes, providing an extra degree of freedom balancing the trade-off between correct attribute binding and the necessary presence of objects.

We summarize our \textbf{contributions} as follows: i) we introduce a novel object-conditioned EBAMA method to address both the incorrect attribute binding and the catastrophic object neglect problems in text-controlled image generation;
ii) extensive qualitative and quantitative experiments, including human evaluation, on several challenging benchmarks demonstrate the superior performance of our method over strong previous approaches.
iii) We showcase that our approach has great promise in further enhancing the text-controlled image editing ability of diffusion models.

\section{Related Work}

\paragraph{EBM Framework for Attention Mechanisms}
Recent advancements in the theoretical exploration of attention mechanisms have increasingly embraced the EBM framework \cite{hopfield1982neural, mceliece1987capacity, ramsauer2020hopfield}.  Modern Hopfield Networks  \cite{ramsauer2020hopfield}  showcases that one of the proposed energy minima is equivalent to the attention mechanism.   Building on this groundwork, Energy Transformer \cite{hoover2023energy} designs an engineered energy function to extract the relationships between tokens.  Furthering this approach, EBCA \cite{park2023energy} first
formulates EBMs of query values condtioned on key values in    each cross-attention
layer. Similarly, our method seeks to exploit the theoretical potential of EBMs, focusing on the unique formulation of object-conditioned EBMs for attention maps.

\paragraph{Text-to-Image Diffusion Models}  
Most large-scale  text-to-image diffusion models \cite{Rombach_2022_CVPR, nichol2021glide, ramesh2022hierarchical, saharia2022photorealistic, balaji2022ediffi, hoogeboom2023simple} utilize classifier-free guidance \cite{ho2022classifier} for  improved conditional
synthesis results.   However, due to its strong linguistic and visual priors injected from the training dataset, these models suffer from diverse semantic misalignment issues related to the objects in the provided text prompts and their attribute(s) \cite{conwell2022testing, saharia2022photorealistic, rassin2022dalle, yu2022scaling, zhang2024flow}. Our approach   better aligns images with its provided texts by mitigating the issues of object neglect and incorrect attribute binding  without fine-tuning the diffusion models or additional training datasets.

\paragraph{Attention-Based Enhancement} PtP \cite{hertz2023prompttoprompt} identifies a correlation between cross-attention maps and image layout. Expanding on this, StrD \cite{feng2023trainingfree} experiments with averaging attention maps from different noun phrases to mitigate object neglect and attribute leakage. AnE \cite{chefer2023attend} introduces a method to enhance object presence by maximizing attention map weights for object tokens. SG \cite{rassin2023linguistic} proposes minimizing distribution distances of the attention maps within attribute-object pairs and maximizing the distances between the pairs and the other tokens. Different from the previous approaches, EBCA \cite{park2023energy} adopts an  EBM framework, focusing on updating text embeddings within cross-attention mechanisms. Our work also introduces an energy-inspired attention map alignment objective, while our objective has a specific emphasis on the object tokens.


\section{Background}

\subsection{Stable Diffusion Models}
For fair comparison with previous approaches, we also conduct experiments on open-sourced state-of-the-art Stable Diffusion Models (SD) \cite{Rombach_2022_CVPR}.
SD first encodes an image $x$ into the latent space using a pretrained encoder \cite{esser2021taming}, i.e., $z = \mathcal{E}(x)$. Given a text prompt $y$, SD  optimizes the conditional denoising autoencoder $\epsilon_\theta$   by minimizing the objective
\begin{align}
\mathcal{L}_\theta = \mathbb{E}_{t, \epsilon \sim \mathcal{N}(0,1), z \sim \mathcal{E}(x)} ||\epsilon - \epsilon_\theta (z_t, t, \phi(y))||^2,
\end{align}
where $\phi$ is a frozen CLIP text encoder \cite{radford2021learning},  $z_t$ is a noised version of the latent $z$, and the time step $t$ is uniformly sampled from $\{1, \ldots, T\}$.   During sampling, $z_T$ is randomly sampled from standard Gaussian and denoised iteratively by the denoising autoencoder $\epsilon_\theta$ from time $T$ to $0$. Finally, a decoder $\mathcal{D}$ reconstructs the image as $\tilde x = \mathcal{D}(z_0)$.

\subsection{Cross-Attention Mechanism}
In the cross-attention mechanism, $K$ is the linear projections of $W_y$, the CLIP-encoded text embeddings of text prompt $y$. $Q$ is the linear projection of the intermediate image representation parameterized by latent variables $z$. Given a set of queries $Q$ and keys $K$, the (unnormalized) attention features and (softmax-normalized) scores between these two matrices are
\begin{align}
\label{equ:attn}
A = \frac{QK^T}{\sqrt{m}},~
\Tilde{A} = \text{softmax} \left( \frac{QK^T}{\sqrt{m}} \right),
\end{align}
where $m$ is the feature dimension. 
We consider both attention features and scores for our modeling here, which we denote as $A_s$ and $\Tilde{A}_s$ for token $s$, respectively.

\begin{figure}[t]
    \centering
    \includegraphics[width=\textwidth]{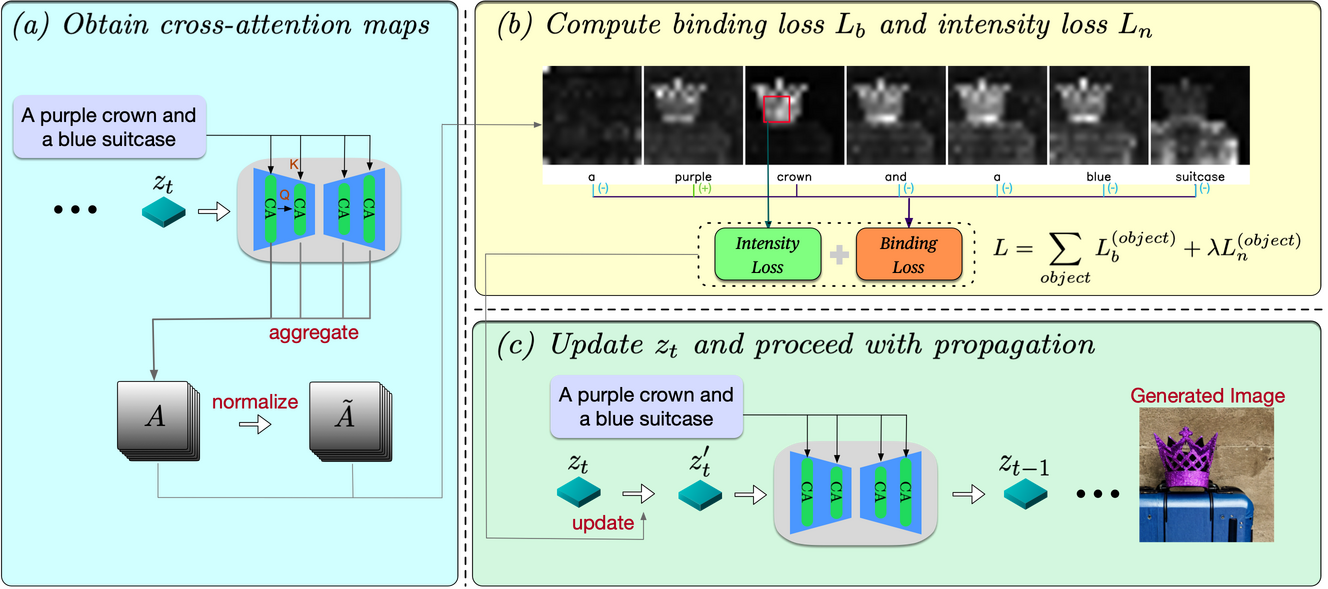}
    \caption{\textbf{An overview of our workflow for optimizing diffusion models.} It includes aggregation of attention maps, computation of object-centric attention loss, and updates to $z_t$. }
    \label{method}
\end{figure}

\section{Method}
The key idea of the proposed method derives from the \textit{object-oriented structure} that underlies most prompts for text-to-image generation. To be specific, syntactically the majority of prompts can be parsed as the modifiers and the entity-nouns, i.e., the nouns that correspond to \textit{objects} in the generated image, such as “A red metal \underline{\textit{crown}}”, and “A \underline{\textit{girl}} in red”, etc.  Based on the observation, we propose to exploit the object-oriented structure by employing an ensemble of object-centric cross-attention losses for inference-time  optimization. Our aim is to address the semantic misalignment issues including both the attribute binding (e.g., semantic leakage and attribute neglect \cite{rassin2023linguistic}) and catastrophic object neglect \cite{chefer2023attend}, with the principally derived and deliberately designed optimization objective. We then discuss the key components of our method as follows.

\subsection{Extraction of the Object-Oriented Structure}
Following the pre-processing step in \cite{rassin2023linguistic}, we parse the prompt using Spacy's \cite{honnibal2017spacy} transformer-based dependency parser to extract the object-oriented structure. We identify a set $S$ of object tokens ${s}$ from the prompt, whose tag is either NOUN (noun) such as `\underline{backpack}' or PROPN (proper noun) such as `\underline{Tesla} company' using the parser; we exclude nouns that serve as direct modifiers of other nouns. The remaining modifiers are grouped by their corresponding object tokens, denoted as the modifier sets for each object token $s$, i.e., $\mathcal{M}(s)$. Note that $\mathcal{M}(s) = \emptyset$ if there are no modifiers corresponding to the object token $s$. We refer to supplementary material for more details about the parsing process.

\subsection{Object-Conditioned Energy-Based Model}
We assume that the distribution of the modifier tokens $l \in \bigcup_s \mathcal{M}(s)$ given the object token $s$ is
\begin{equation}
\label{equ:ebm}
    p_{z}(l | s) = \frac{1}{Z(s)} \exp\left(f(A_l, A_s)\right),
\end{equation}
where $Z(s) = \sum_l \exp\left(f(A_l, A_s)\right)$ is the normalizing constant and $f$ is the negative energy function. How to choose the energy function for attention maps remains an interesting and open problem. Prior works \cite{ramsauer2020hopfield, park2023energy} utilize a log-sum-exp term to model the exponential interaction and alignment between state patterns (query) and stored patterns (key).  However, the alignment measurement  for attentions maps across different tokens is unclear. To bridge the gap, we propose the application of  a non-crafted yet effective  energy function --- cosine similarity defined as $f(A_l, A_s) = \left<A_l, A_s\right>/(||A_l|| \cdot ||A_s||)$. The efficacy of this choice of energy function is validated in the experimental section. Eqn. (\ref{equ:ebm}) therefore defines a multinomial token distribution as a $z$-parameterized conditional energy-based model, where $z$ is the latent variables of SD. The inference-time optimization over the latent variables $z$ is then equivalently maximizing the log-likelihood of this EBM, which increases the probabilities of the syntatically related modifier tokens of the given object $s$. To be specific, it can be shown that (see the supplementary material)
\begin{equation}
\label{equ:ebm_grad}
\nabla_z \log p_{z}(l | s) = 
\nabla_z f(A_l, A_s) - 
\mathbb{E}_{p_{z}(l | s)} \left[
\nabla_z f(A_l, A_s)
\right].
\end{equation}
Since the vocabulary size of modifier tokens can be large in practice (in the order of $10^4$), we consider resorting to negative sampling \cite{mikolov2013distributed} for the approximation of the expectation term, where we uniformly sample tokens unrelated to the object token and calculate the Monte Carlo average. This particular implementation choice of Eqn. (\ref{equ:ebm_grad}) then leads to the object-centric attribute binding loss below.
 
\subsection{Object-Conditioned Energy-Based Attention Map Alignment}
For each object token $s \in S$, we design the following two components that consist of the object-centric attention loss:

\paragraph{\textbf{Object-centric attribute binding}} 
First, instead of operating on the noun-modifier normalized attention score pairs as in \cite{rassin2023linguistic}, we focus on optimizing the log-likelihood of the object-conditioned EBM using negative sampling. This gives us the attribute binding loss:
\begin{align}\label{loss_b}
    L_b^{(s)} =   -\frac{1}{|\mathcal{M}(s)|}\sum_{l\in \mathcal{M}(s)} f(A_s,A_l)   
    + \frac{1}{N-|\mathcal{M}(s)|-1}\sum_{l \notin \mathcal{M}(s), l \neq s} f(A_s, A_l),
\end{align}
whose negative gradient w.r.t. $z$ could be seen as the Monte Carlo approximation of Eqn. (\ref{equ:ebm_grad}).
The goal of $L_b^{(s)}$ is to: i) maximize the cosine similarity between the given object $s$ and its syntactically-related modifier tokens, while ii) enforcing the repulsion of grammatically unrelated ones in the feature space. 
Note that the loss above only applies to the cases where $\mathcal{M}(s)$ is a non-empty set. For the case where $\mathcal{M}(s) = \emptyset$, only the \textbf{repulsive term} of Eqn. (\ref{loss_b}) is used.

\paragraph{\textbf{Object-centric intensity regularizer}} Although the proposed attribute binding loss mitigates the catastrophic object neglect problem (see $\lambda=0$ entries in Tab. \ref{ane_table}), we observe that the object-related attention feature can still be overly shifted when there are multiple modifier tokens in the $\mathcal{M}(s)$ or multiple object tokens in a prompt; this could again potentially leads to the object neglect phenomenon. To address this issue, we follow \cite{chefer2023attend} and propose an object-centric intensity regularizer to maintain the attention intensity level of object $s$:
\begin{align}
    L_n^{(s)} = -||\mathcal{K}(\Tilde{A_s})||_\infty,
\end{align}
where $\mathcal{K}$ is a 3x3 Gaussian kernel, and $||\cdot||_\infty$ denotes the maximum value of a vector. We use the attention scores in Eqn. (\ref{equ:attn}) as its input. We refer to $||\mathcal{K}(\Tilde{A_s})||_\infty$ as the  {\textbf{intensity level}} of the object token $s$.

The final object-centric attention loss $L$ is the linear combination of the binding loss and the regularizer, i.e. 
\begin{align}\label{loss}
    L = \sum_{s\in S} L^{(s)} =\sum_{s\in S} L_b^{(s)} + \lambda L_n^{(s)},
\end{align}
where intensity weight $\lambda$ is a hyper-parameter to specify. $\lambda > 0$ enforces the presence of object $s$, but excessively intensified object attention can hinder the attribute binding performance and lower visual image quality. We provide empirical analysis on how to tune the weight in practice (see Section \ref{sec:ablation} and supplementary material for details).

\subsection{Workflow}
Our workflow is illustrated in Fig. \ref{method}. To begin, at each time step $t$, we aggregate the attention map features   denoted as $A$   at a resolution of 16x16. This aggregation is performed after one step of propagating $z_t$ through the denoising model. 
Subsequently, we calculate the object-centric attention loss, as described in Eqn. (\ref{loss}).  
Finally, we backpropagate the computed loss and update $z_t$  for each time step, following the formula $z_t' \leftarrow z_t - \alpha \nabla_{z_t} L$, where $\alpha$ represents the step size. In our experimental setup, we set $\alpha=20$.
Note that we only perform updates on $z_t$ during the first half of the sampling steps, which corresponds to 25 steps since we use  a DDIM sampler with a total of 50 steps.
More details about the workflow can be found in the supplementary material.

\section{Experiments}

\begin{table*}[t]
\centering
\caption{\textbf{Comparison of 
Full Sim.,  Min. Sim., and  T-C Sim. across
different methods on the AnE dataset.}  Note that the performance of SG on Animal-Animal is degraded to SD, as the prompts do not contain any attribute-object pairs. The best and second-best performances are marked in bold numbers and underlines, respectively; tables henceforth follows this format.}
\label{tab:comparison}
\begin{tabular}{lccccccccc}
\toprule
& \multicolumn{3}{c}{\scriptsize Animal-Animal} & \multicolumn{3}{c}{\scriptsize Animal-Object} & \multicolumn{3}{c}{\scriptsize Object-Object} \\
\cmidrule(lr){2-4} \cmidrule(lr){5-7} \cmidrule(lr){8-10}
\scriptsize Method  & \tiny Full Sim.  & \tiny Min. Sim.  & \tiny T-C Sim.  &  \tiny Full  Sim.  &  \tiny Min.  Sim.  &  \tiny T-C Sim.  &  \tiny Full  Sim.  &  \tiny Min. Sim.  &  \tiny T-C Sim.  \\
\midrule
\scriptsize SD\cite{Rombach_2022_CVPR} & \footnotesize 0.311 & 0.213 & 0.767 & 0.340 & 0.246 & 0.793 & 0.335 & 0.235 & 0.765 \\
\scriptsize CD\cite{liu2022compositional}& 0.284 & 0.232 & 0.692 & 0.336 & 0.252 & 0.769 & 0.349 & 0.265 & 0.759 \\
\scriptsize StrD\cite{feng2023trainingfree} & 0.306 & 0.210 & 0.761 & 0.336 & 0.242 & 0.781 & 0.332 & 0.234 & 0.762 \\
\scriptsize EBCA\cite{park2023energy}& 0.291& 0.215 & 0.722& 0.317 & 0.229 & 0.732 & 0.321 & 0.231 & 0.726 \\
\scriptsize AnE\cite{chefer2023attend} & {0.332} & {0.248} & {0.806} & 0.353 & {0.265} & {0.830} & {0.360} & {0.270} & {0.811} \\ 
\scriptsize SG\cite{rassin2023linguistic}  & 0.311 & 0.213 & 0.767 & {0.355} & 0.264& {0.830} & 0.355 & 0.262 & {0.811}\\ 
\midrule 
\rowcolor{gray}
\scriptsize Ours($\lambda = 0)$ &  \textbf{0.340} & \underline{0.255} & \underline{0.814} & \underline{0.362} & \textbf{0.271} & \textbf{0.851} & \underline{0.360} & \underline{0.270} &  \underline{0.823}\\ 
\rowcolor{gray}
\scriptsize Ours & \textbf{0.340} & \textbf{0.256} & \textbf{0.817} &  \textbf{0.362} & \underline{0.270} &\textbf{0.851} & \textbf{0.366} & \textbf{0.274 }& \textbf{0.836}\\
\bottomrule
\end{tabular} \label{ane_table}
\end{table*}

We compare our generation results with SD, CD, StrD, EBCA, 
 AnE, and SG on two artificial datasets, AnE dataset \cite{chefer2023attend} and DVMP \cite{rassin2023linguistic}, and one natural-language dataset, ABC-6K \cite{feng2023trainingfree}. We refer to supplementary material for implementation details and computational efficiency comparison.


\paragraph{Datasets} The AnE dataset \cite{chefer2023attend} comprises three benchmarks: Animal-Animal, Animal-Object, and Object-Object. Each benchmark varies in complexity and incorporates a combination of potentially colored animals and objects. The prompt patterns for these benchmarks include two unattributed animals, one unattributed animal and one attributed object, and two attributed objects, respectively. The DVMP dataset \cite{rassin2023linguistic} features a diverse set of objects (e.g., daily objects, animals, fruits, etc.) and diverse modifiers including colors, textures and so on. It features more than three attribute descriptor per prompt. The ABC-6K dataset \cite{feng2023trainingfree}, derived from natural MSCOCO \cite{lin2014microsoft} captions, includes prompts with at least two color words modifying different objects. The first two datasets are artificial, while ABC-6K is composed of natural language captions.

 \begin{wrapfigure}{l}{0.48\textwidth} 
  \centering
  \includegraphics[width=0.48\textwidth]{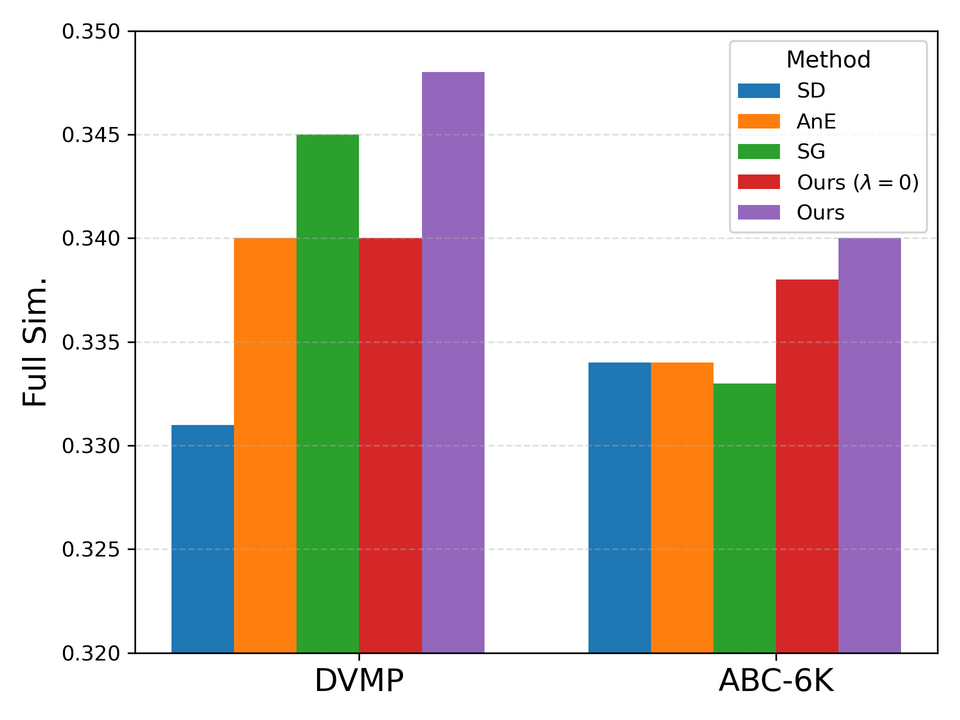 }
  \caption{\textbf{Full Sim. results on DVMP and ABC-6K datasets.} We randomly sample 200 prompts from each dataset and generate 4 images for each prompt.}
  \label{abcDVMP}
\end{wrapfigure}

\subsection{Quantitative Comparison}
Following the setting of \cite{chefer2023attend}, we compare the Text-Image Full Similarity (Full Sim.), Text-Image Min Similarity (Min. Sim.), and Text-Caption Similarity (T-C Sim.) on the AnE dataset. Additionally, we present the Full  Sim. results on the DVMP and ABC-6K datasets.

Full Sim. is the CLIP \cite{radford2021learning} cosine similarity score between the text prompt and the generated image. 
Furthermore, we assess CLIP similarity for the most neglected object independently from the full text by computing the CLIP similarity scores between each sub-prompt and the generated image. The smaller score is denoted as Min. Sim..  T-C Sim. is  the average CLIP similarity between the prompt and all captions generated by a pre-trained BLIP image-captioning model \cite{li2022blip} with the generated image as input.  

We generate 64 images for each prompt using the same seed across all methods and compute the average score between each prompt and its corresponding images. Our method consistently demonstrates superior performance across all datasets, as shown in Tab. \ref{ane_table}. 
We stress the following advantages of our method:
(1) Our method distinguishes itself from SG by its adaptability to the Animal-Animal dataset, even when the prompts lack specific attributes;
(2) Our method with $\lambda=0$ surpasses AnE and SG in all cases, underscoring the effectiveness of our object-centric attribute binding loss;
(3) As the dataset becomes more complicated, our method with hyper-picked $\lambda$ gains a more significant advantage over that with $\lambda=0$.   

In Fig.~\ref{abcDVMP}, despite SG's deliberate design for multi-attribute prompts, our method consistently surpasses SG. Furthermore, in the ABC-6K dataset, AnE and SG exhibit performance levels similar to that of SD, while our method consistently achieves superior results. These advantages are further confirmed by our human evaluation in \ref{sec:human_eval}.

\subsection{Qualitative Comparison}

\begin{figure*}[ht]
\centering
\renewcommand{\arraystretch}{0.8} 
\setlength{\tabcolsep}{4pt} 
\begin{tabular}{cccc}
    \raisebox{0.05\textwidth}{ \hspace{-5pt} \rotatebox{90}{SD} }& \includegraphics[width=0.25\textwidth]{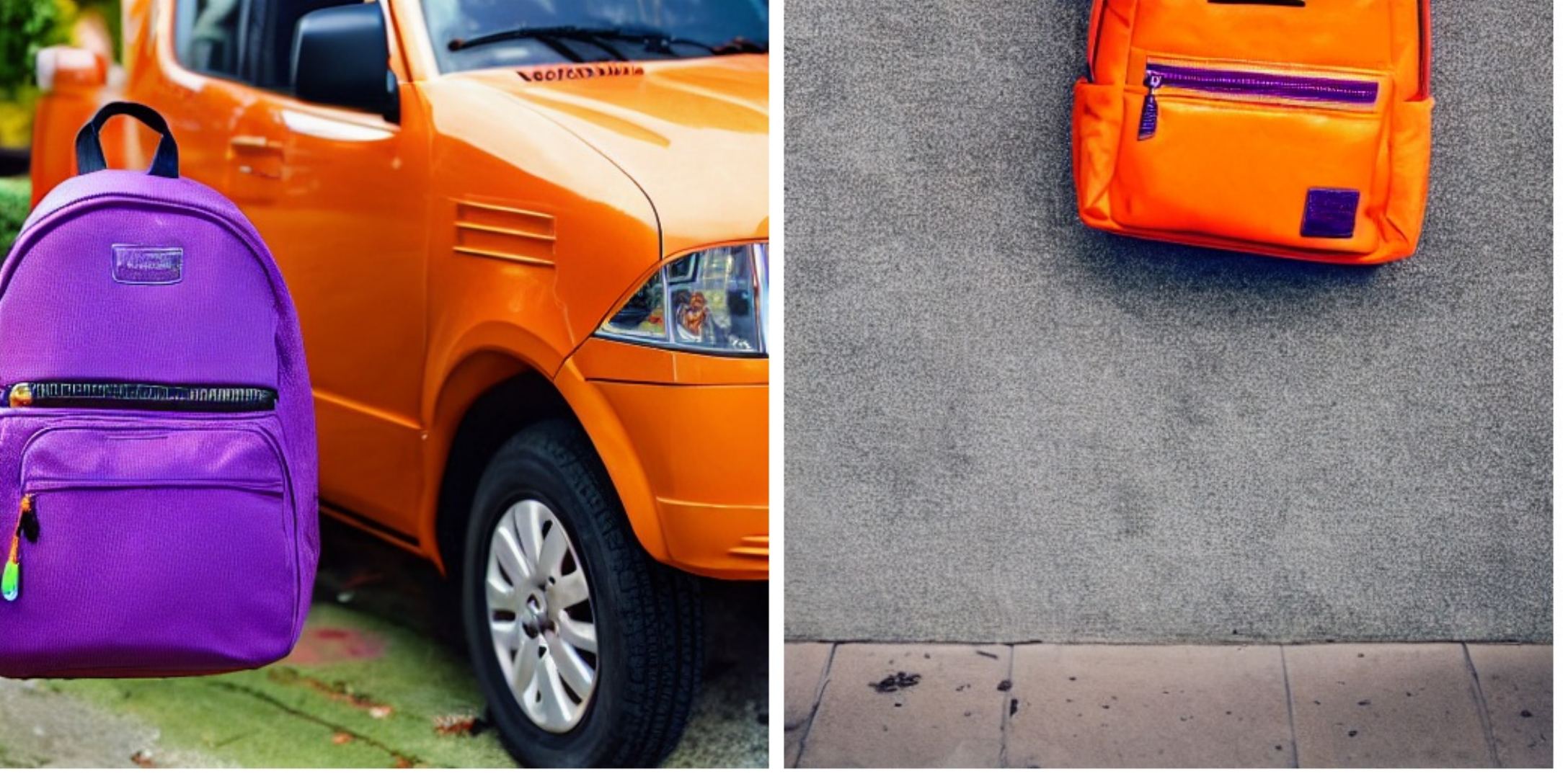} & \includegraphics[width=0.25\textwidth]{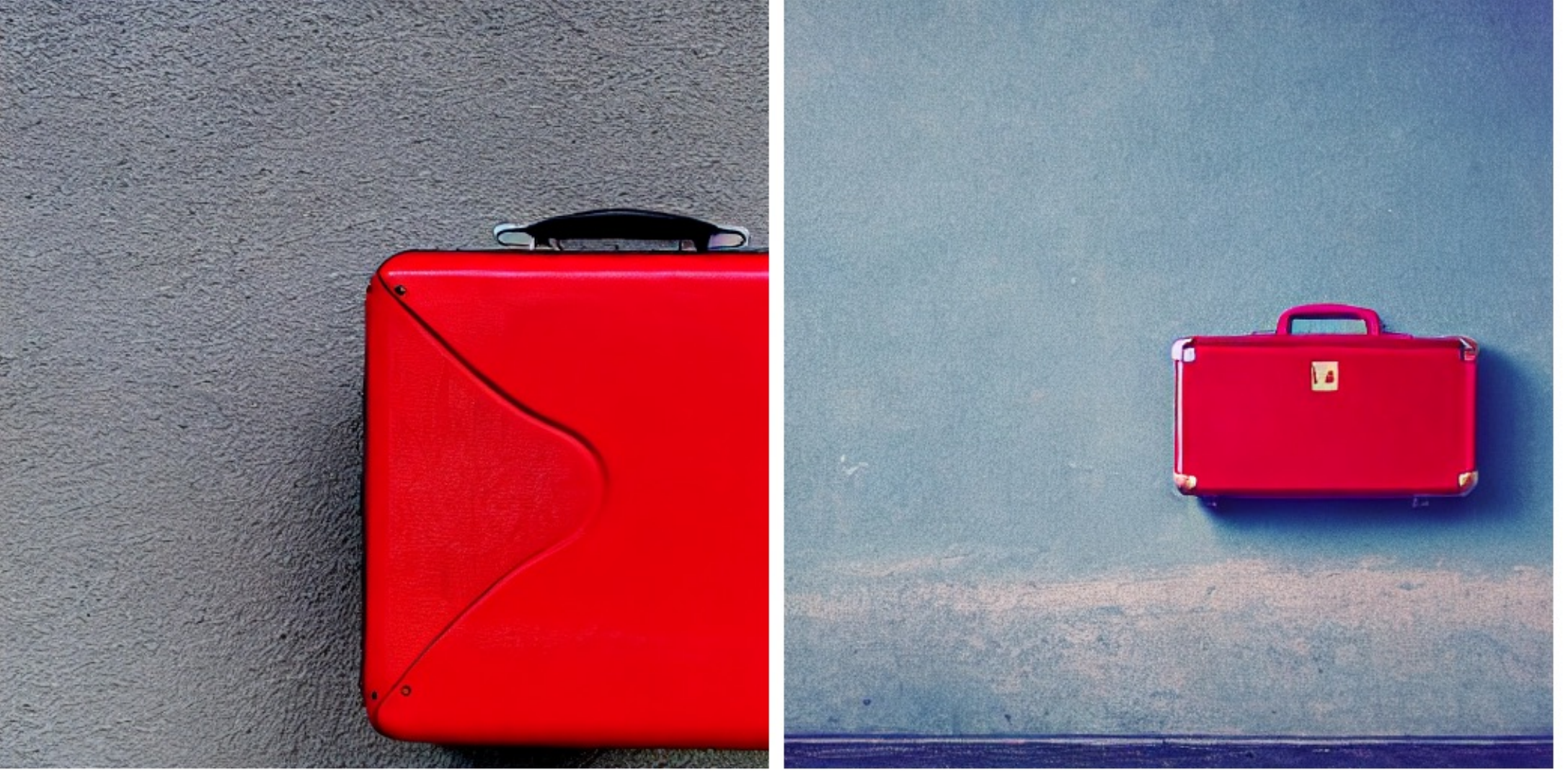} & \includegraphics[width=0.25\textwidth]{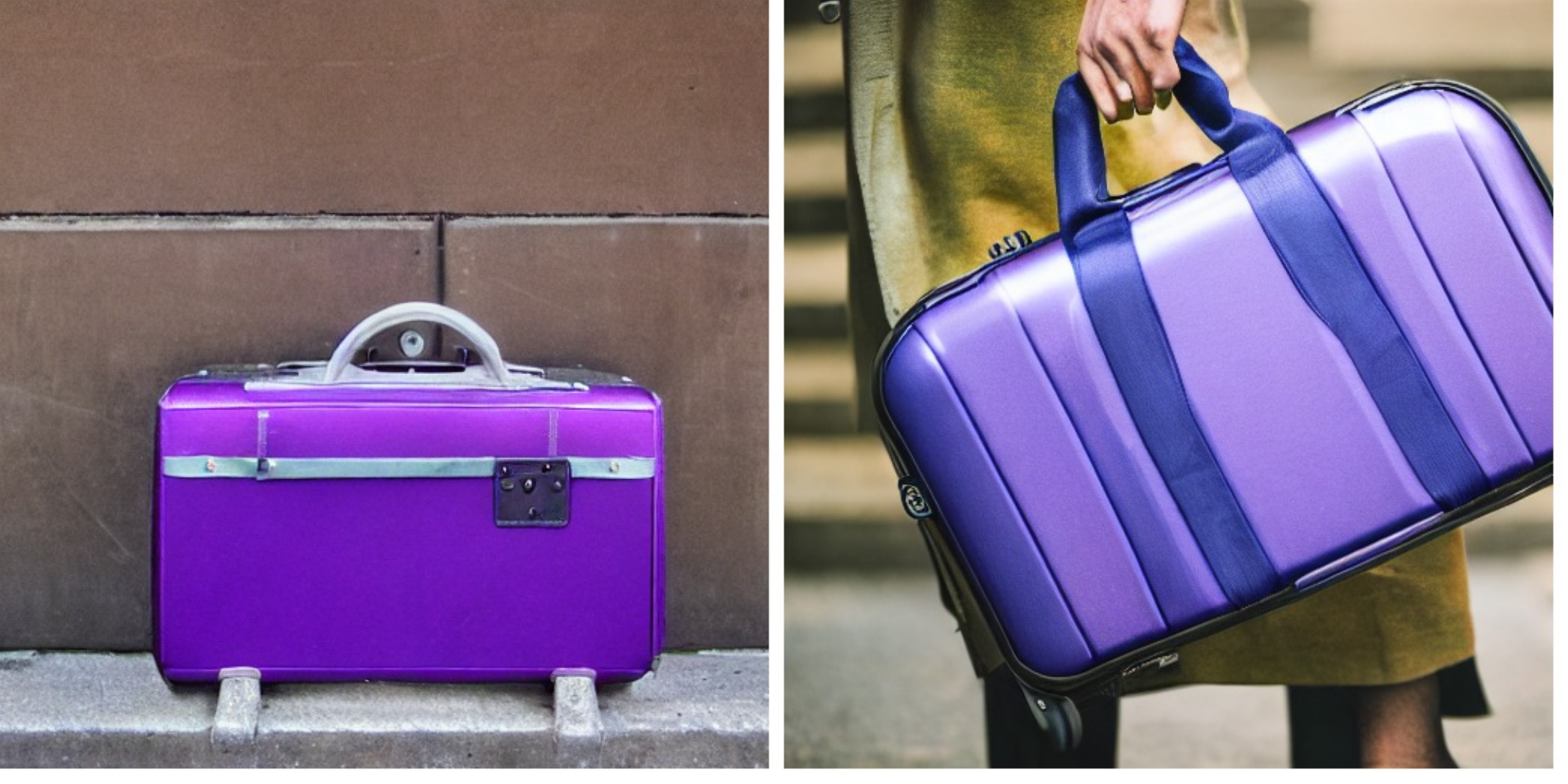}   \\
    \raisebox{0.05\textwidth}{ \hspace{-5pt}\rotatebox{90}{AnE}} & \includegraphics[width=0.25\textwidth]{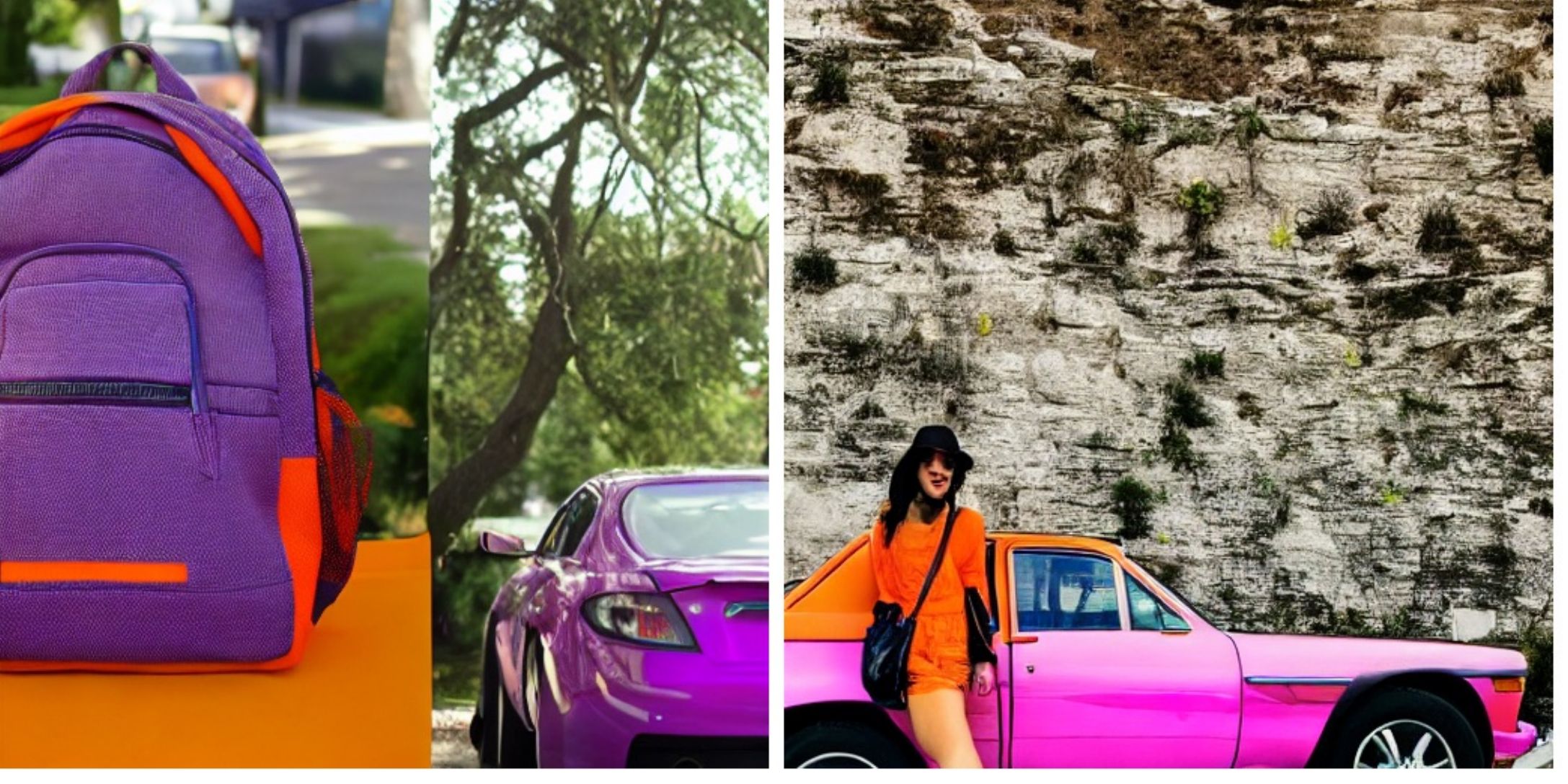} & \includegraphics[width=0.25\textwidth]{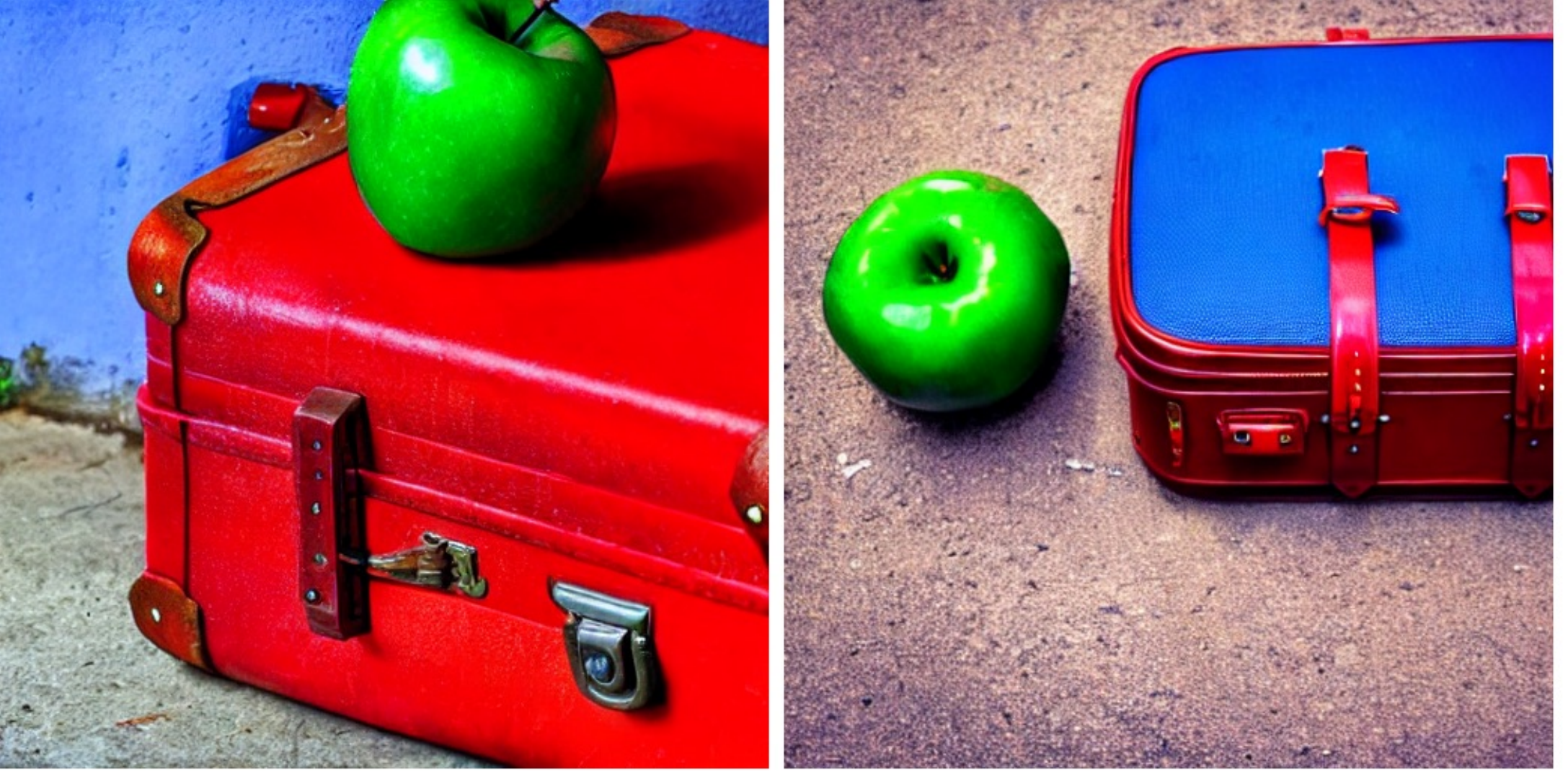} & \includegraphics[width=0.25\textwidth]{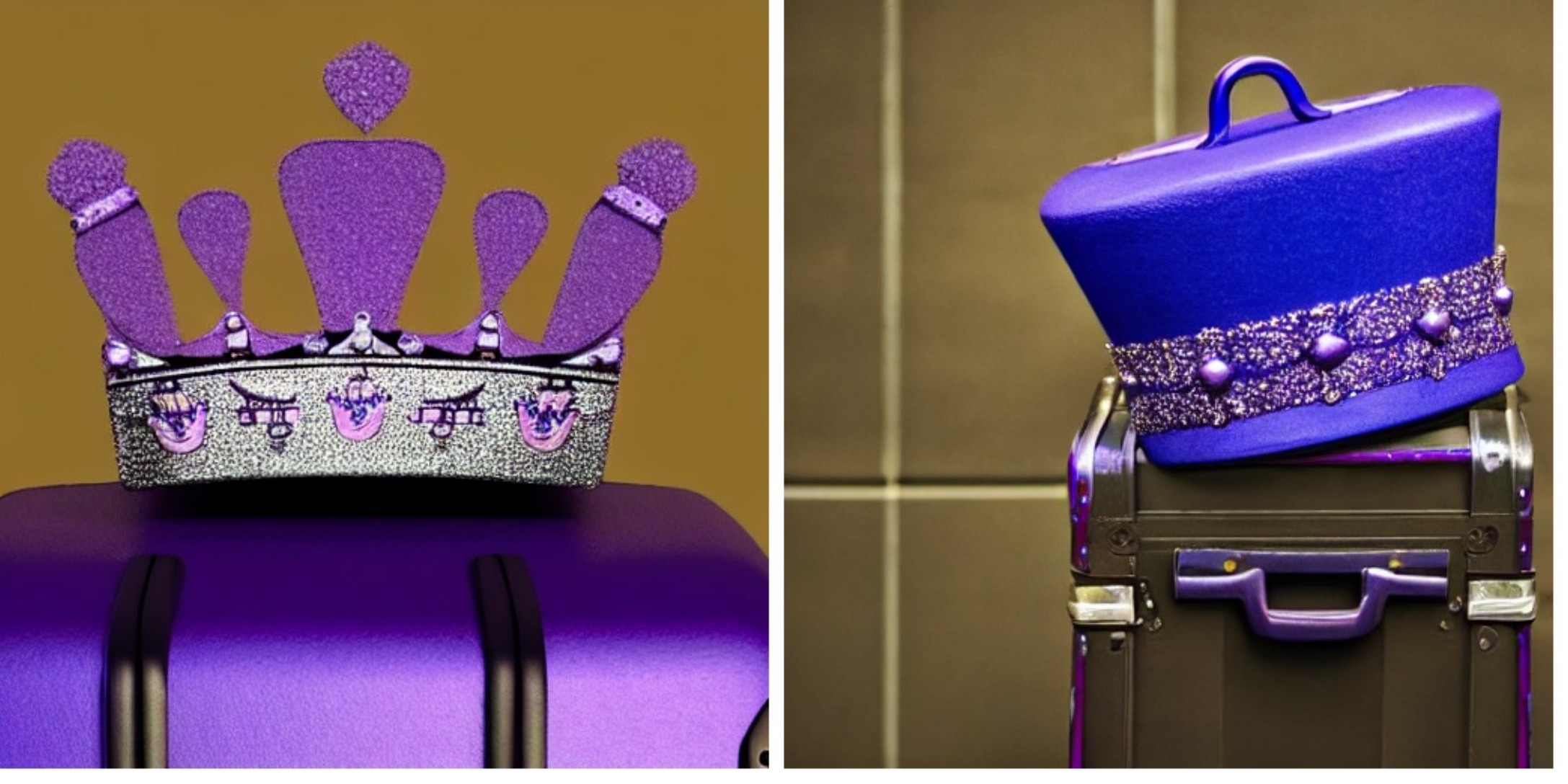}  \\
    \raisebox{0.05\textwidth}{ \hspace{-5pt}\rotatebox{90}{SG}} & \includegraphics[width=0.25\textwidth]{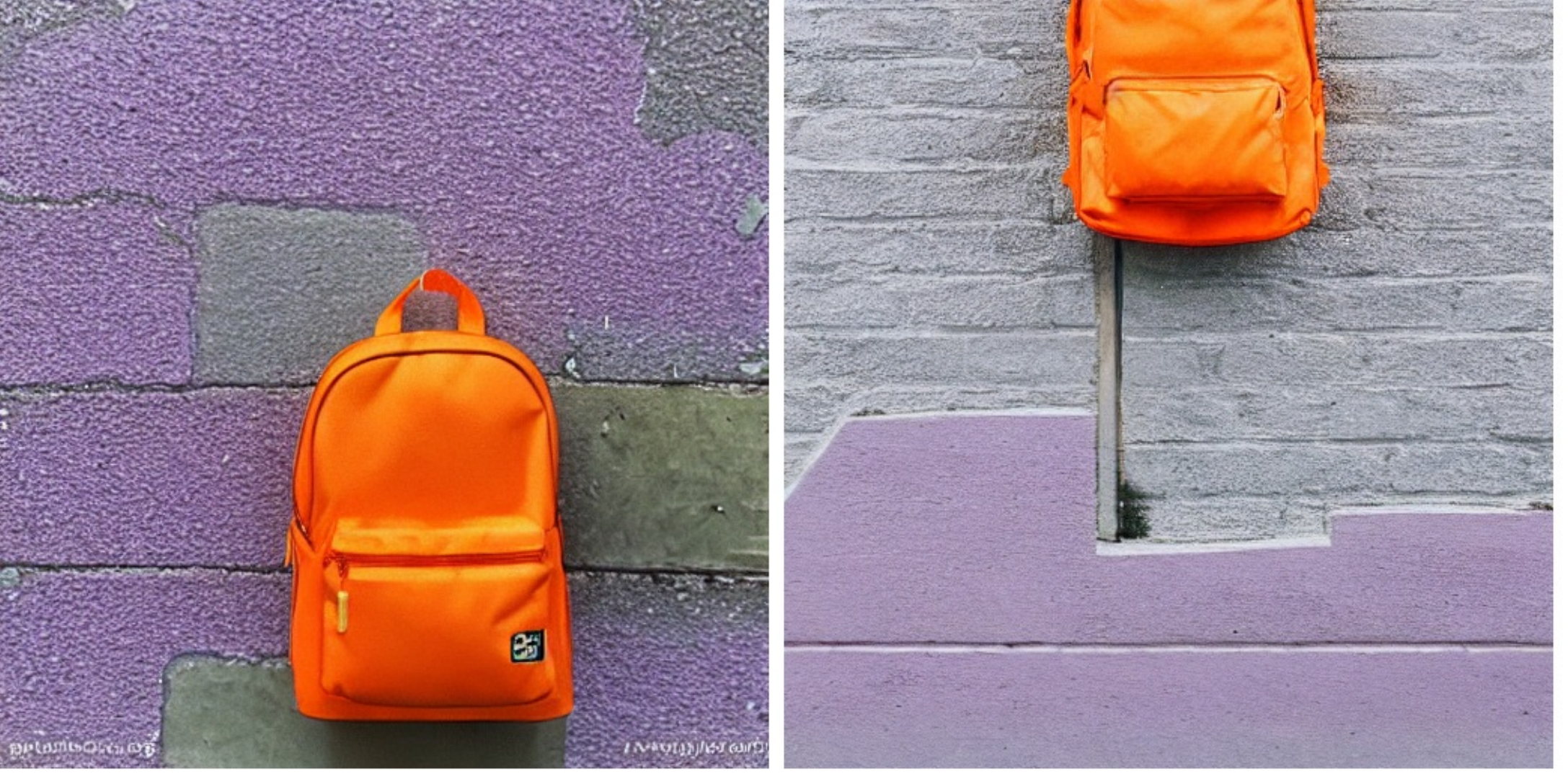} & \includegraphics[width=0.25\textwidth]{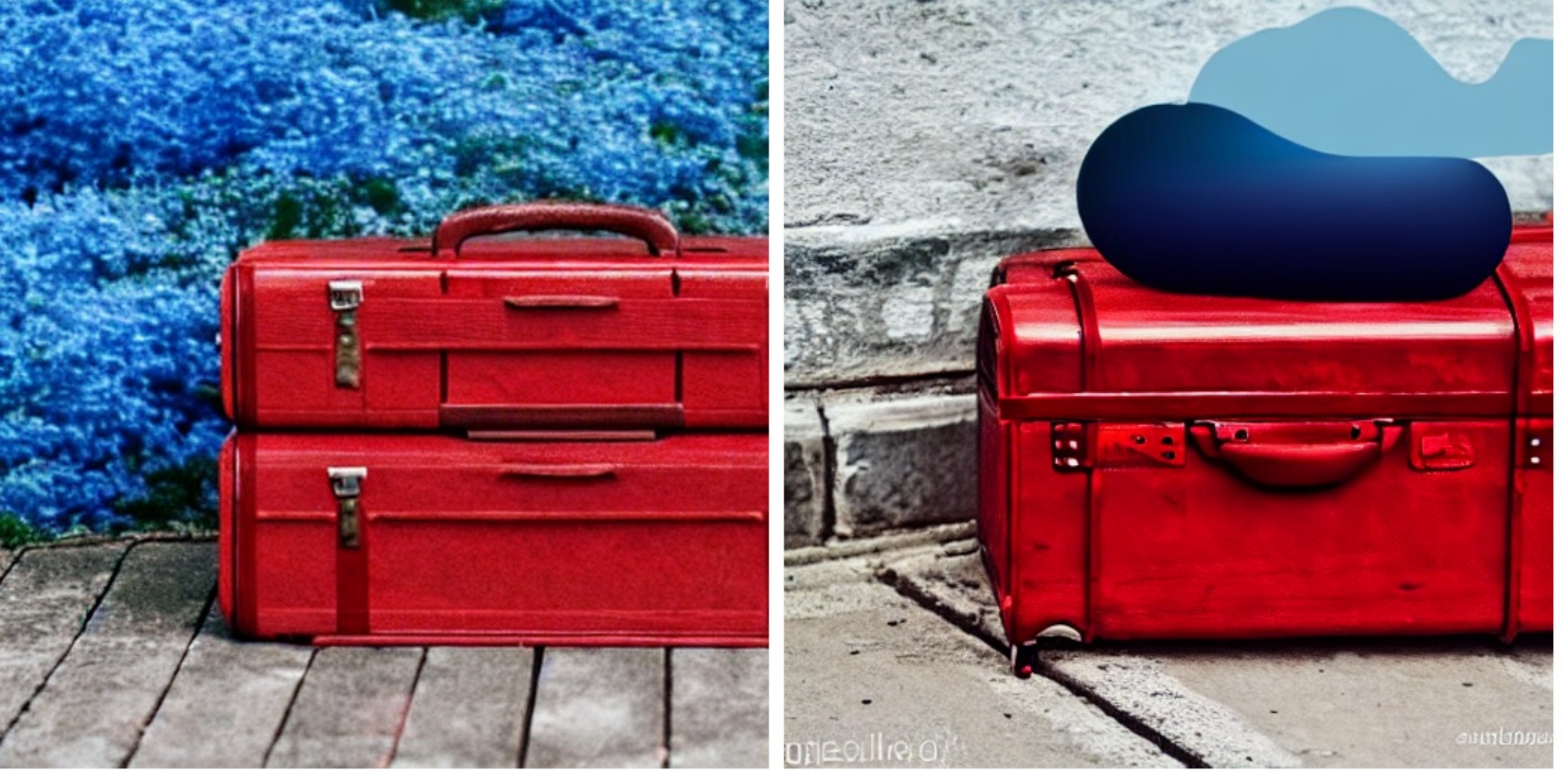} & \includegraphics[width=0.25\textwidth]{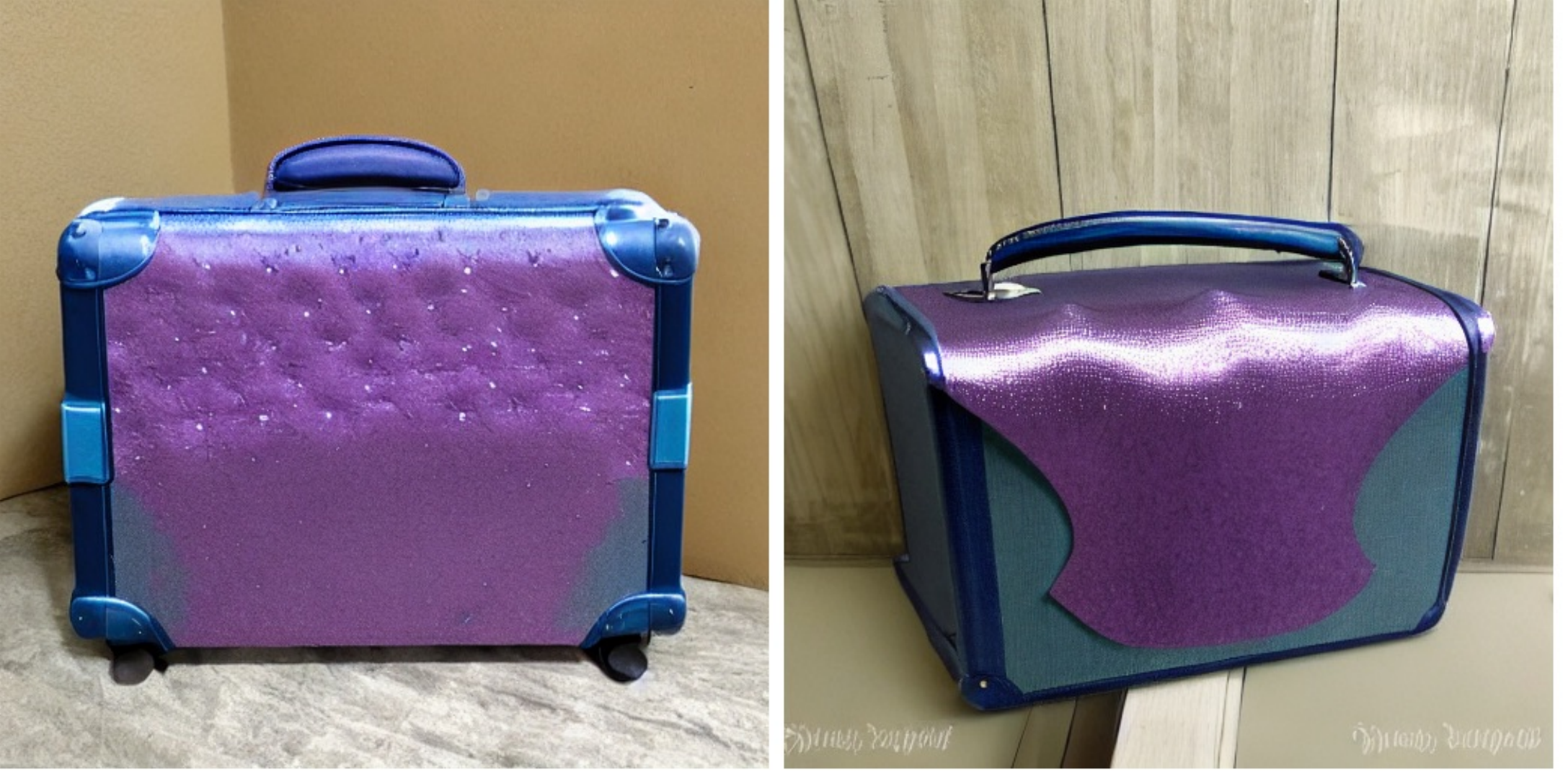}  \\
     \raisebox{0.05\textwidth}{ \hspace{-5pt}\rotatebox{90}{Ours}} & \includegraphics[width=0.25\textwidth]{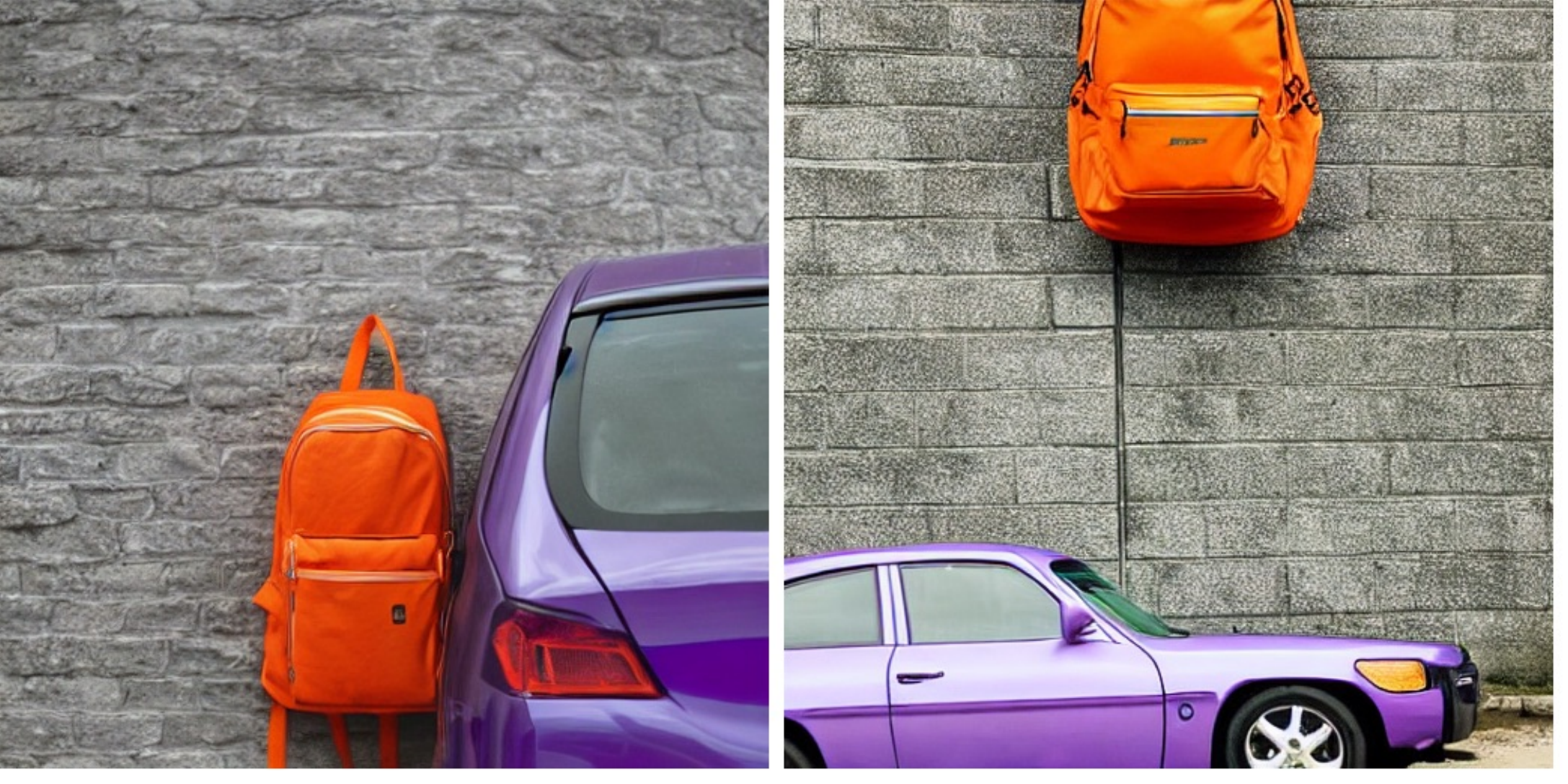} & \includegraphics[width=0.25\textwidth]{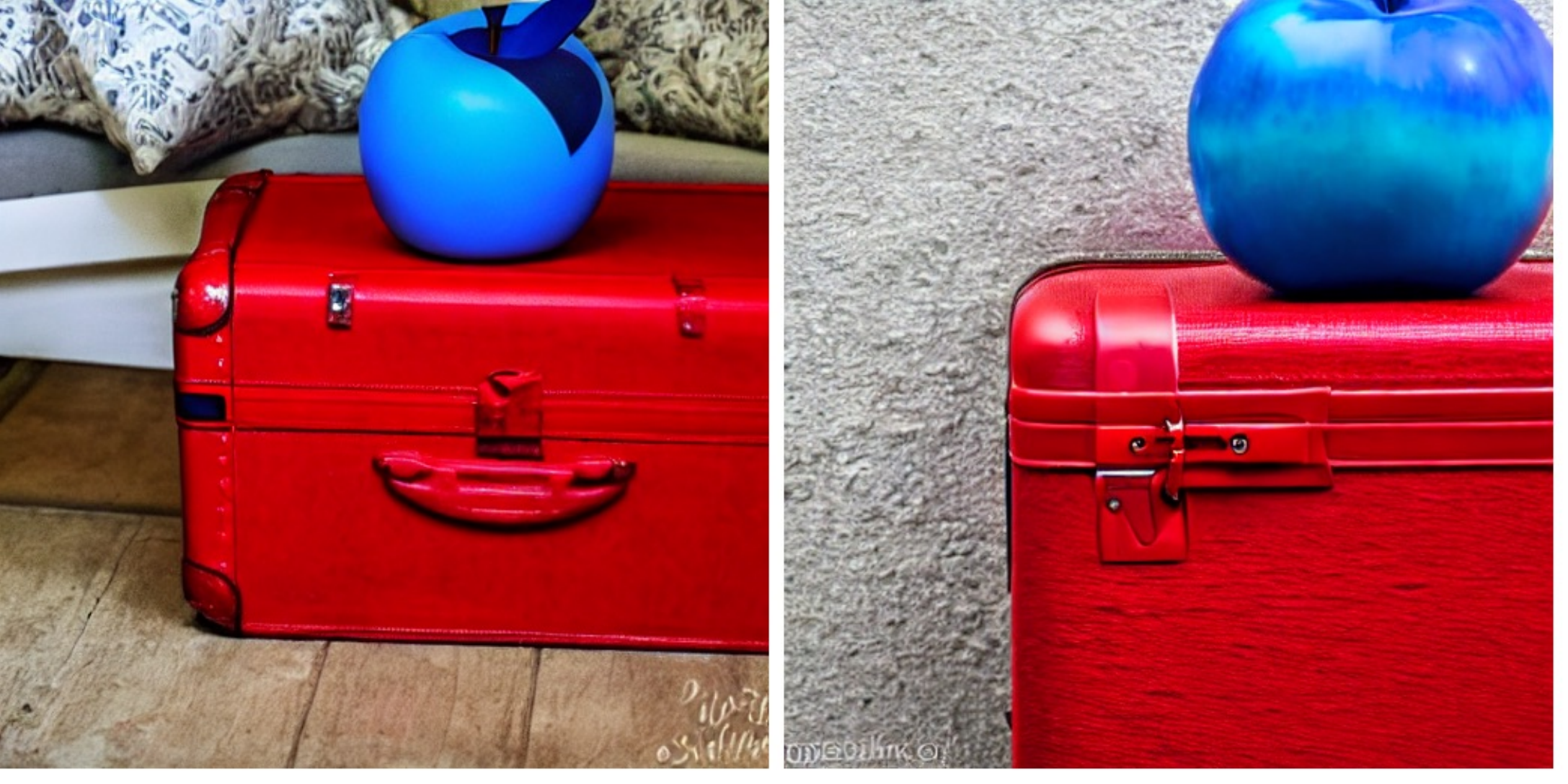} & \includegraphics[width=0.25\textwidth]{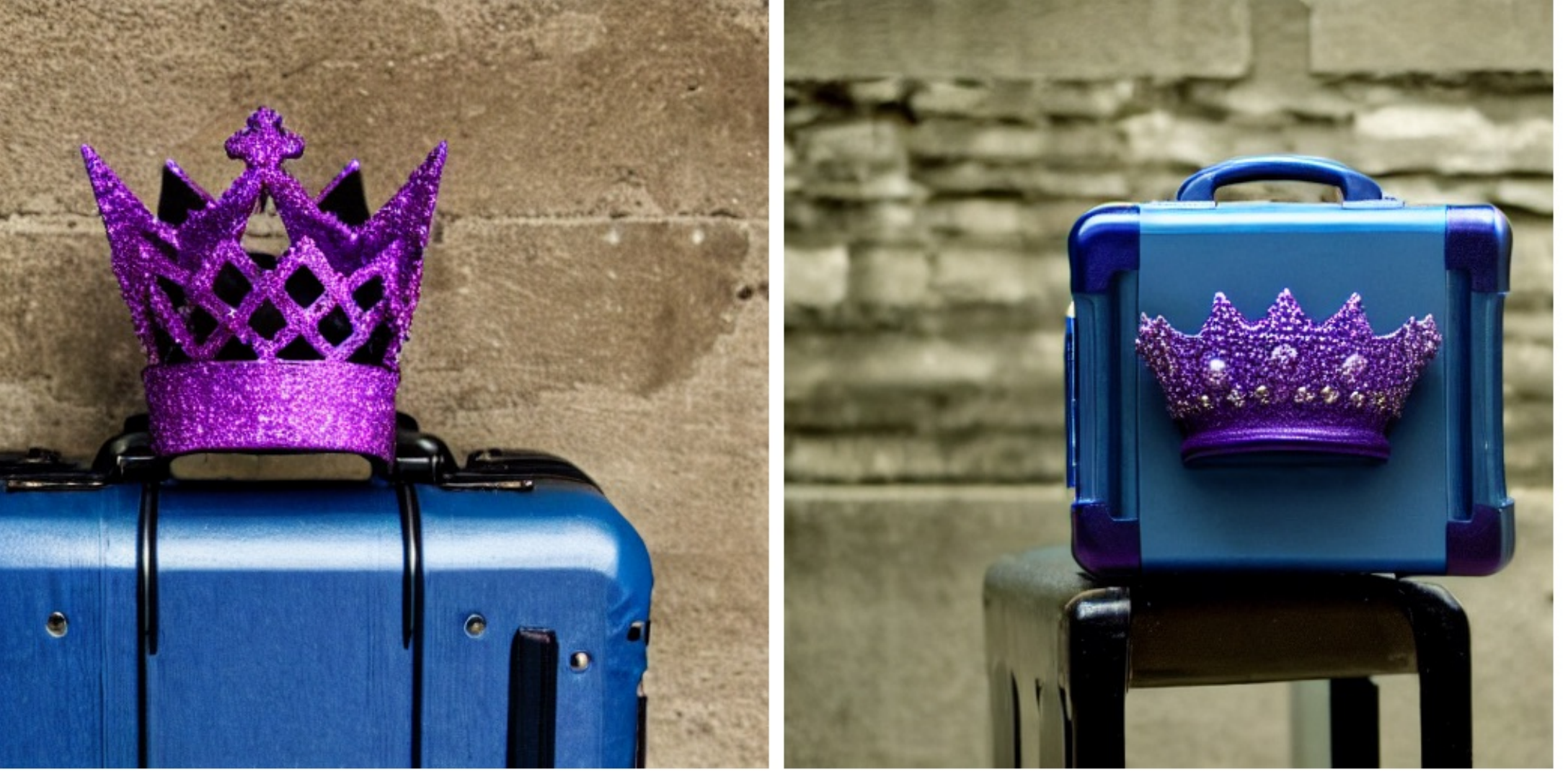}   \\
       &   \multicolumn{1}{p{0.25\textwidth}}{\small  \centering   an \textcolor{orange}{orange} \underline{backpack} and  a \textcolor{Orchid}{purple} \underline{car}}  & 
     \multicolumn{1}{p{0.25\textwidth}}{\small \centering   a \textcolor{red}{red} \underline{suitcase} and a \textcolor{blue}{blue} \underline{apple} }  &
  \multicolumn{1}{p{0.25\textwidth}}{\small   \centering    a \textcolor{Orchid}{purple} \underline{crown}  and a \textcolor{blue}{blue} \underline{suitcase}}  
\end{tabular}
\caption{\textbf{Qualitative comparison on the AnE dataset.} Each column shares the same random seed.}
\label{ane_image_comparison}
\end{figure*}

 \begin{figure*}[!h]
\centering
\renewcommand{\arraystretch}{0.8} 
\setlength{\tabcolsep}{4pt} 
\begin{tabular}{cccc}
    \raisebox{0.05\textwidth}{ \hspace{-5pt} \rotatebox{90}{SD} }& \includegraphics[width=0.25\textwidth]{  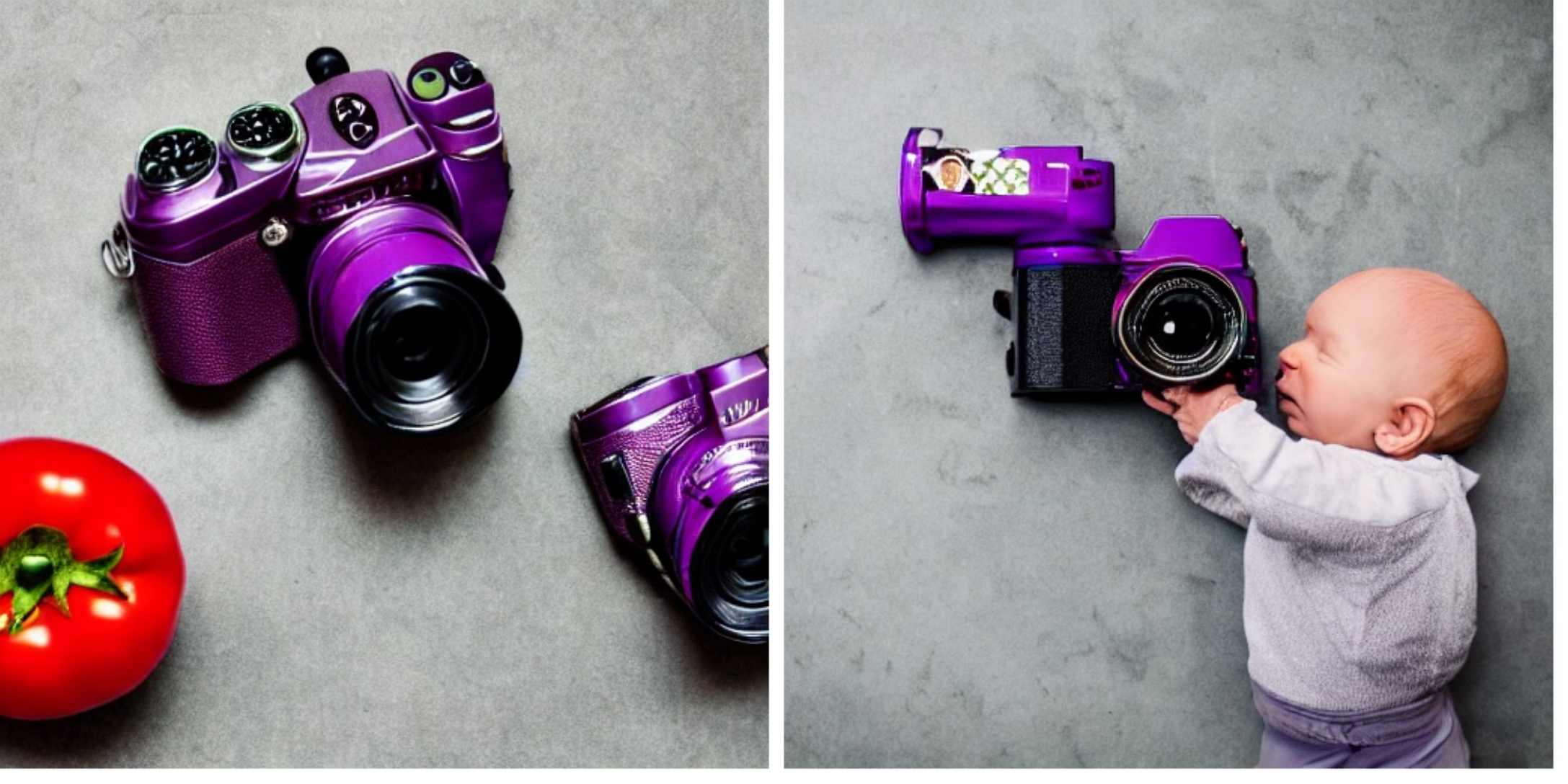} & \includegraphics[width=0.25\textwidth]{  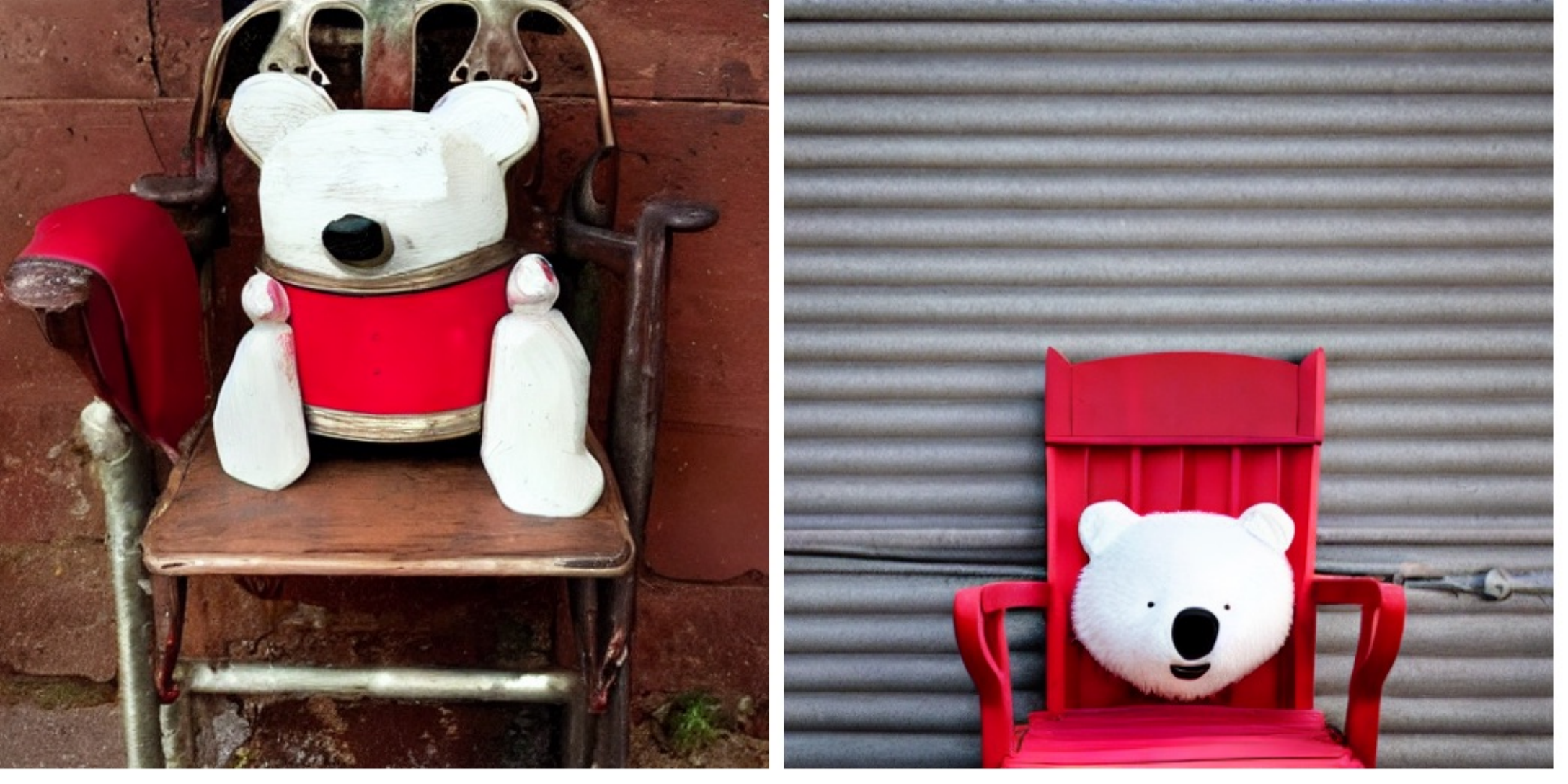 } & \includegraphics[width=0.25\textwidth]{  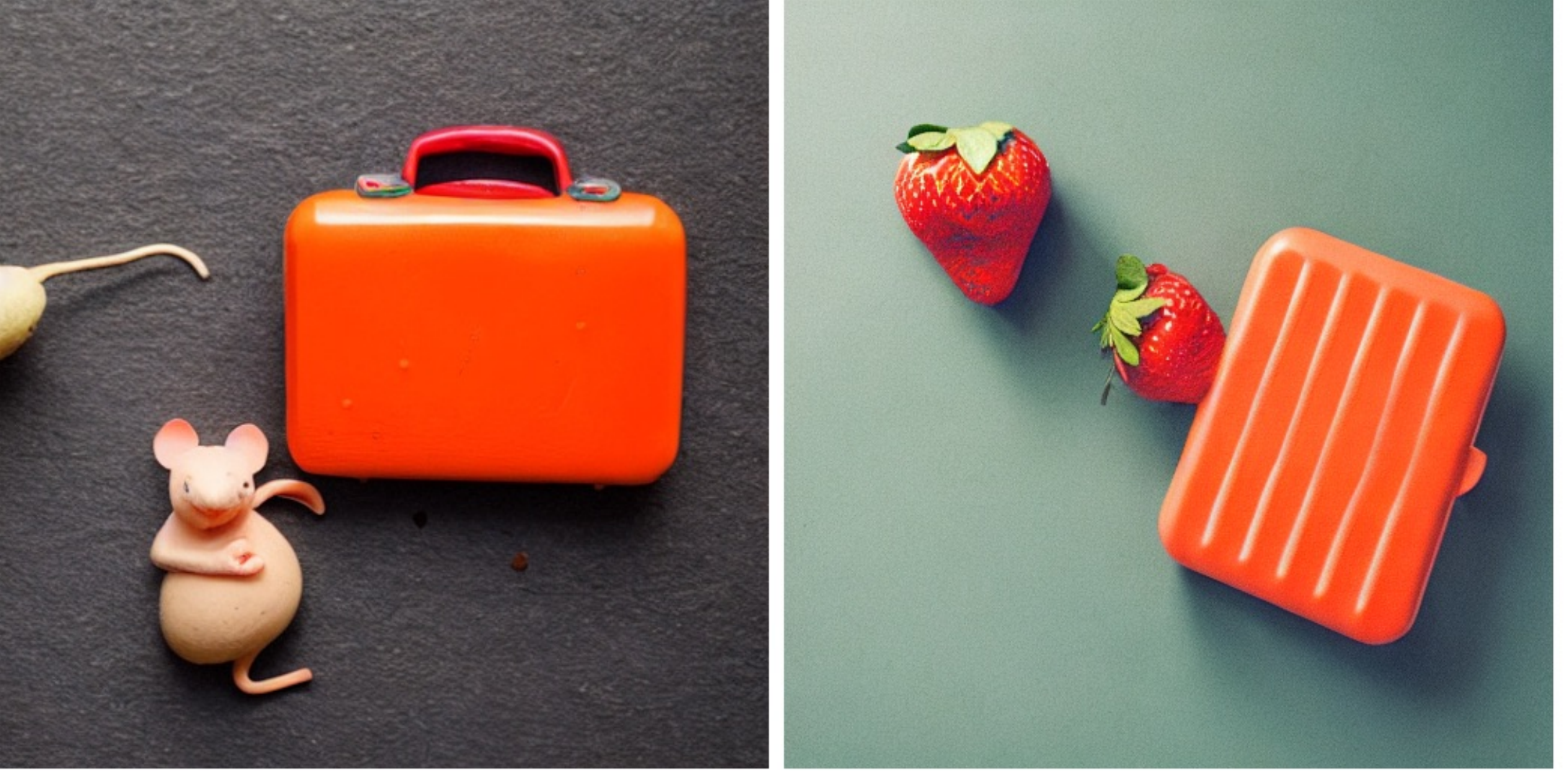 }   \\
    \raisebox{0.05\textwidth}{ \hspace{-5pt}\rotatebox{90}{AnE}} & \includegraphics[width=0.25\textwidth]{  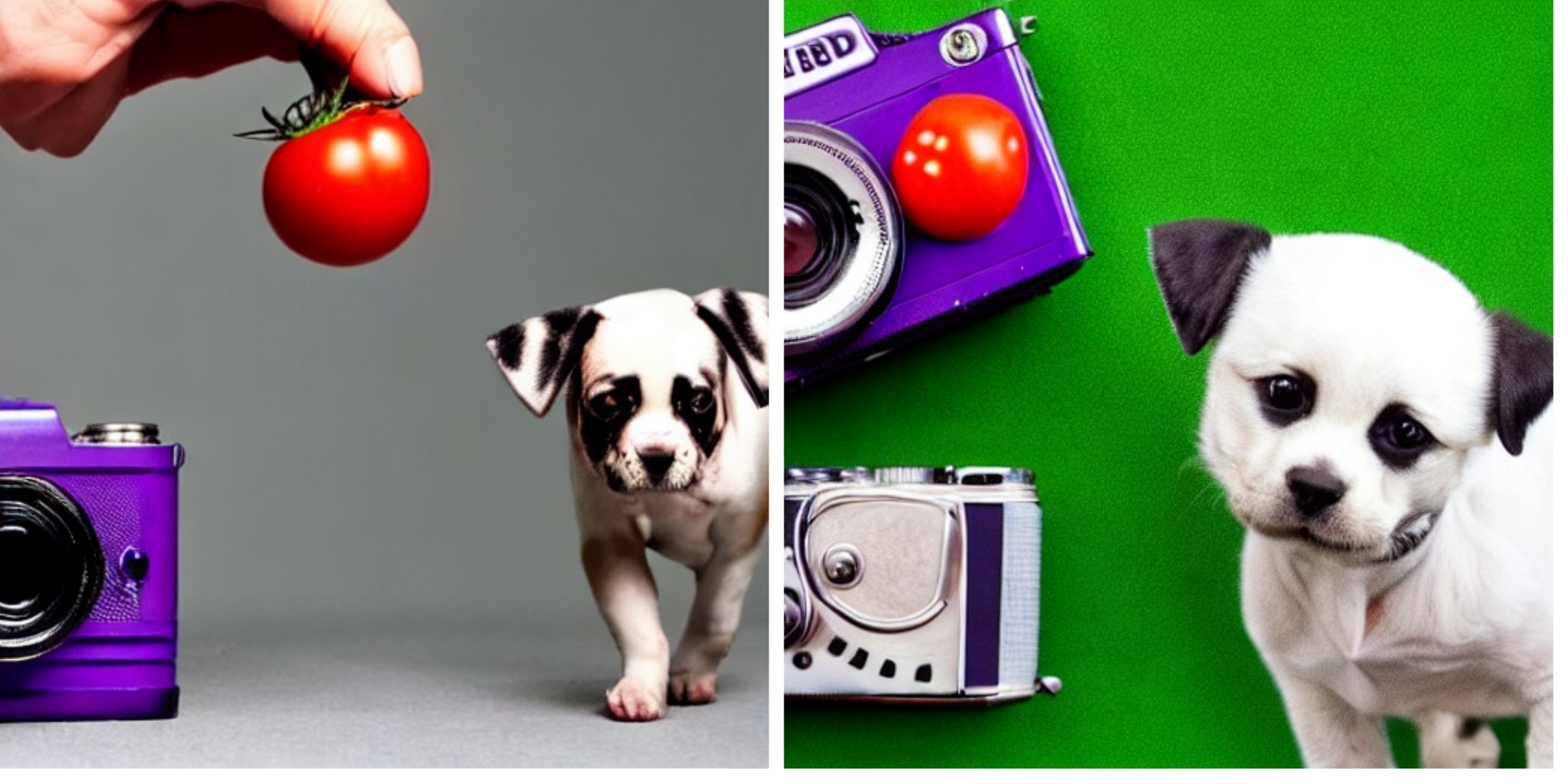 } & \includegraphics[width=0.25\textwidth]{  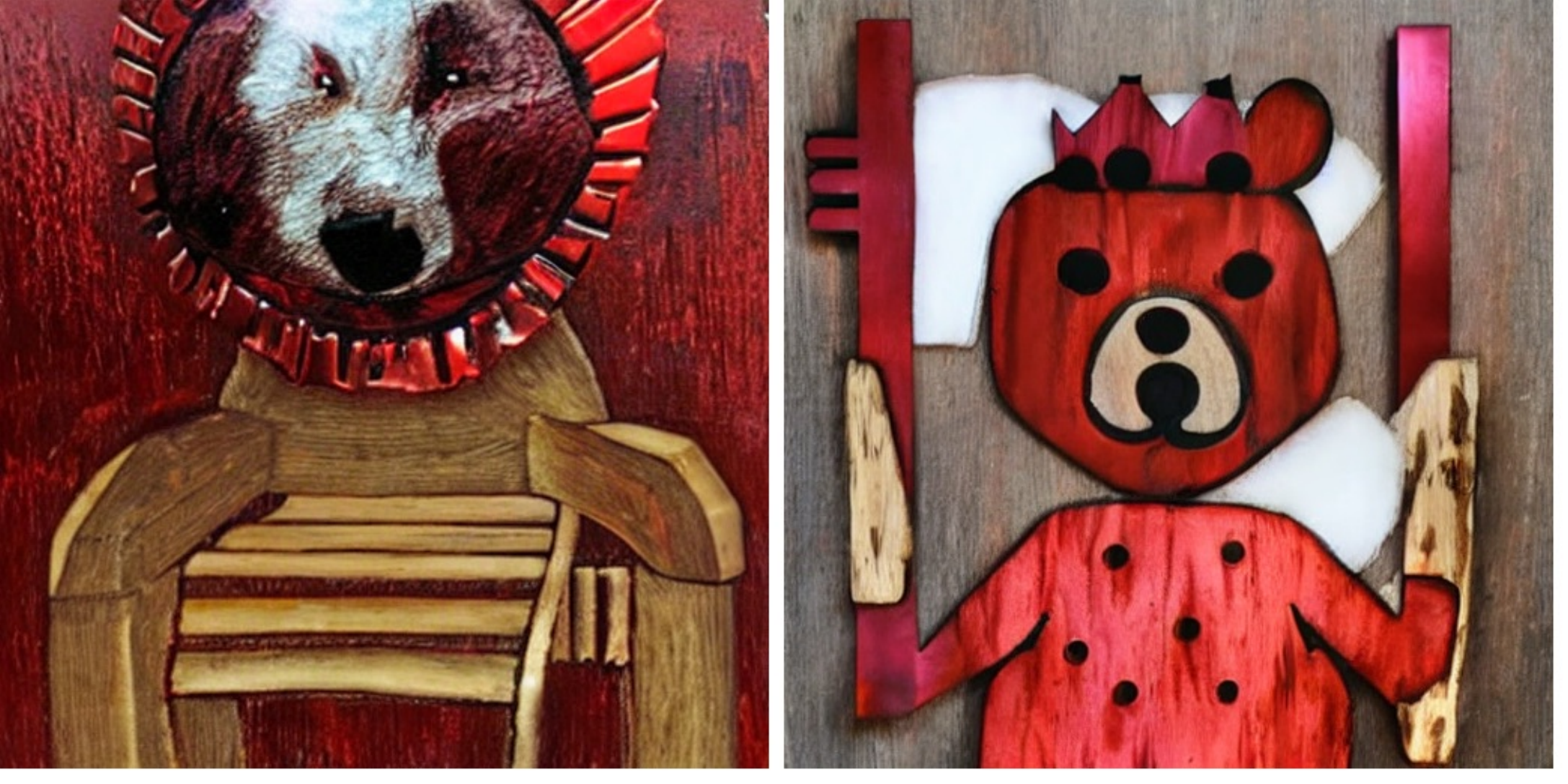} & \includegraphics[width=0.25\textwidth]{  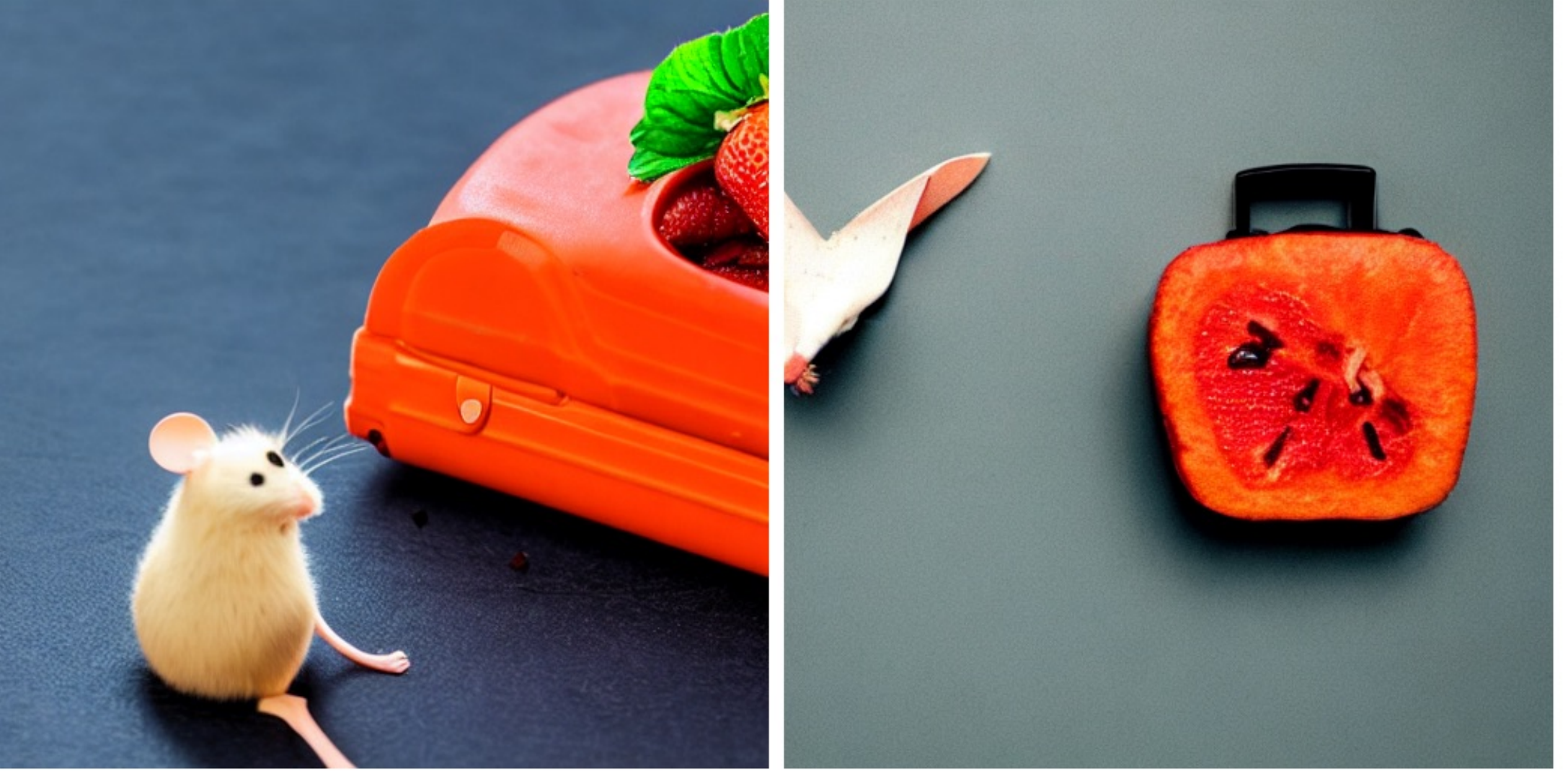 }  \\
    \raisebox{0.05\textwidth}{ \hspace{-5pt}\rotatebox{90}{SG}} & \includegraphics[width=0.25\textwidth]{  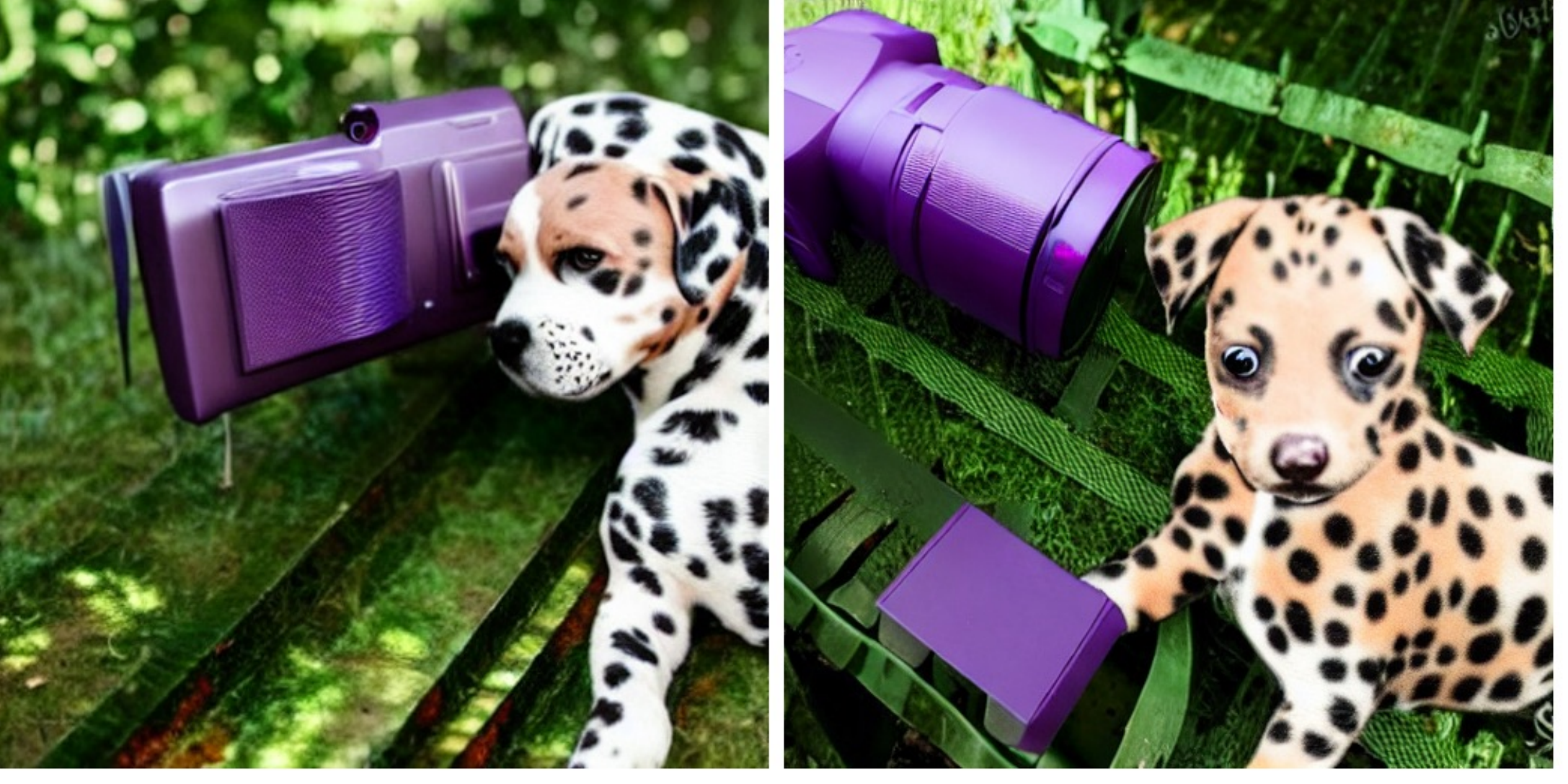 } & \includegraphics[width=0.25\textwidth]{  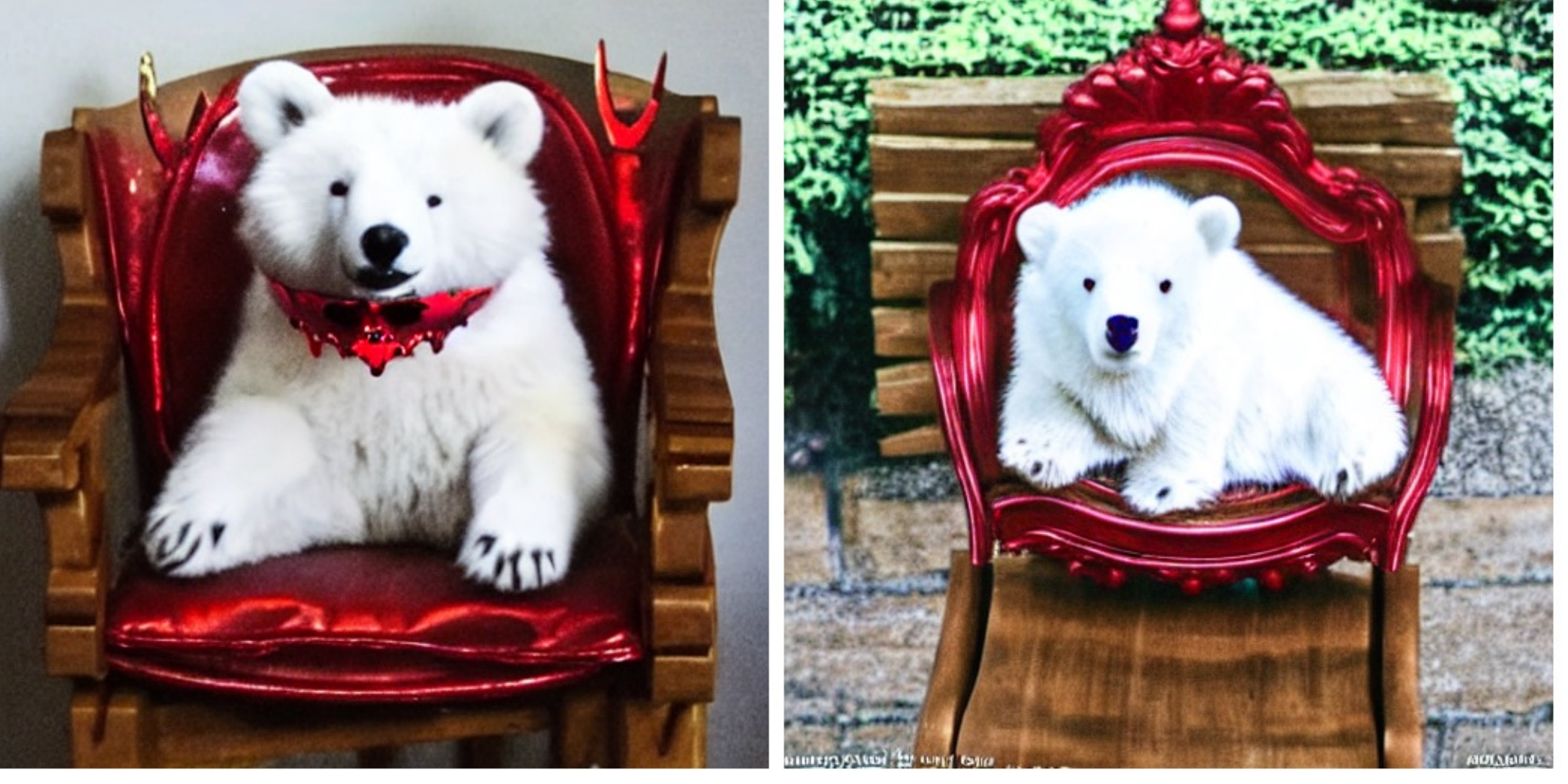 } & \includegraphics[width=0.25\textwidth]{  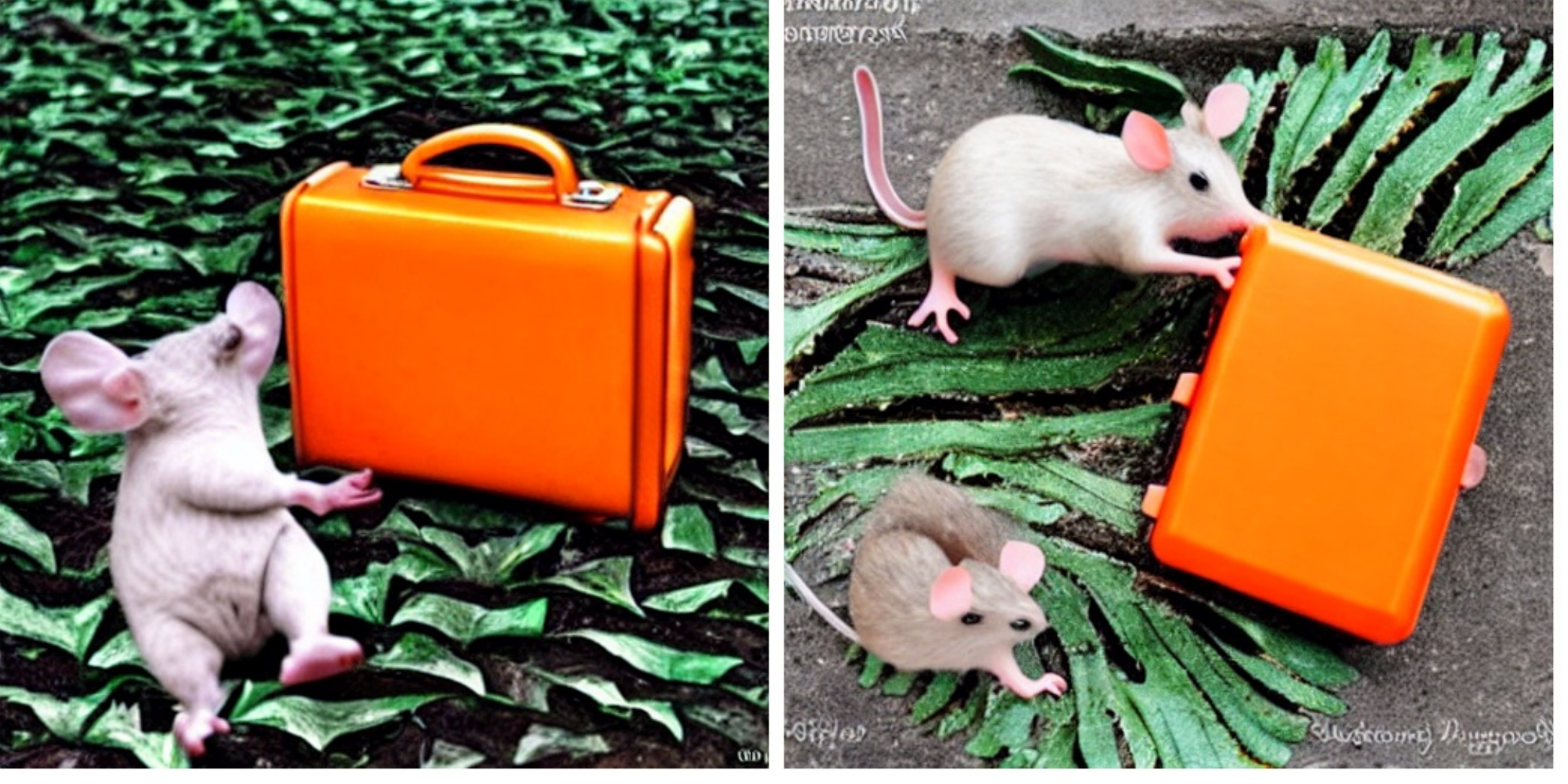 }  \\
     \raisebox{0.05\textwidth}{ \hspace{-5pt}\rotatebox{90}{Ours}} & \includegraphics[width=0.25\textwidth]{  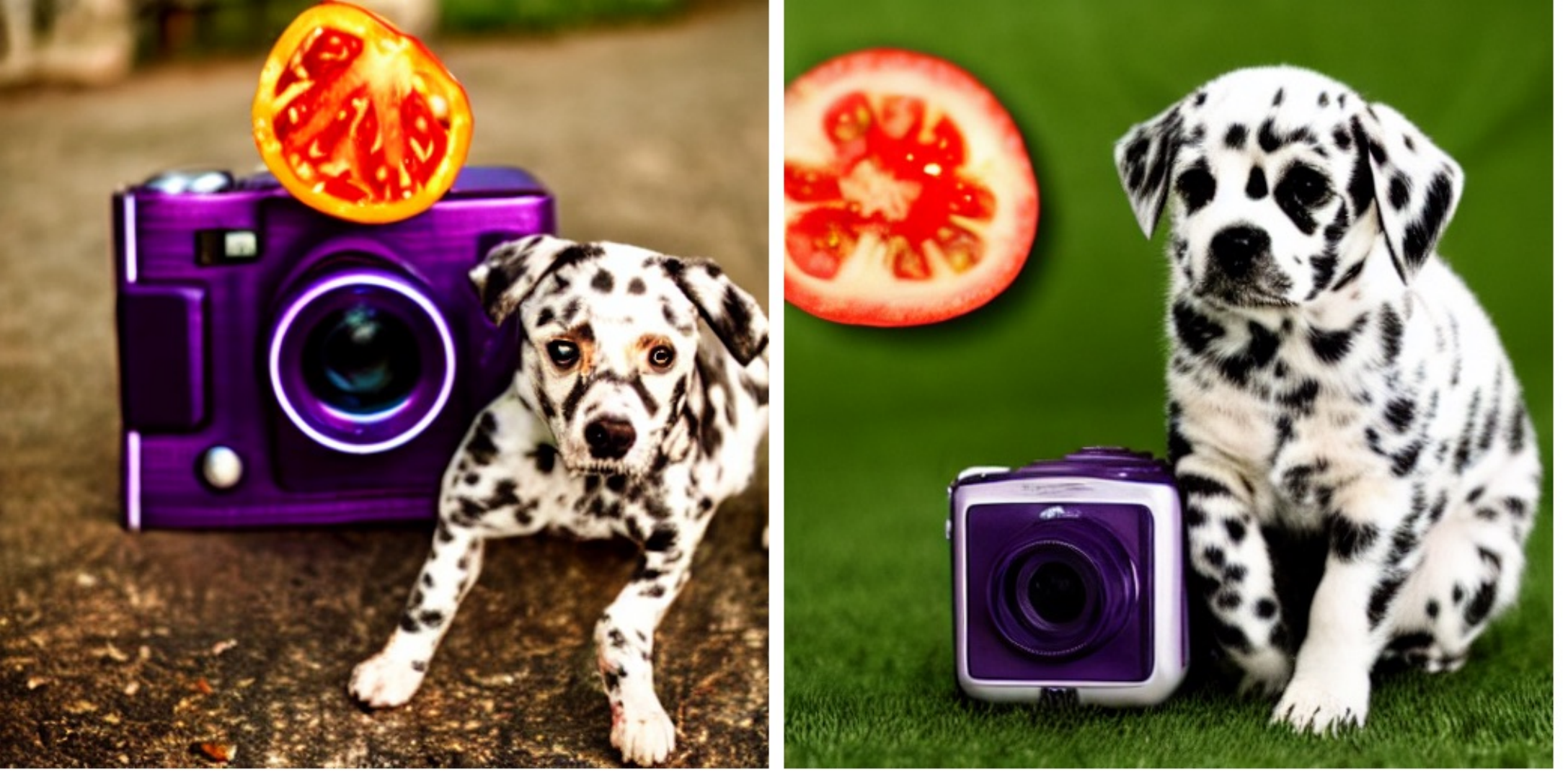 } & \includegraphics[width=0.25\textwidth]{  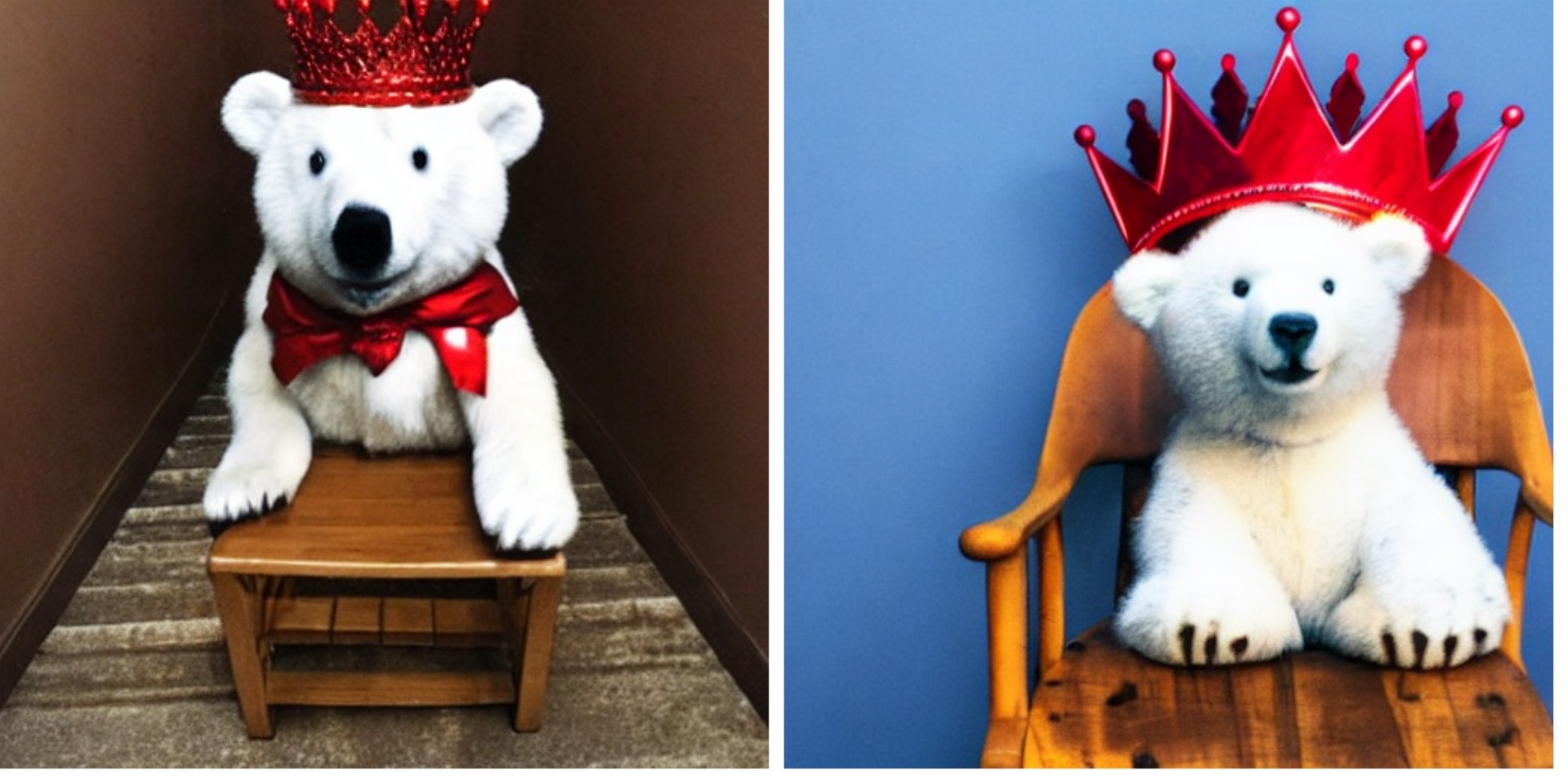 } & \includegraphics[width=0.25\textwidth]{  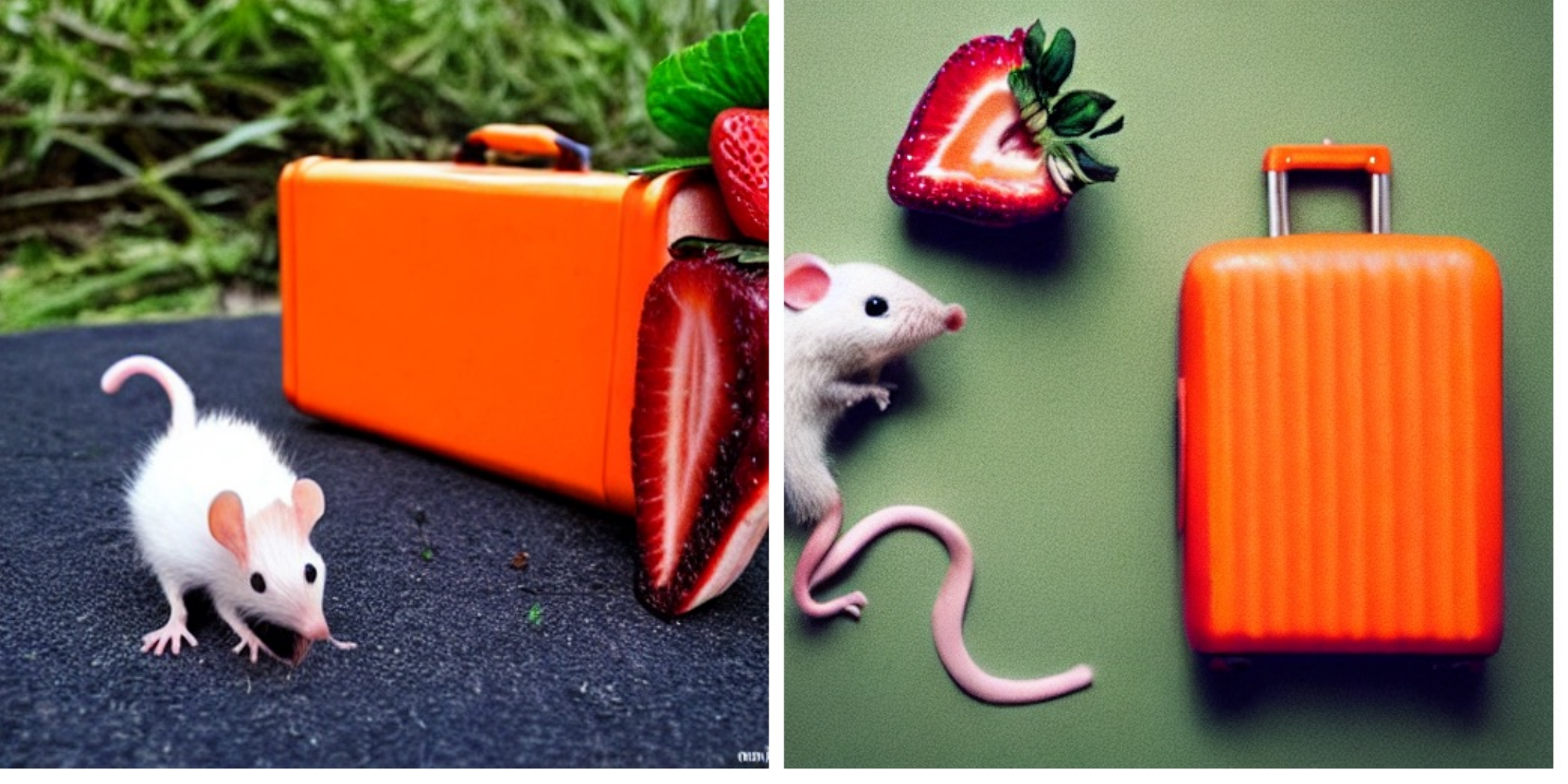 }   \\
       &  \multicolumn{1}{p{0.25\textwidth}}{\small \centering a \textcolor{Orchid}{purple} modern \underline{camera} and a spotted  baby \underline{dog} and a sliced \underline{tomato} } & 
     \multicolumn{1}{p{0.25\textwidth}}{\small \centering    a \textcolor{red}{red} metal \underline{crown} and a \textcolor{tgray}{white}   \underline{bear} and a wooden \underline{chair}}   &
        \multicolumn{1}{p{0.25\textwidth}}{\small \centering    an \textcolor{orange}{orange} \underline{suitcase} and a sliced  \underline{strawberry} and a baby \underline{mouse}} 
\end{tabular}
\caption{\textbf{Qualitative comparison on the DVMP dataset.} Each column shares the same random seed.}
\label{DVMP_image_comparison}
\end{figure*}

 \begin{figure*}[!h]
\centering
\renewcommand{\arraystretch}{0.8} 
\setlength{\tabcolsep}{4pt} 
\begin{tabular}{cccc}
    \raisebox{0.05\textwidth}{ \hspace{-5pt} \rotatebox{90}{SD} }& \includegraphics[width=0.25\textwidth]{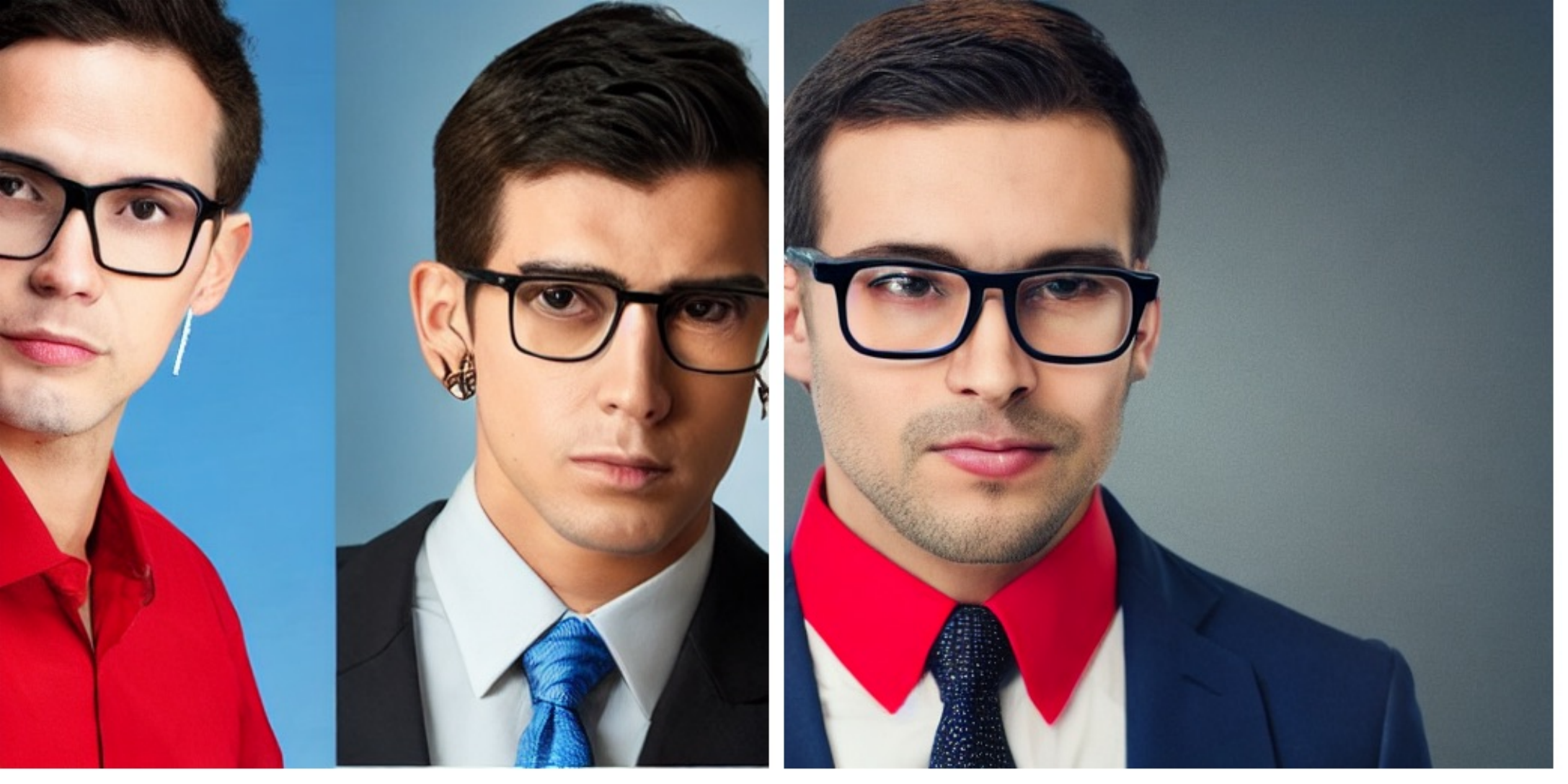} & \includegraphics[width=0.25\textwidth]{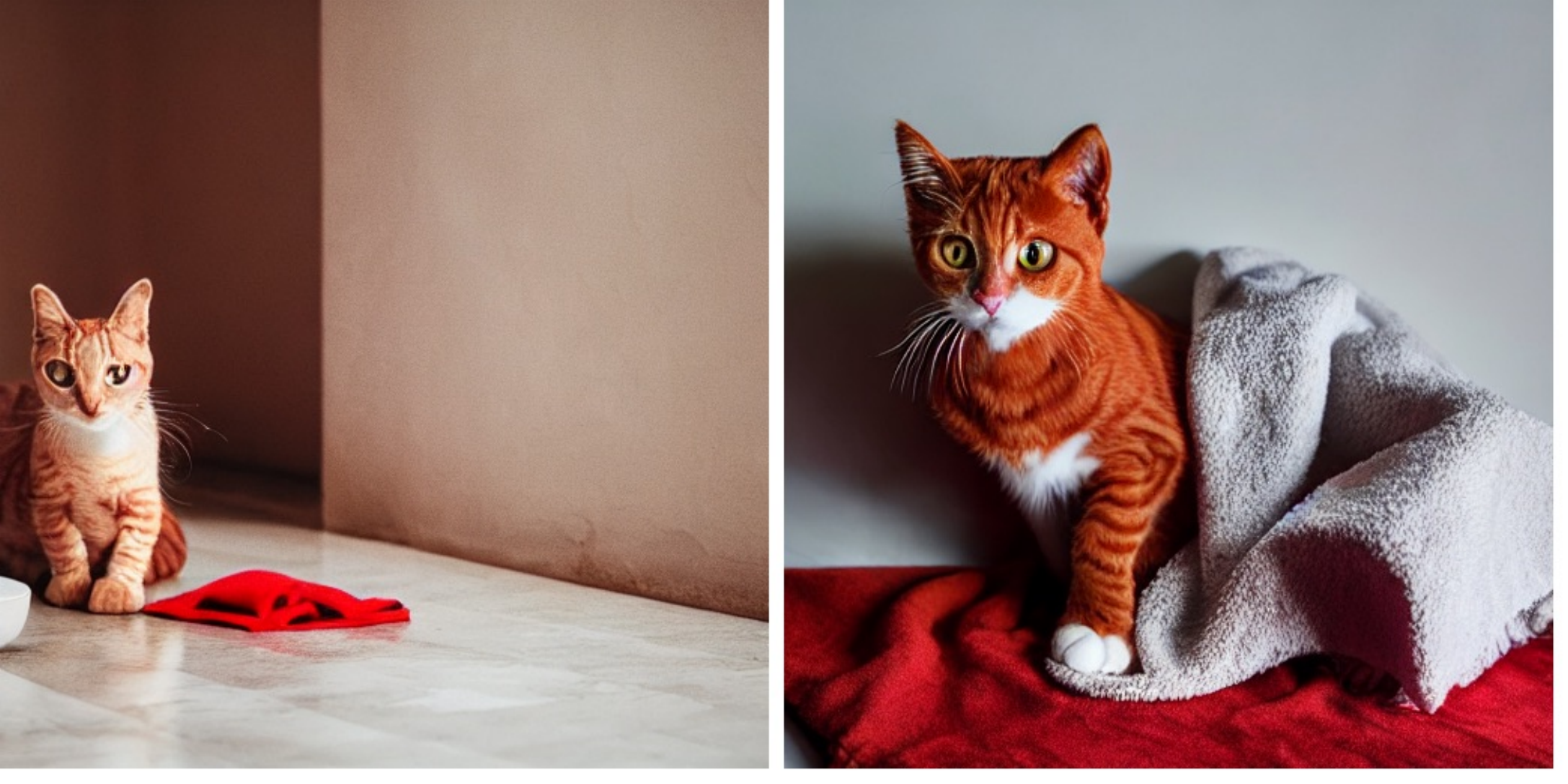} & \includegraphics[width=0.25\textwidth]{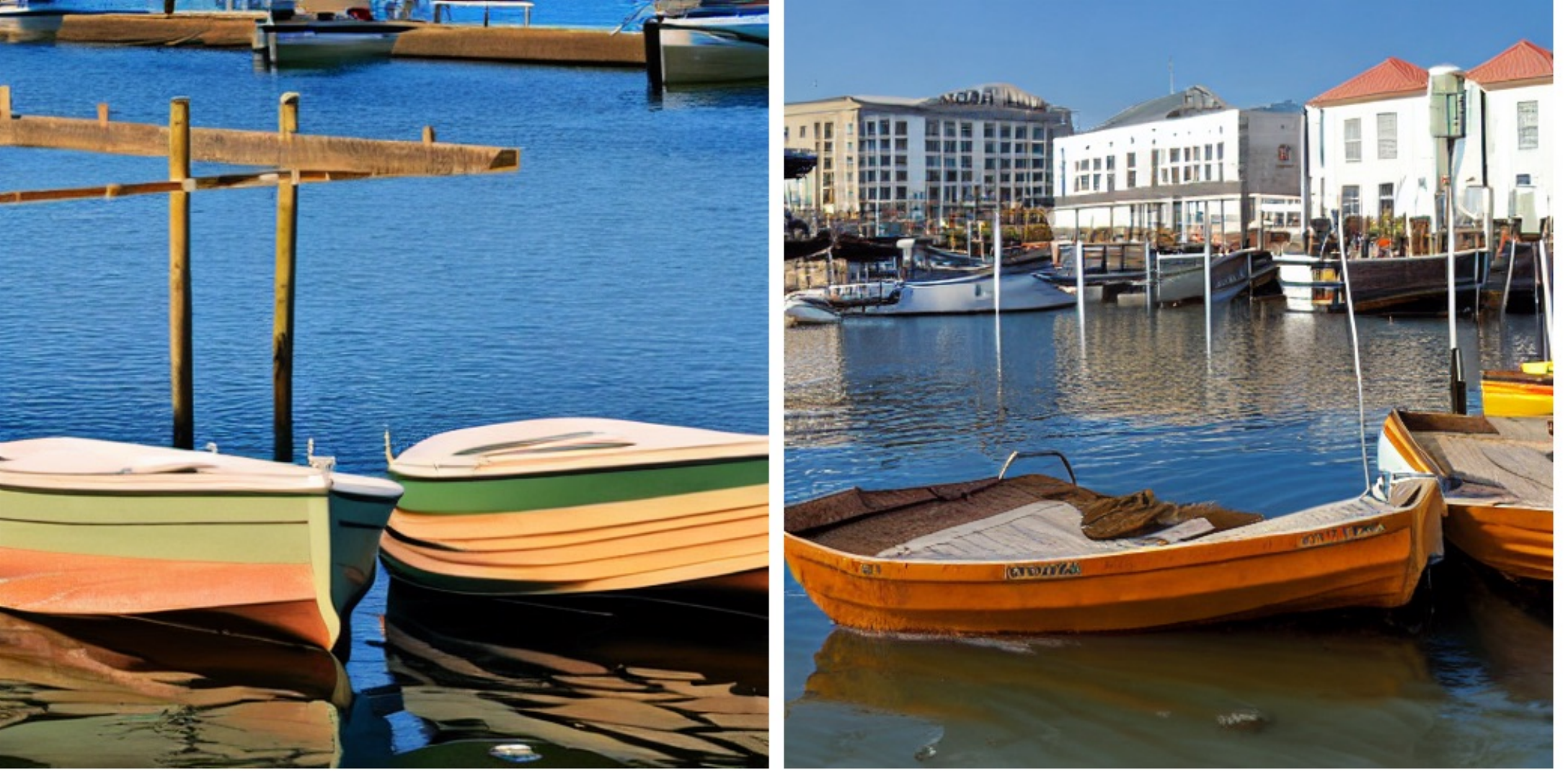}   \\
    \raisebox{0.05\textwidth}{ \hspace{-5pt}\rotatebox{90}{AnE}} & \includegraphics[width=0.25\textwidth]{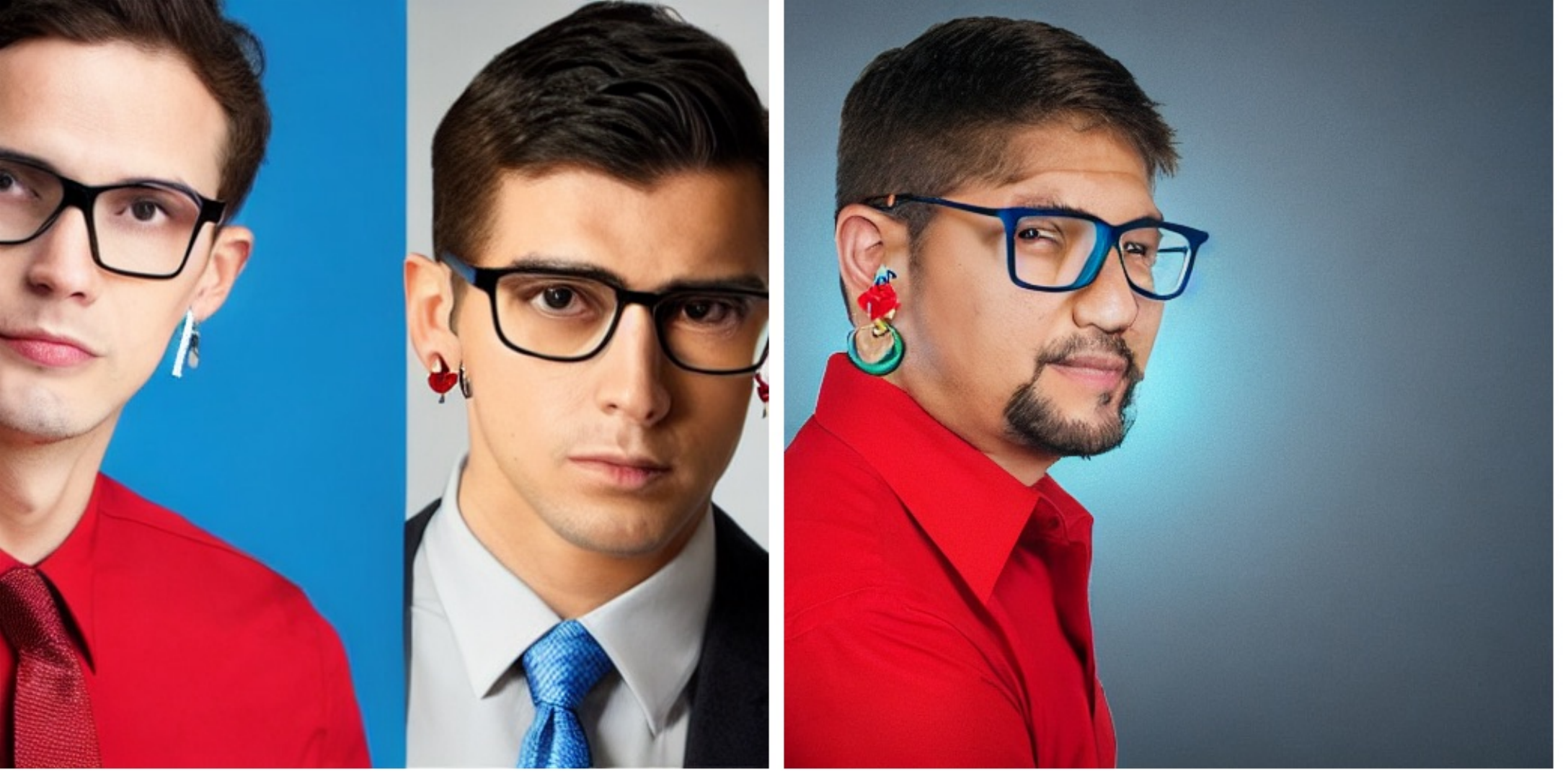} & \includegraphics[width=0.25\textwidth]{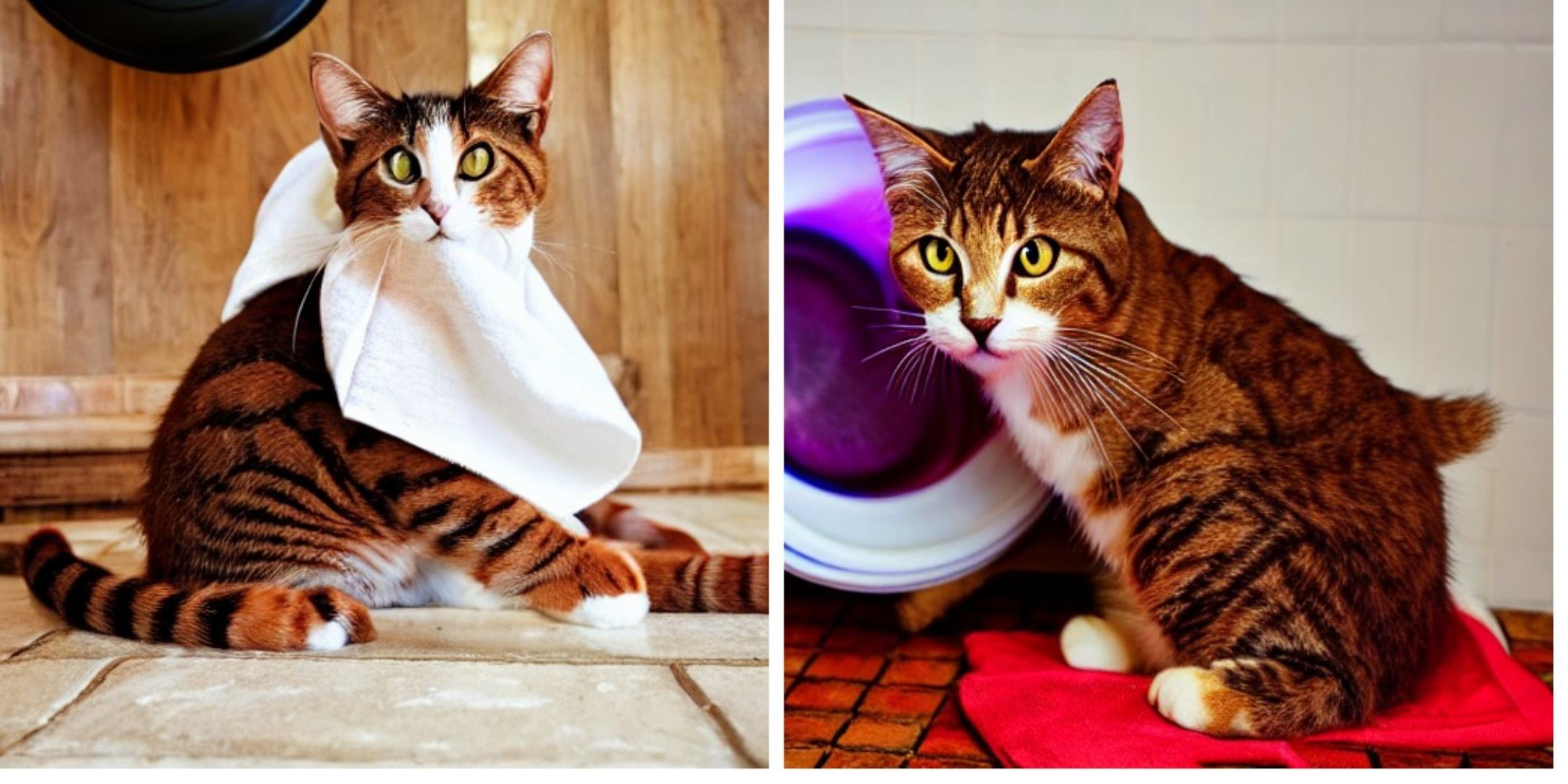} & \includegraphics[width=0.25\textwidth]{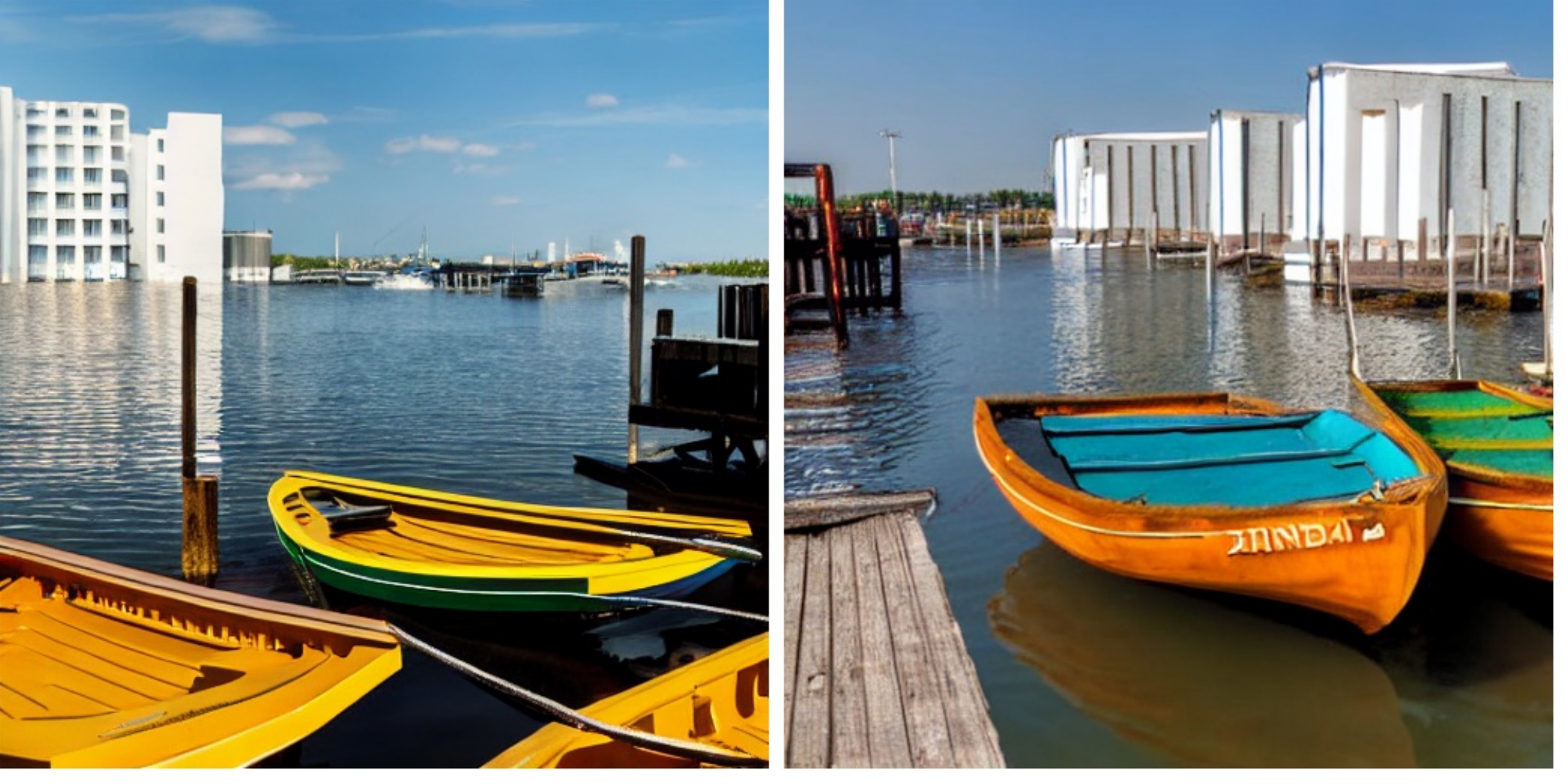}  \\
    \raisebox{0.05\textwidth}{ \hspace{-5pt}\rotatebox{90}{SG}} & \includegraphics[width=0.25\textwidth]{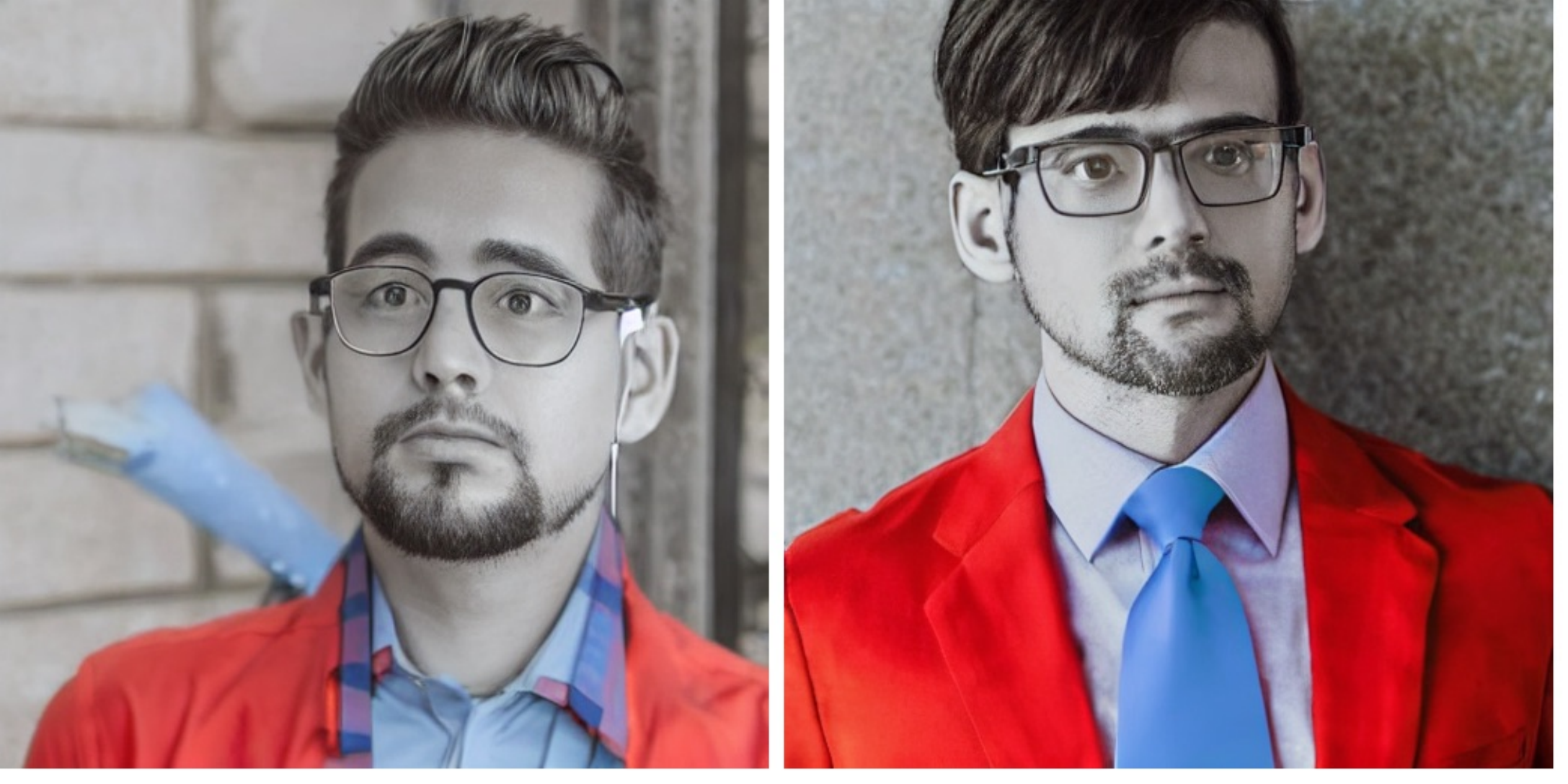} & \includegraphics[width=0.25\textwidth]{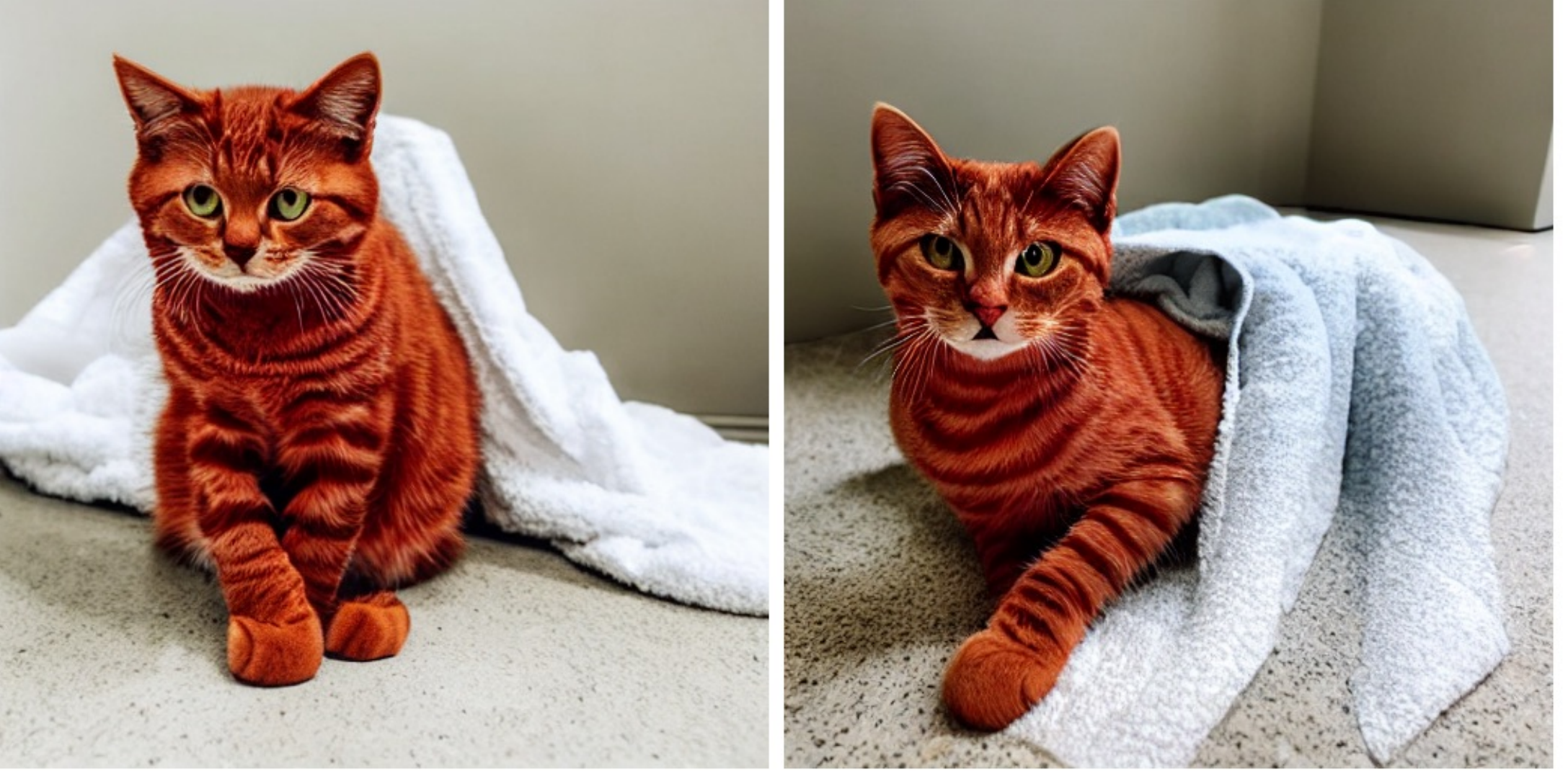} & \includegraphics[width=0.25\textwidth]{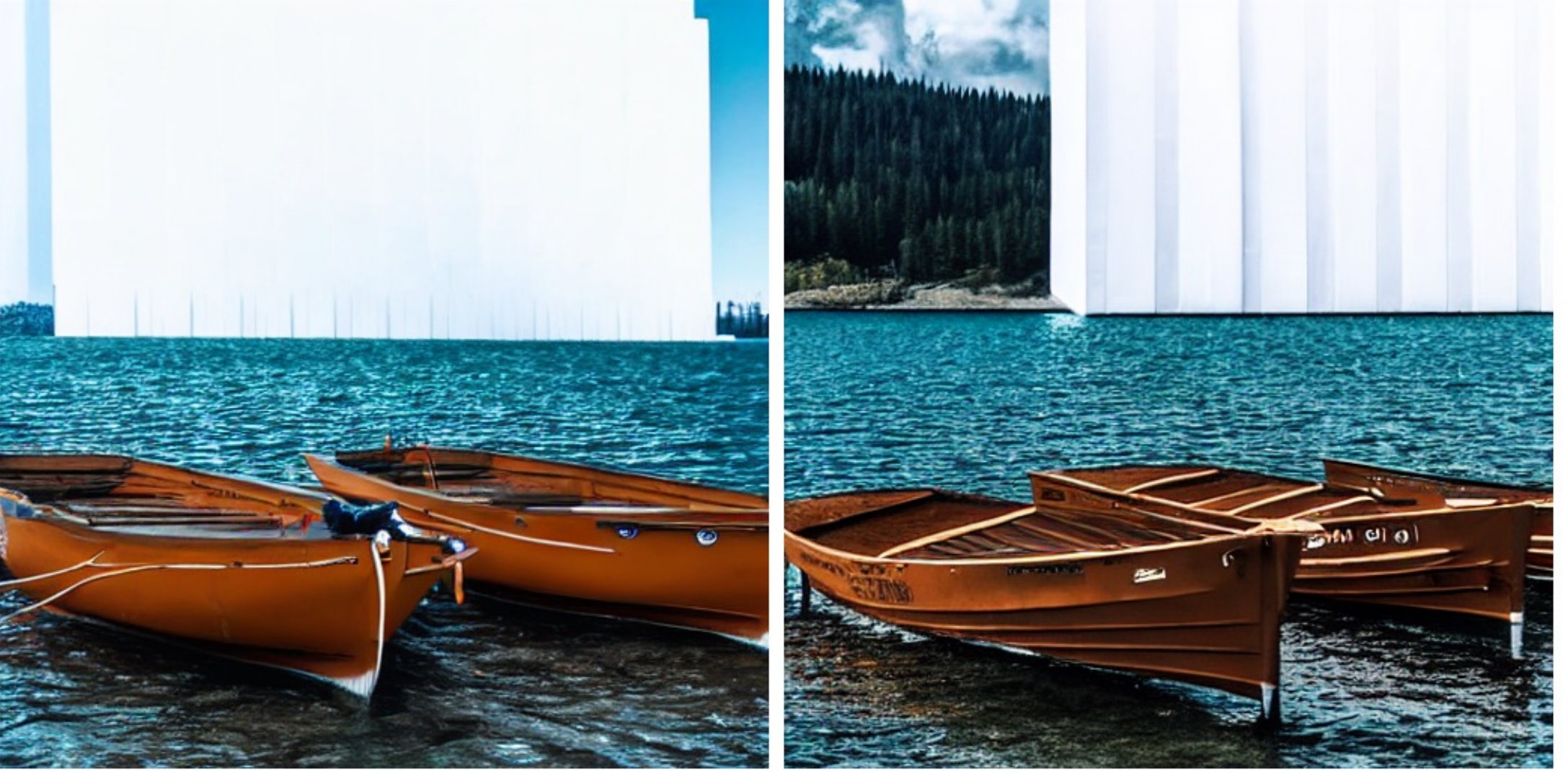}  \\
     \raisebox{0.05\textwidth}{ \hspace{-5pt}\rotatebox{90}{Ours}} & \includegraphics[width=0.25\textwidth]{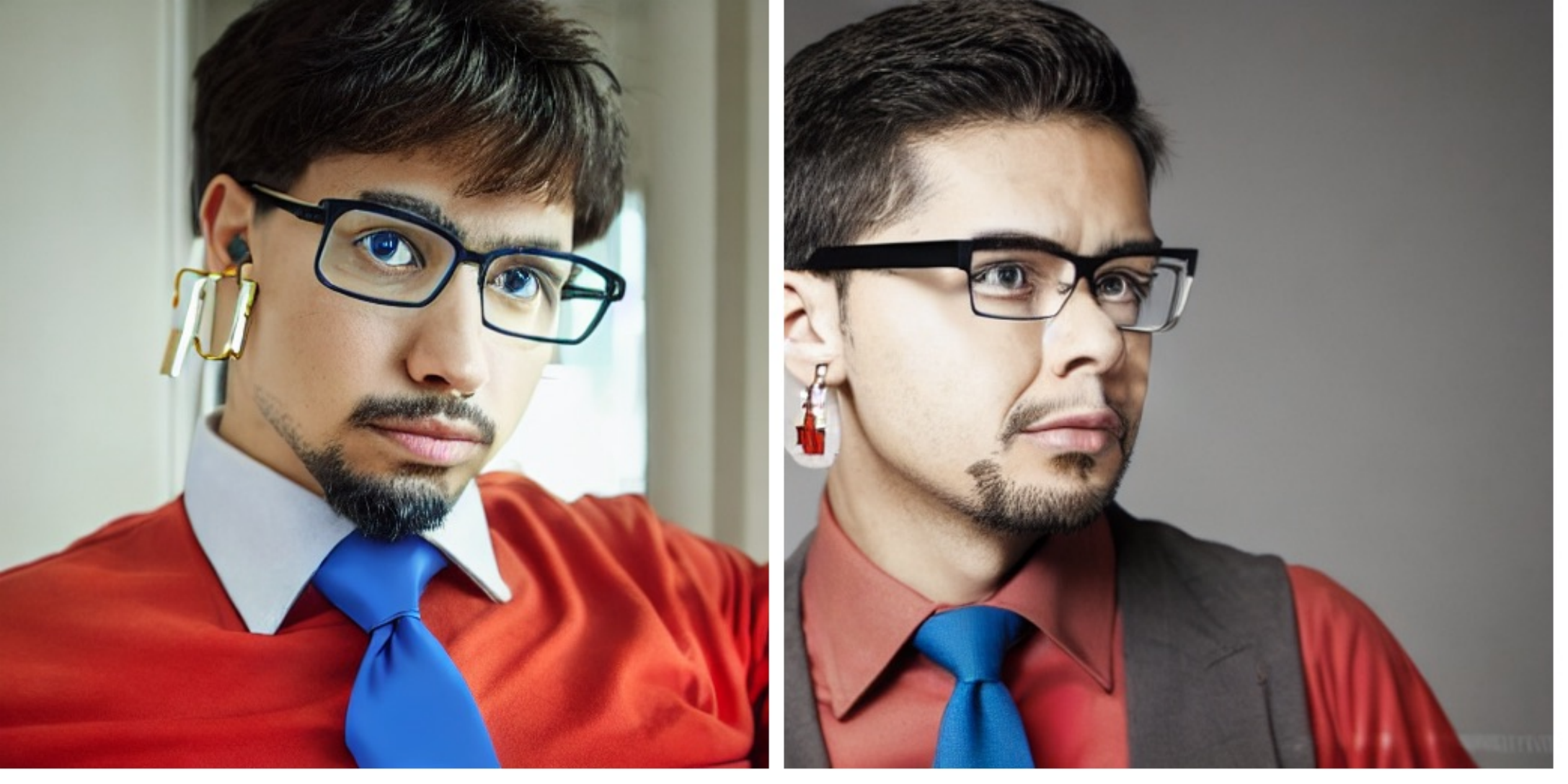} & \includegraphics[width=0.25\textwidth]{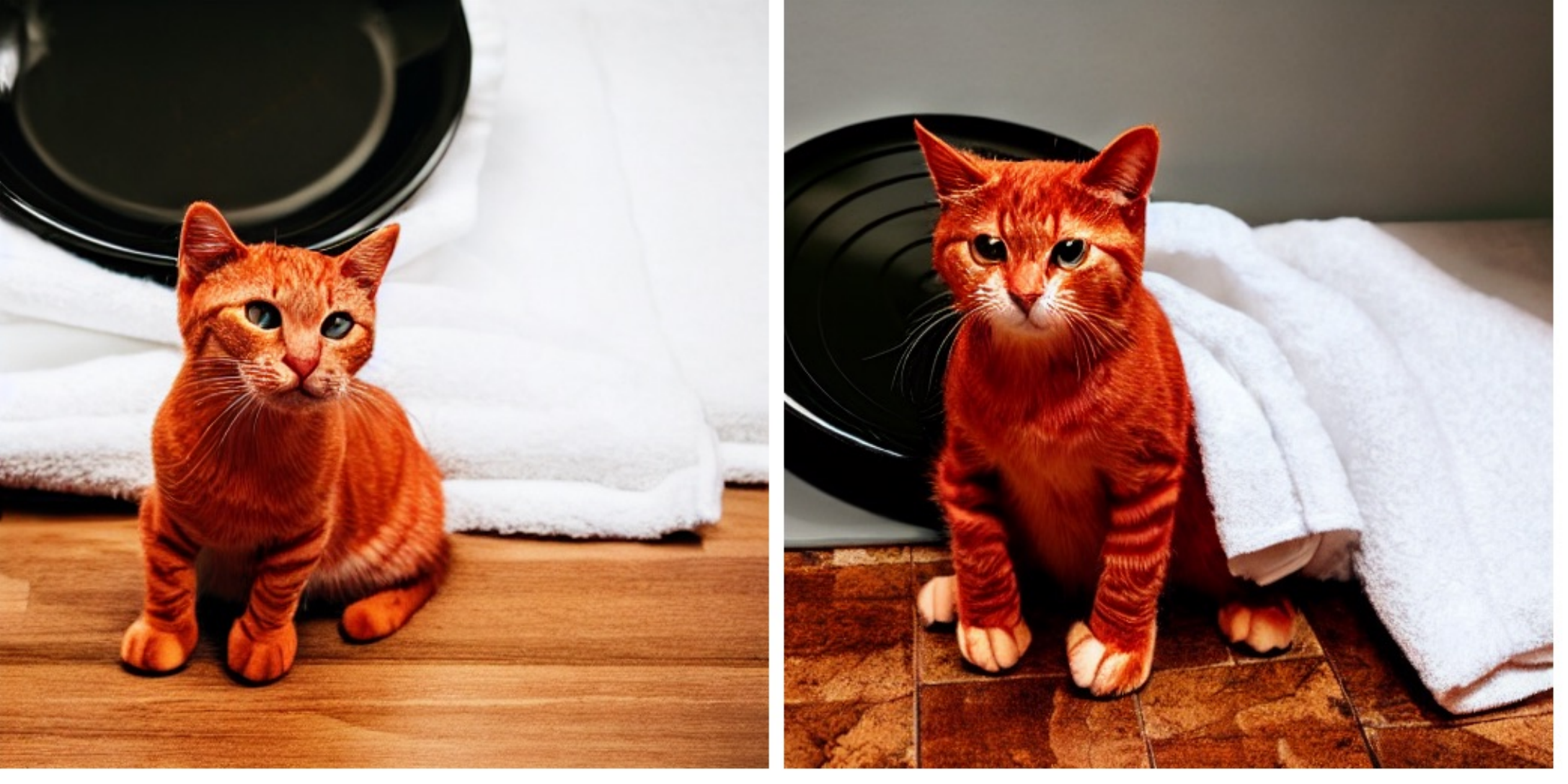} & \includegraphics[width=0.25\textwidth]{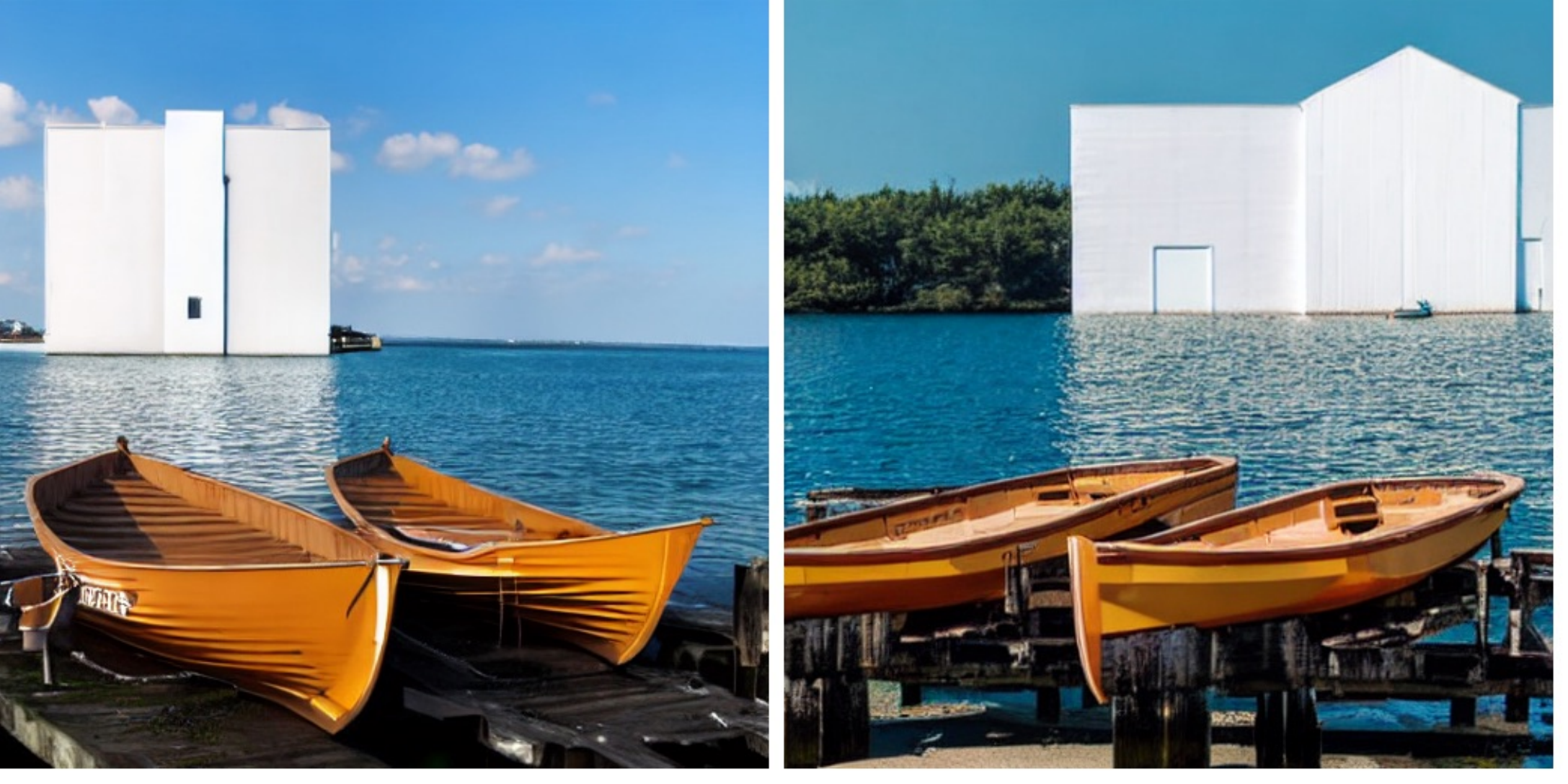}   \\
       & \multicolumn{1}{p{0.25\textwidth}}{\small \centering   A \underline{man} with \underline{glasses}, \underline{earrings},  and a \textcolor{red}{red} \underline{shirt} with \textcolor{blue}{blue} \underline{tie}.} & 
    \multicolumn{1}{p{0.25\textwidth}}{\small  \centering A \textcolor{red}{red} kitty \underline{cat} sitting on a \underline{floor} near a \underline{dish} and a \textcolor{tgray}{white} \underline{towel}.}  &
    \multicolumn{1}{p{0.25\textwidth}}{\small \centering   Two \textcolor{Tan}{tan} \underline{boats} on \underline{dock}  next to large \textcolor{tgray}{white} \underline{buildings}.}
\end{tabular}
\caption{\textbf{Qualitative comparison on the ABC-6K dataset.} Each column shares the same random seed.}
\label{abc_image_comparison}
\end{figure*}

In Figs. \ref{ane_image_comparison}-\ref{abc_image_comparison}, we identify recurrent failure modes in SG and AnE, attributable to the ineffectiveness of their objective design. AnE frequently struggles with incorrect attribute association, whereas SG often fails to generate multiple objects simultaneously. In contrast, our method attains high-quality semantic alignment with  deliberately designed optimization objective. It also exhibits more stable performance across different random seed selections.

 \paragraph{Object Omission} SG, due to its pair-centric approach, frequently omits objects, as evidenced by missing items like cars, apples, and crowns in Fig. \ref{ane_image_comparison}, tomatoes, crowns, strawberries in Fig. \ref{DVMP_image_comparison}, and earrings, ties, etc. in Fig. \ref{abc_image_comparison}.

\paragraph{Attribute Omission} Due to a lack of concern for attribute tokens, AnE fails to overcome the strong visual priors over objects. e.g.,  the green apple in Fig. \ref{ane_image_comparison}, the non-spotted dog in Fig. \ref{DVMP_image_comparison}, and the brown cat in Fig. \ref{abc_image_comparison}.

\paragraph{Attribute Leakage} In the case of SG, examples include purple on the wall, blue spilled on the plants, and purple on the suitcase  in Fig. \ref{ane_image_comparison}, illustrating how attributes emerge as leakage when the respective object is absent. Additional examples include the tomato's color spilling onto the dog and red metal leaking onto the chair in Fig. \ref{DVMP_image_comparison}, as well as blue color leaking and forming artifacts in Fig. \ref{abc_image_comparison}. AnE, with its sole focus on intensity, also suffers significantly from attribute leakage, evident in the purple backpack and blue suitcase in Fig. \ref{ane_image_comparison}, the red metal chair in Fig. \ref{DVMP_image_comparison}, and the blue glasses and earrings, and red towel in Fig. \ref{abc_image_comparison}.

We argue that addressing object neglect or attribute binding in isolation is insufficient, as these issues are intrinsically interconnected. Our method adeptly balances these two concerns with a chosen intensity weight 
$\lambda$, demonstrating its success in addressing the challenges above.

\subsection{Human Evaluation}
\label{sec:human_eval}
Recent work \cite{yuksekgonul2022and, chang2024skews} has found  that large Vision-and-Language Models (VLMs)  \cite{radford2021learning, singh2022flava, li2022blip, zeng2021multi}  demonstrate a significant lack of compositional understanding, failing to reflect human preferences accurately. Given this, we conducted human evaluations across all three datasets to rigorously assess our model's performance.

Raters were enlisted online, with the requirement that each participant possessed an educational level of a bachelor's degree or higher. 
In the process of evaluation, they were presented with 2-way multiple choice problems consisting of a text prompt and two images generated by our method and one of four baselines, including SD, AnE, SG, and our method with 
$\lambda=0$. For each dataset, 100 prompts were randomly sampled for evaluation. The effectiveness of prompt-image alignment was assessed by asking raters, "Which image better matches the given description?". More details are provided in supplementary material.

  \begin{wrapfigure}{l}{0.48\textwidth} 
  \centering
  \includegraphics[width=0.48\textwidth]{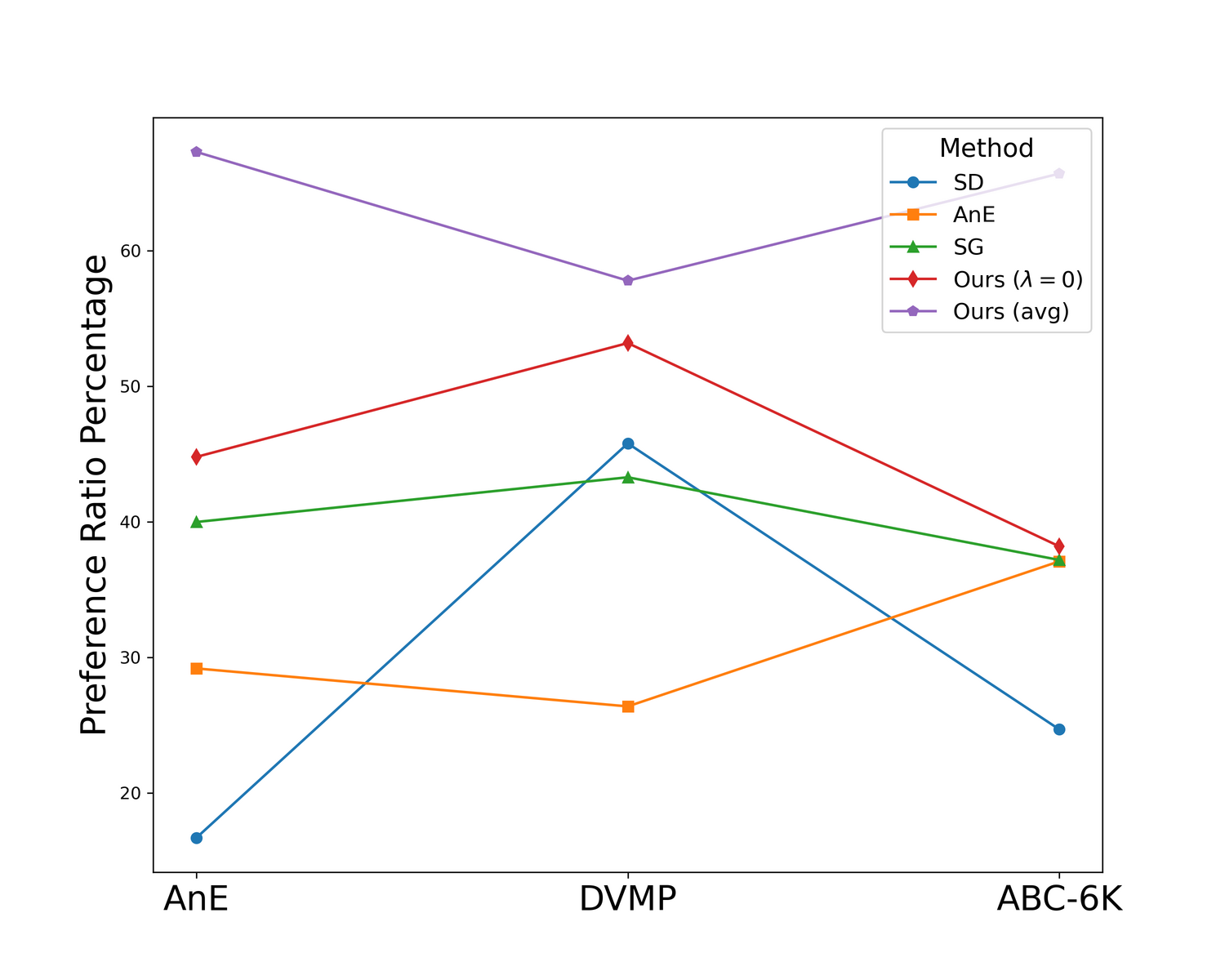 }
\caption{\textbf{Preference ratio percentage on text-image alignment by human evaluation}. Ours(avg) represents the average preference ratio of our method compared with the other four methods.}
  \label{fig:user}
\end{wrapfigure}

The human evaluation results are shown in Fig. \ref{fig:user}. 
 We observe that: 1) our method consistently surpasses SD, AnE, and SG aligned with quantitative results in Tab. \ref{ane_table} and  Fig. \ref{abcDVMP}; 2) our method shows a more pronounced advantage over other methods on the natural-language ABC-6K dataset. We argue that our object-centric objective,   in harmony with the object-oriented patterns prevalent in naturally occurring prompts, exhibits superior efficacy in handling complex real-world, natural-language-based prompts.

 \begin{table*}[b]
\centering
\caption{\textbf{Ablation results  on repulsive term. } Both Ours and Ours($\lambda=0$) benefit from the repulsive term as defined in Eqn. (\ref{loss_b}).}
\label{tab:repulsive}
\begin{tabular}{lcccccccccc}
\toprule
& & \multicolumn{3}{c}{\scriptsize Animal-Animal} & \multicolumn{3}{c}{\scriptsize Animal-Object} & \multicolumn{3}{c}{\scriptsize Object-Object} \\
\cmidrule(lr){3-5} \cmidrule(lr){6-8} \cmidrule(lr){9-11}
\scriptsize Method & \scriptsize Repul.  & \tiny Full Sim.  & \tiny Min. Sim.  & \tiny T-C Sim.  &  \tiny Full  Sim.  &  \tiny Min.  Sim.  &  \tiny T-C Sim.  &  \tiny Full  Sim.  &  \tiny Min. Sim.  &  \tiny T-C Sim.  \\
\midrule 

\scriptsize Ours($\lambda = 0)$ & \xmark& 0.311&0.213&0.767   &   0.343&0.246&0.794 &  0.334&0.237&0.765\\ 
\rowcolor{gray}
   & \cmark& \textbf{0.340} & \underline{0.255} & \underline{0.814} & \underline{0.362} & \textbf{0.271} & \textbf{0.851} & \underline{0.360} & \underline{0.270} &  \underline{0.823}\\ 
\midrule
\scriptsize Ours & \xmark & 0.338&0.250&0.810  &  0.360&0.267&0.841&   0.359&0.269&0.819\\
\rowcolor{gray}
  & \cmark & \textbf{0.340} & \textbf{0.256} & \textbf{0.817} &  \textbf{0.362} & \underline{0.270} &\textbf{0.851} & \textbf{0.366} & \textbf{0.274 }& \textbf{0.836}\\
\bottomrule
\end{tabular} \label{repul}
\end{table*}

\subsection{Ablation Study} \label{sec:ablation}
 \begin{wrapfigure}{l}{0.48\textwidth}
     \centering
       \renewcommand{\arraystretch}{0.7} 
\setlength{\tabcolsep}{0.5pt} 
    \begin{tabular}{cccc}
       \raisebox{0.05\textwidth}{ \hspace{-5pt} \rotatebox{90}{$\lambda=0.0$} }& \includegraphics[width= 0.14\textwidth]{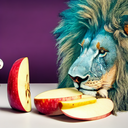} & \includegraphics[width= 0.14\textwidth]{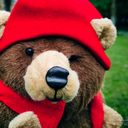} & \includegraphics[width= 0.14\textwidth]{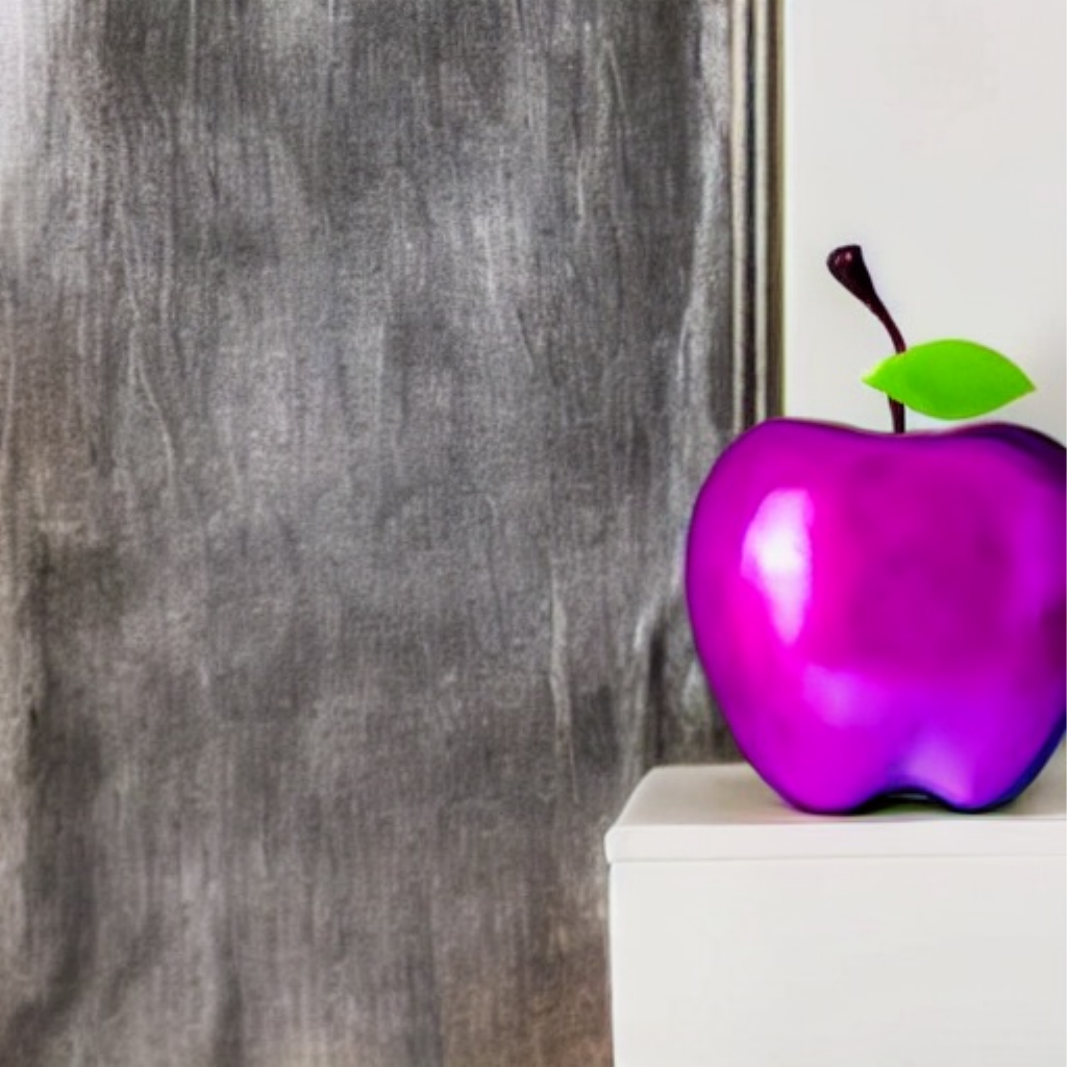}   \\
    \raisebox{0.05\textwidth}{ \hspace{-5pt}\rotatebox{90}{$\lambda=0.5$}} & \includegraphics[width= 0.14\textwidth]{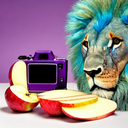} & \includegraphics[width= 0.14\textwidth]{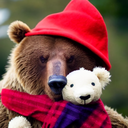} & \includegraphics[width= 0.14\textwidth]{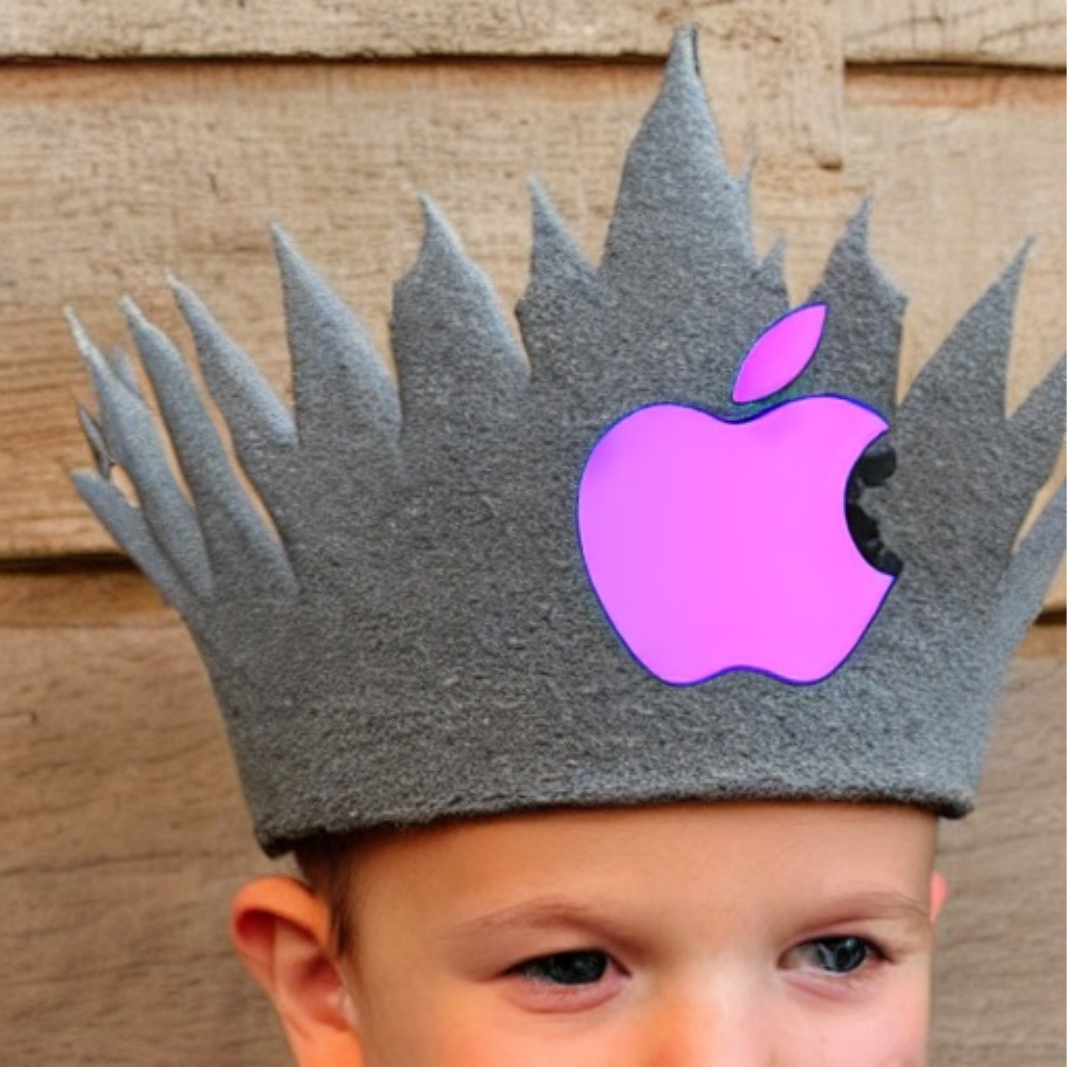}  \\
    \raisebox{0.05\textwidth}{ \hspace{-5pt}\rotatebox{90}{$\lambda=1.0$}} & \includegraphics[width= 0.14\textwidth]{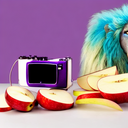} & \includegraphics[width= 0.14\textwidth]{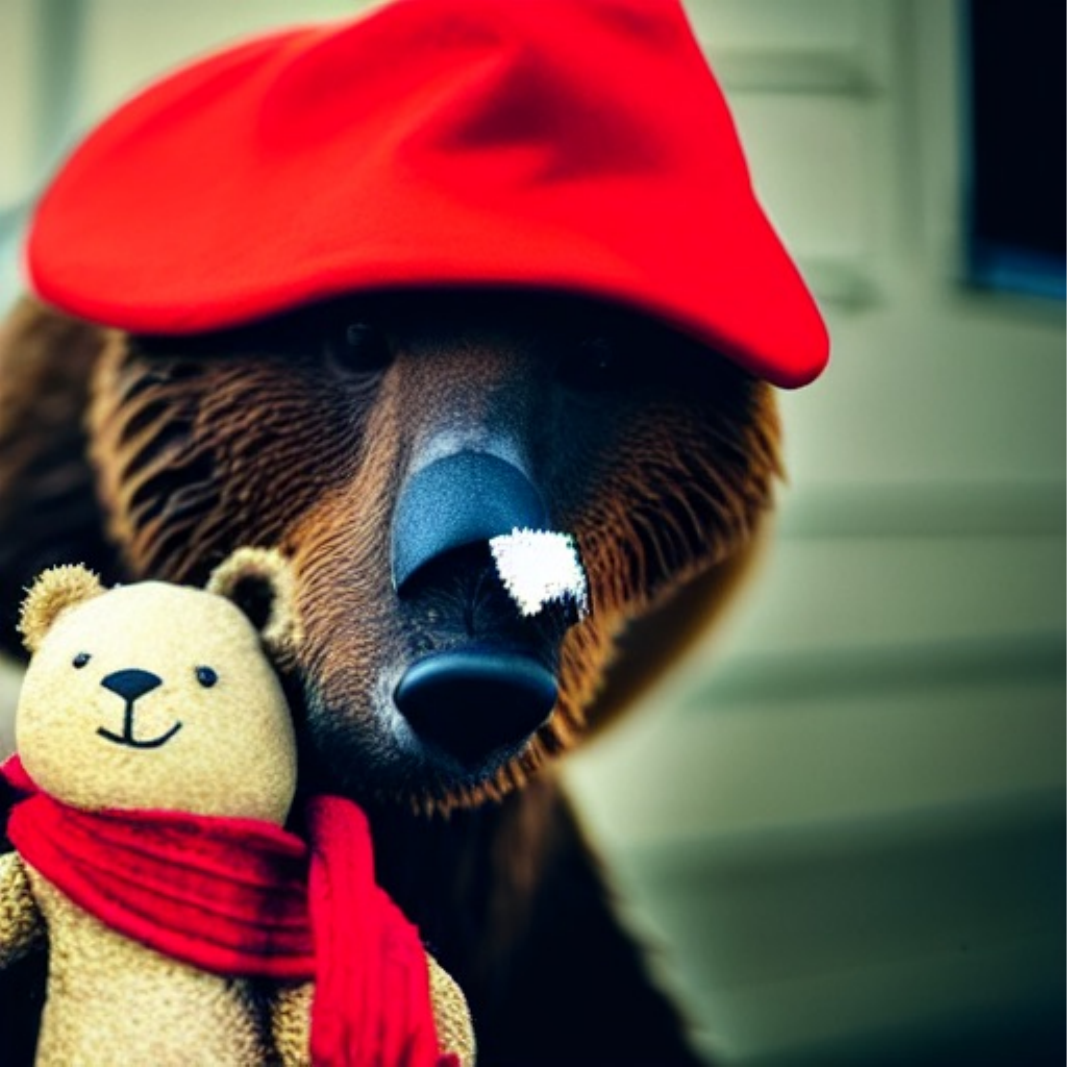} & \includegraphics[width= 0.14\textwidth]{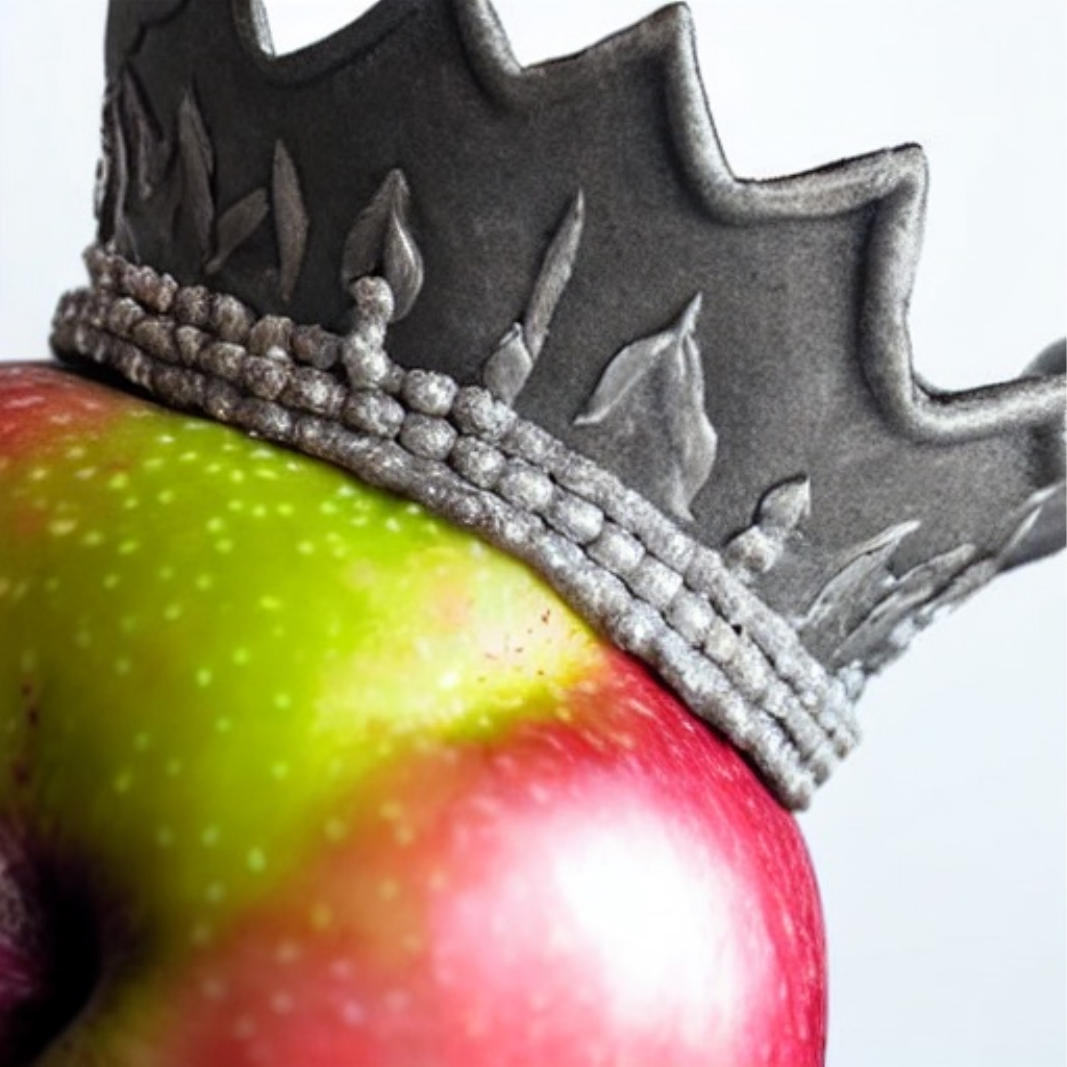}  \\
   & (a) & (b) & (c)
    \end{tabular}
     \caption{\textbf{Ablation demonstration for intensity weight $\lambda$.} (a) a sliced \underline{apple} and a \textcolor{Orchid}{purple} \underline{camera} and a \textcolor{teal}{teal} \underline{lion}; (b) a \textcolor{Brown}{brown} \underline{bear} with \textcolor{red}{red} \underline{hat} and \underline{scarf} and a small stuffed \underline{bear}; (c) a \textcolor{tgray}{gray} \underline{crown} and a \textcolor{Orchid}{purple} \underline{apple}. We have selected one prompt from each dataset to showcase the stability of our method.  Each column shares the same random seed.}
     \label{Ablation}  
 \end{wrapfigure}

\paragraph{Repulsive Term} Tab. \ref{repul}  presents the results of  ours(${\lambda=0}$)  w/o and w/ the repulsive term in rows 1 and 2, and similarly,   ours  w/o and w/ this term in rows 3 and 4,  under the same settings as Tab. \ref{ane_table}. Row 2/4 demonstrates a significant performance increase than Row 1/3 due to the repulsive term, validating the effectiveness of negative sampling approximation.

\paragraph{Intensity Weight} We demonstrate the impact of different choices for the intensity weight $\lambda$, which plays a role in enhancing the intensity level. In Fig. \ref{Ablation}, we present some representative examples where the model needs to generate multiple objects with certain modifiers. When $\lambda=0.5$, the generation is balanced. However, when $\lambda=0.0$, all images more or less suffer from object neglect. Conversely, when $\lambda=1.0$, artifacts are likely to appear and attribute binding becomes less effective.

\begin{wraptable}{r}{0.48\textwidth}
    \centering
        \caption{\textbf{Ablations on object-conditioned perspective and energy function choice.}}
    \begin{tabular}{l|ccc}
    \toprule
     &\tiny  A-A &\tiny  A-O &\tiny  O-O \\
     \midrule
  \tiny      ours($\lambda=0$) &  \tiny  0.814&  \tiny 0.851 & \tiny  0.823 \\
   \tiny    - obj cond.  & \tiny   0.814 \textcolor{Red}{(-0.0\%)}   &\tiny   0.832   \textcolor{Red}{(-2.2\%)}  &\tiny   0.813  \textcolor{Red}{(-1.2\%)}  \\
   \tiny    - cos sim. &\tiny   0.812 \textcolor{Red}{(-0.2\%)}   &\tiny    0.846 \textcolor{Red}{(-0.6\%)}  & \tiny   0.817 \textcolor{Red}{(-0.7\%)}  \\
      \midrule
  \tiny     SG  &\tiny  0.767 \textcolor{Red}{(-5.8\%)} &\tiny  0.830  \textcolor{Red}{(-2.5\%)}& \tiny  0.811 \textcolor{Red}{(-1.5\%)} \\
     \bottomrule
    \end{tabular}
    \label{tab:component}
\end{wraptable}

\paragraph{Object-Conditioned Perspective} Besides being able to handle more flexible prompts where no attribute-object pairs exist, our method is object-centric and thus, in the repulsive term, we only calculate the energy of objects and non-modifiers. The way SG handles it is to treat objects and modifiers equivalently when faced with non-modifiers. In the row `- obj cond.' of Tab. \ref{tab:component}, we replace the energy of object and non-modifier $f(A_s,A_l)$ with the average energy of $f(A_s, A_l)$ and $f(A_s, A_m)$, where $s, l, m$ represent object, non-modifier, and modifier, respectively. Note that as no modifiers exist in A-A, the results remain the same. In the other datasets,  our object-conditioned perspective plays a vital role in the success of our method as the performance significantly decreases. 

\paragraph{Energy function Choice} To calculate KL div., SG assumes attention maps follow a multinomial distribution. Yet,  cosine similarity does not pose any assumption on the distribution and achieves superior performance. In the row `- cos sim.' of Tab. \ref{tab:component}, we replace cosine similarity with the average KL div..

\subsection{Augmented Attribute Editing}

\begin{wrapfigure}{l}{0.48\textwidth}
    \centering
    \renewcommand{\arraystretch}{1} 
   \setlength{\tabcolsep}{1pt} 
    \begin{tabular}{cccc}  
        \multicolumn{2}{c}{\scriptsize Original   $\longrightarrow$   \scriptsize Edited    }  &    \multicolumn{2}{c}{\scriptsize Original  $\longrightarrow$ \scriptsize Edited  }     \\
         \includegraphics[width = 0.11\textwidth]{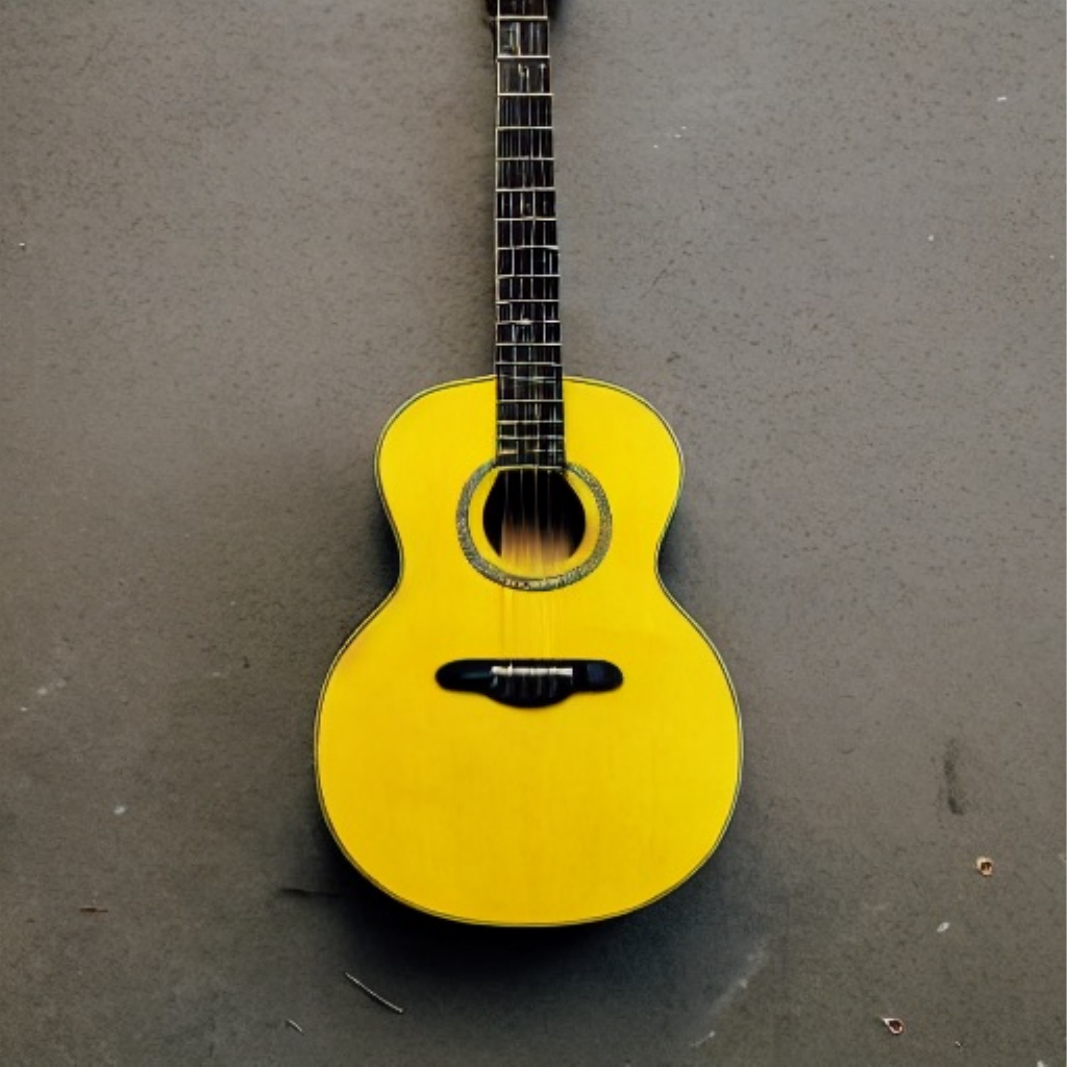} & \includegraphics[width = 0.11\textwidth]{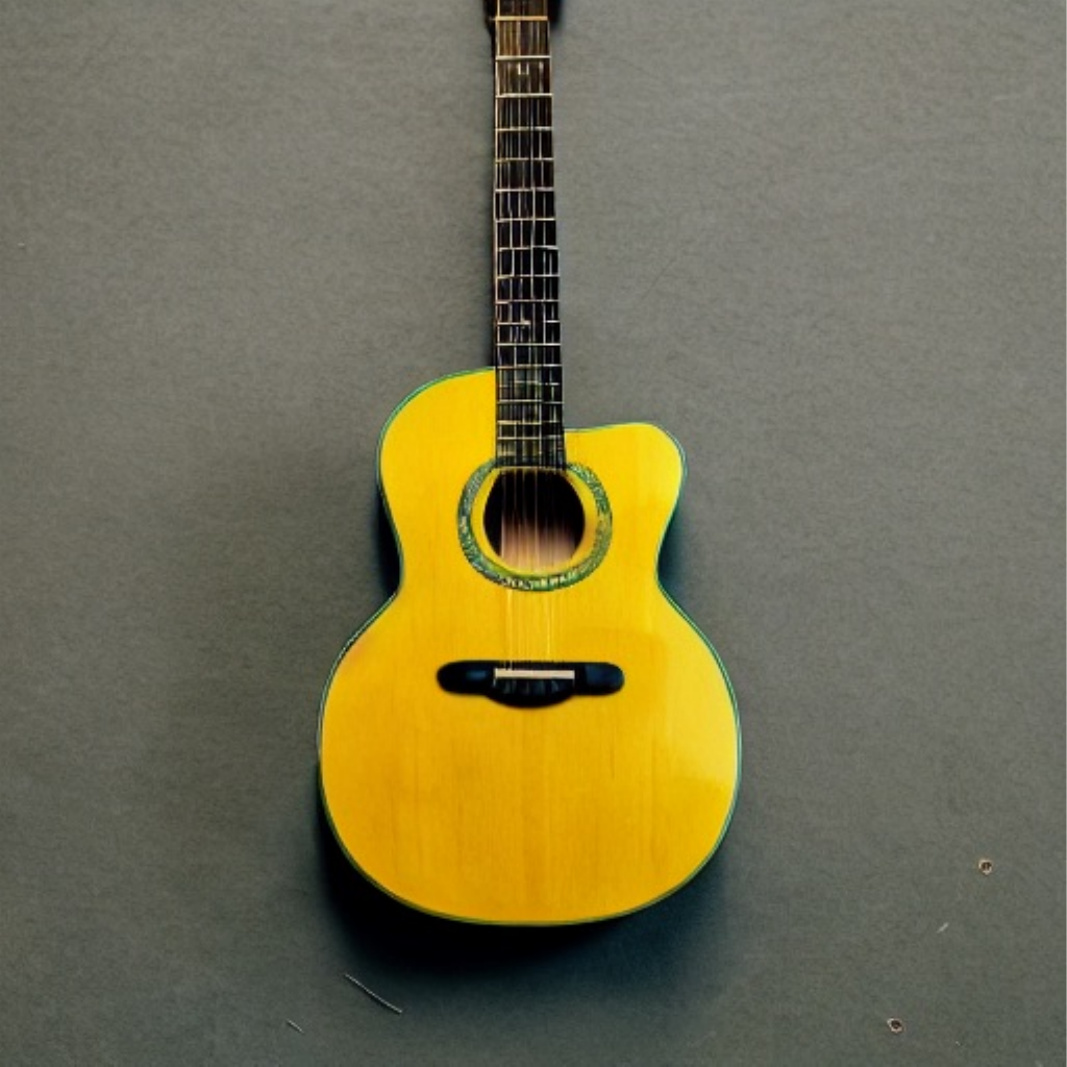} & \includegraphics[width = 0.11\textwidth]{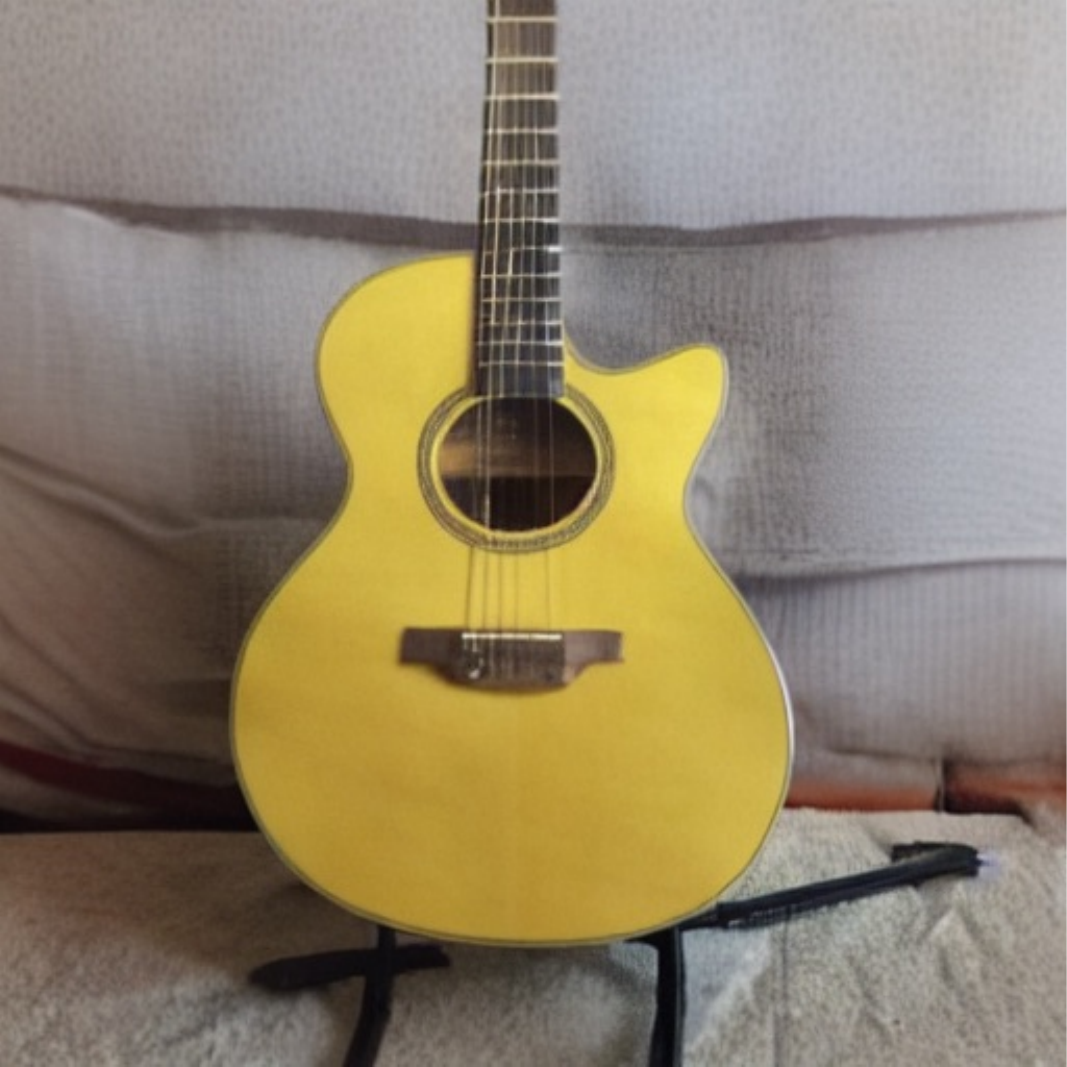} & \includegraphics[width = 0.11\textwidth]{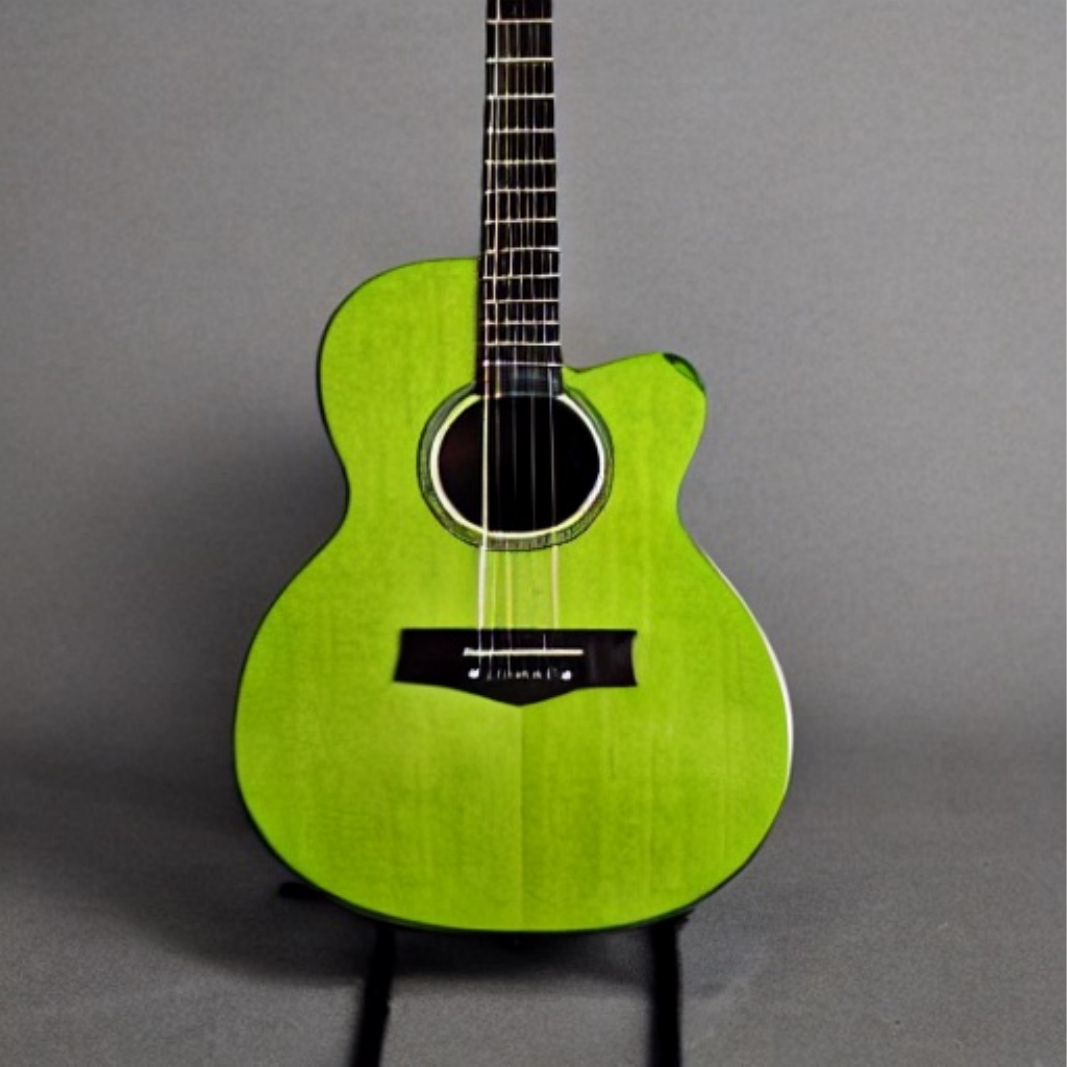} \\
         \multicolumn{4}{p{0.48\textwidth}}{ \small (a) a \textcolor{Yellow}{yellow}($\rightarrow$ \textcolor{Green}{green}) guitar} \\
          \includegraphics[width = 0.11\textwidth]{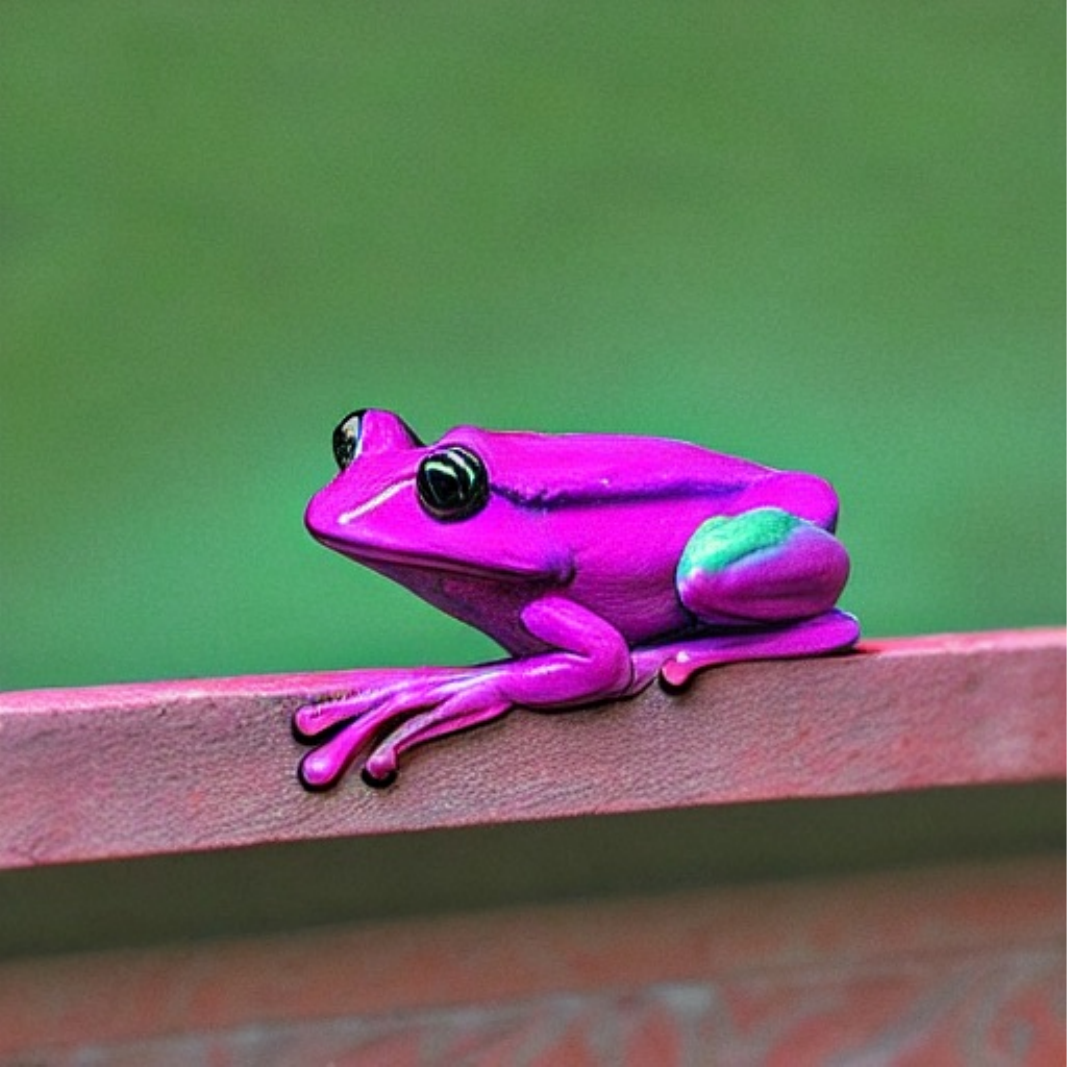} & \includegraphics[width = 0.11\textwidth]{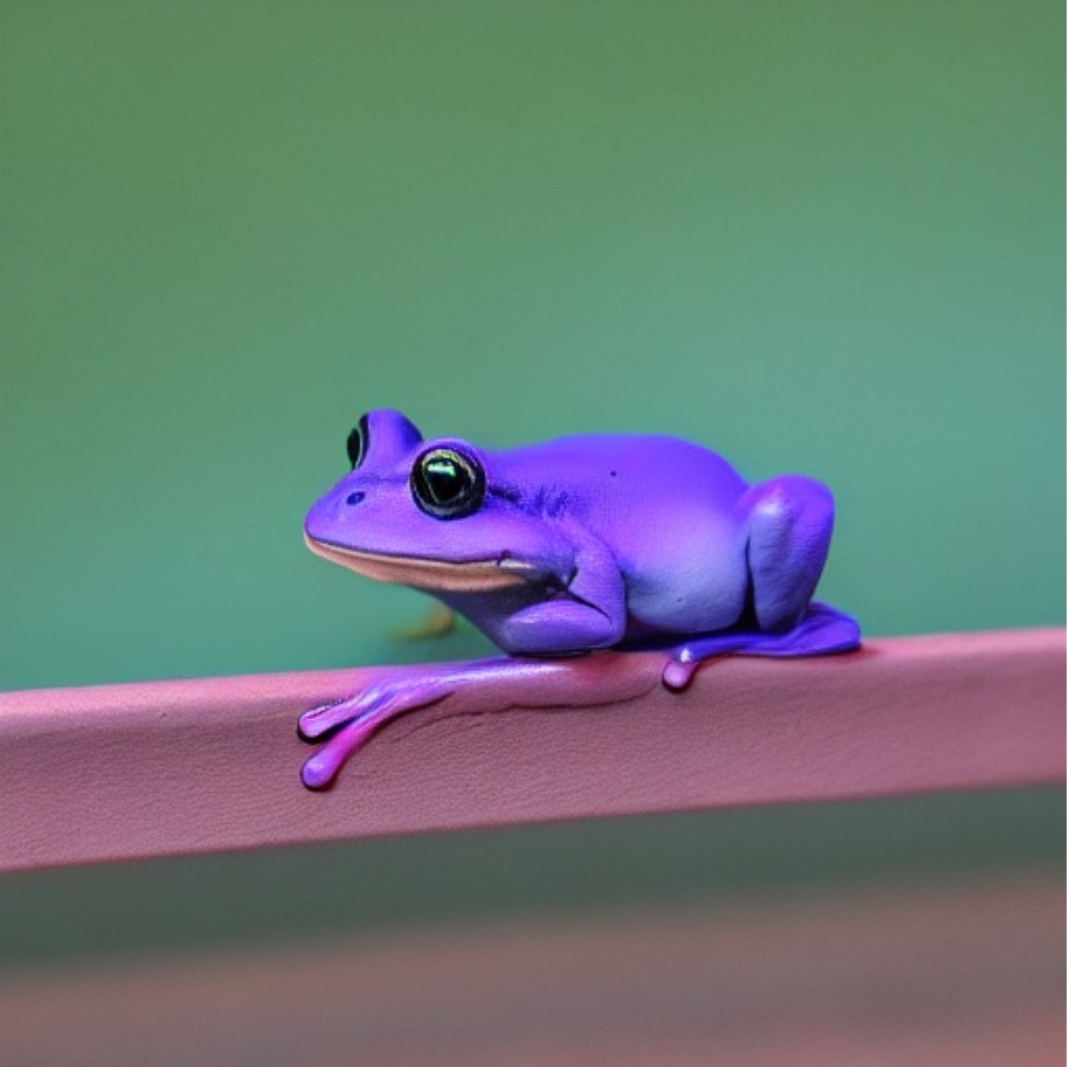} & \includegraphics[width = 0.11\textwidth]{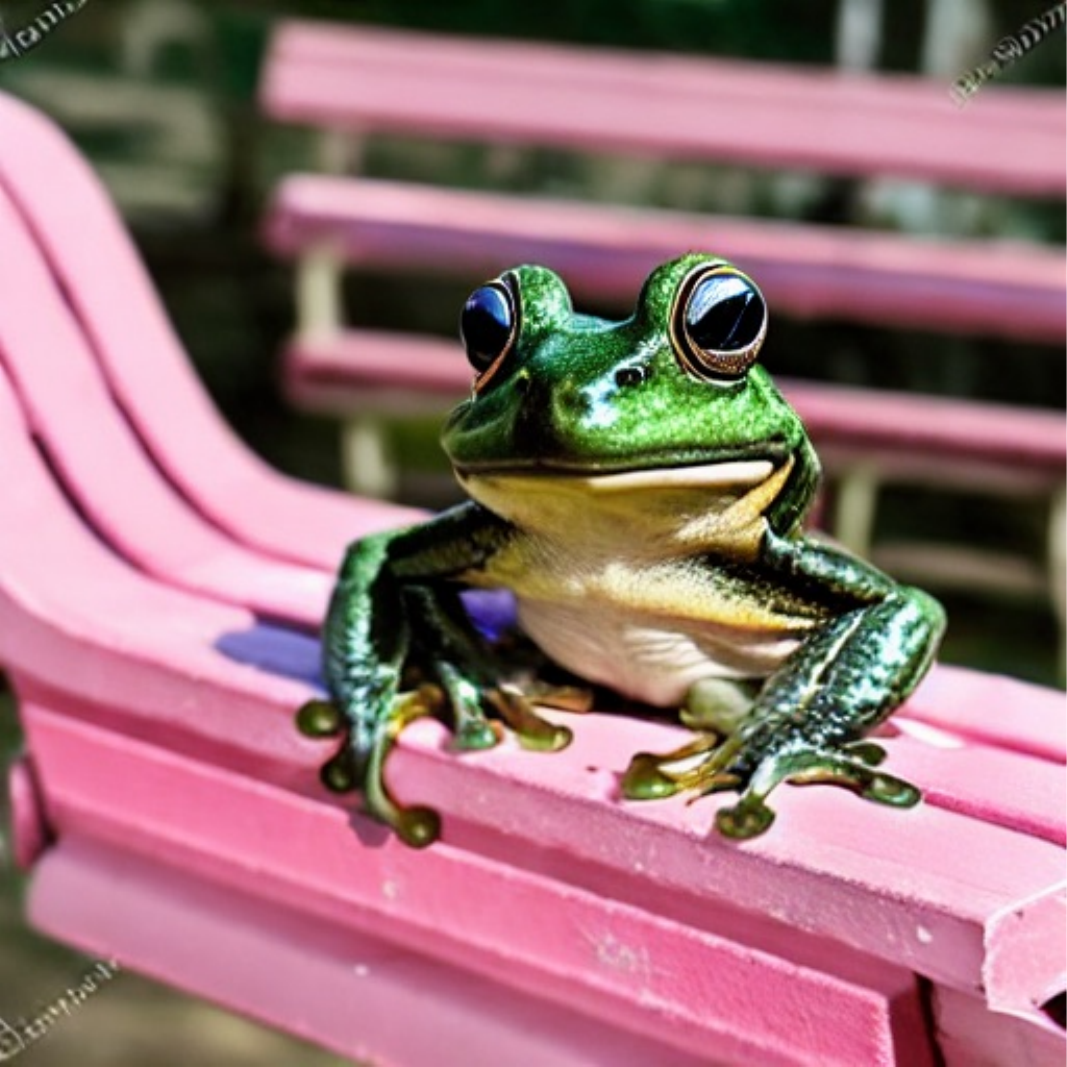} & \includegraphics[width = 0.11\textwidth]{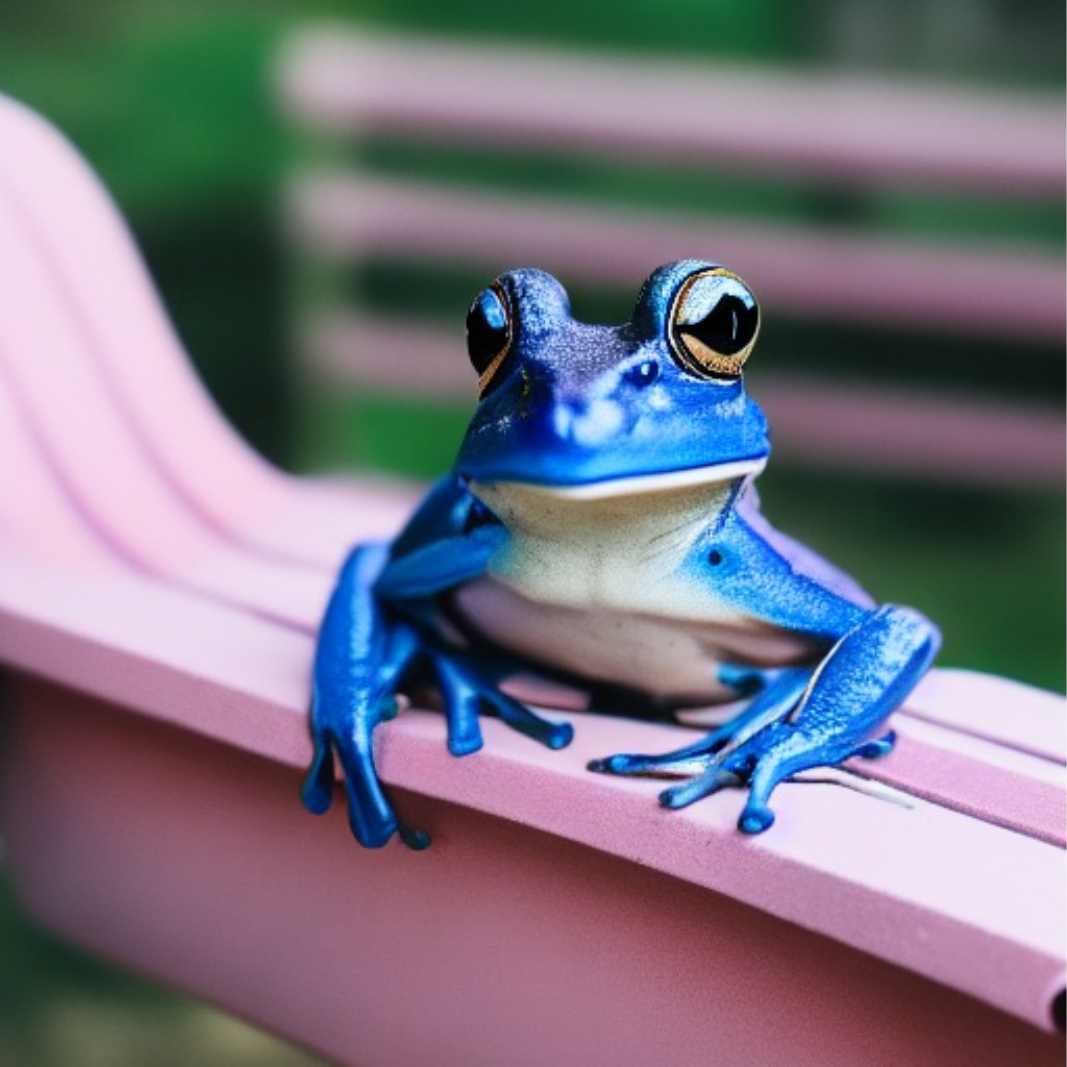} \\
          \multicolumn{4}{p{0.48\textwidth}}{  \small (b) a (+ \textcolor{blue}{blue}) frog on a pink bench} \\
           \includegraphics[width = 0.11\textwidth]{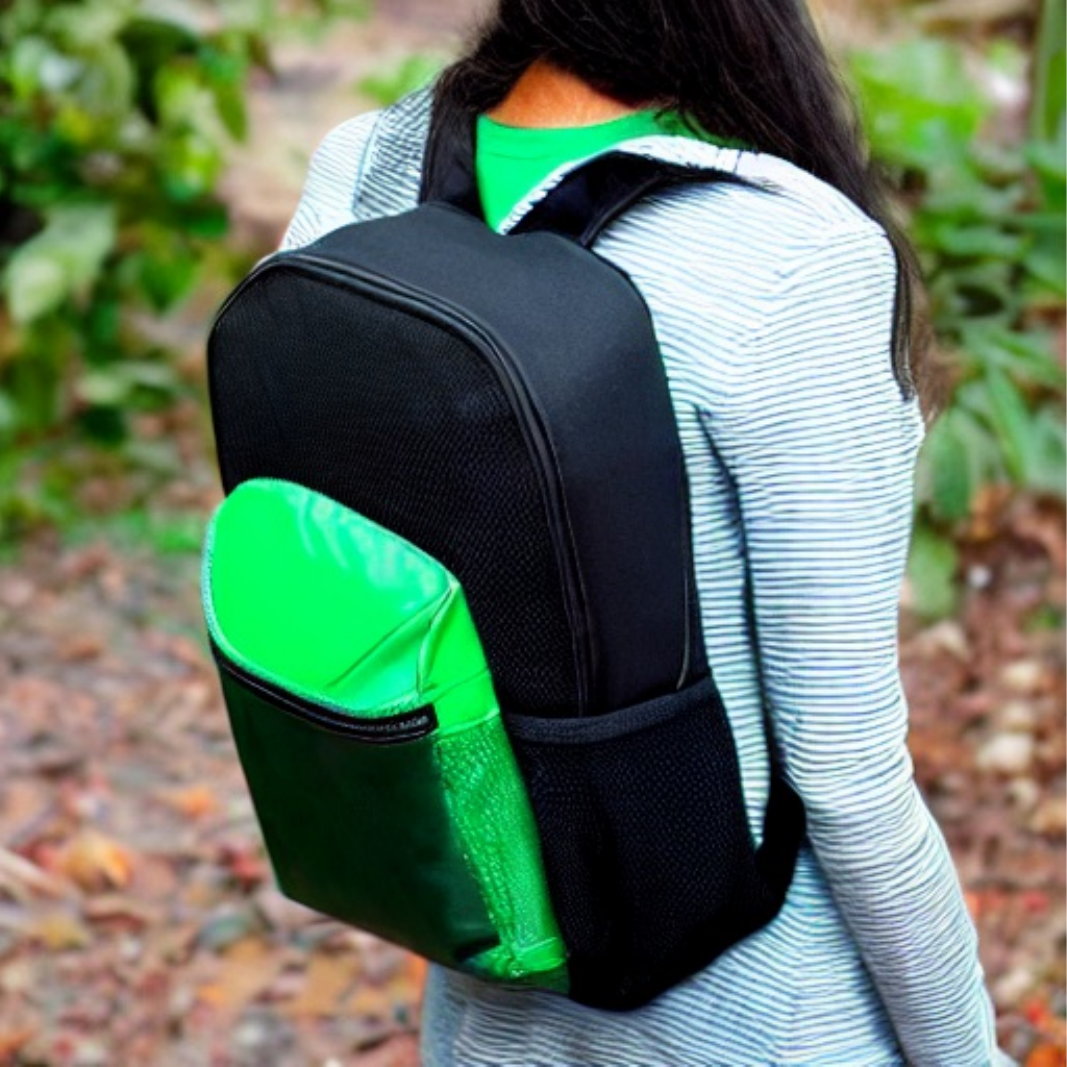} & \includegraphics[width = 0.11\textwidth]{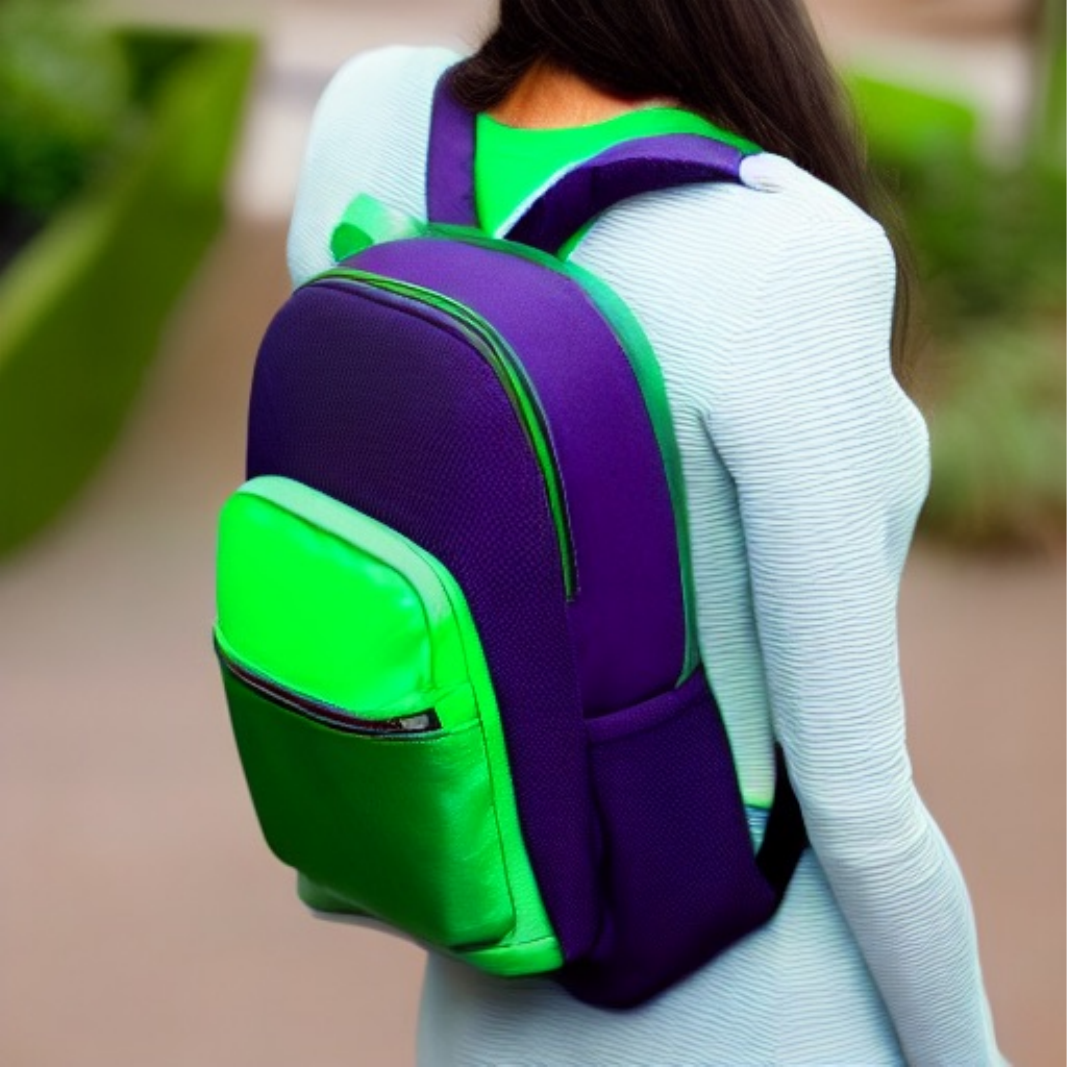} & \includegraphics[width = 0.11\textwidth]{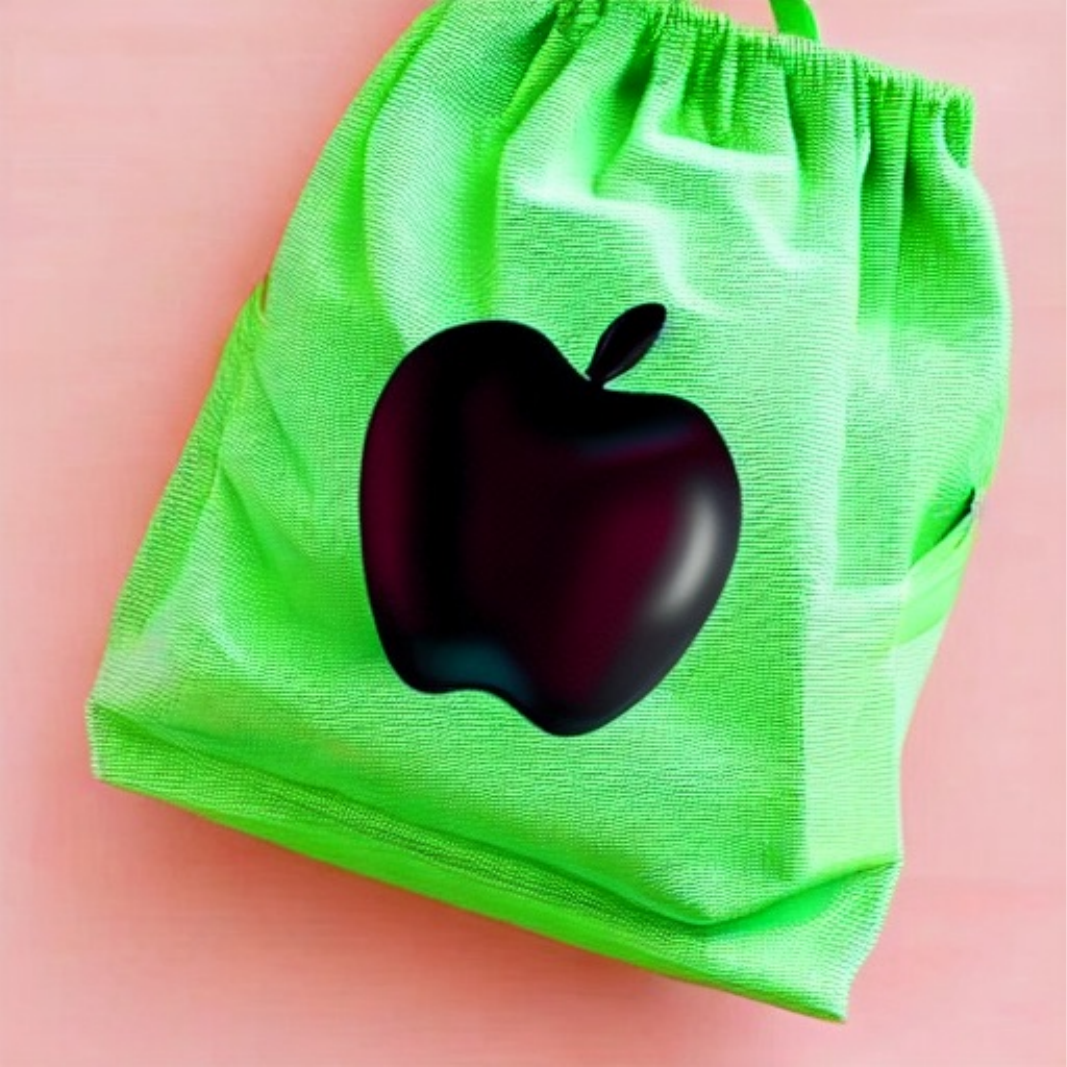} & \includegraphics[width = 0.11\textwidth]{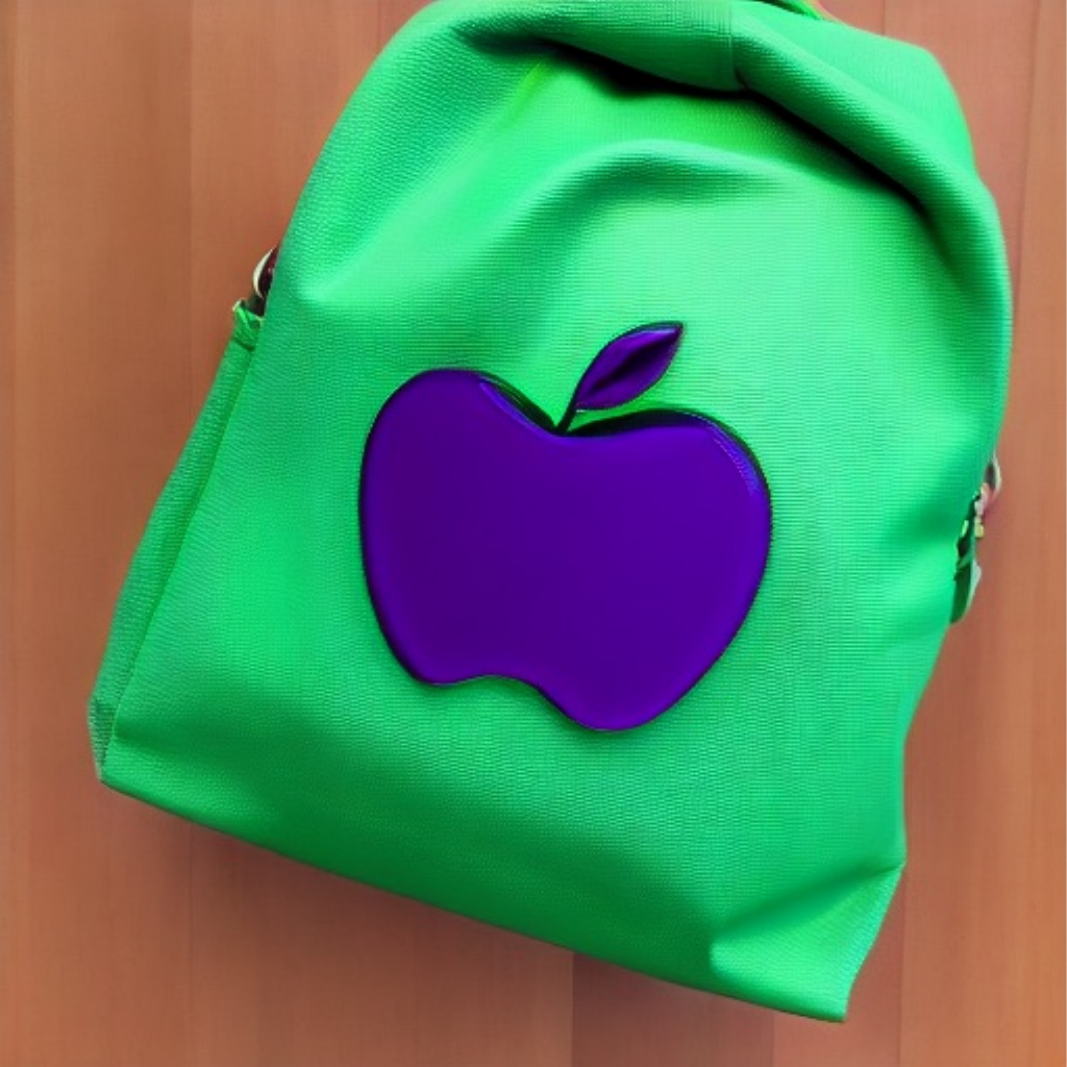} \\
             \multicolumn{4}{p{0.48\textwidth}}{  \small (c) a \textcolor{Green}{green} backpack and a  \textbf{black}($\rightarrow$ \textcolor{Orchid}{purple}) apple }  \\
            \includegraphics[width = 0.11\textwidth]{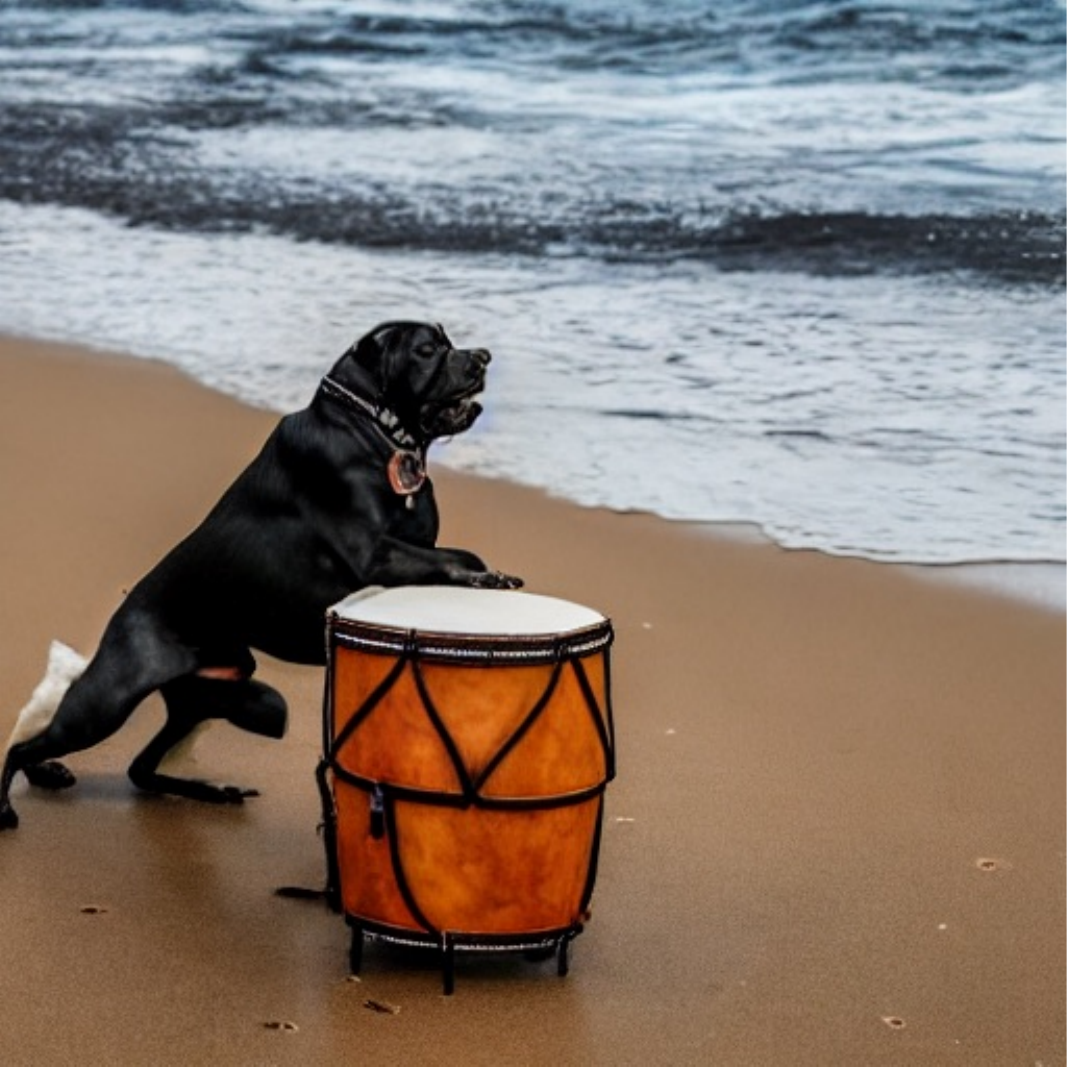} & \includegraphics[width = 0.11\textwidth]{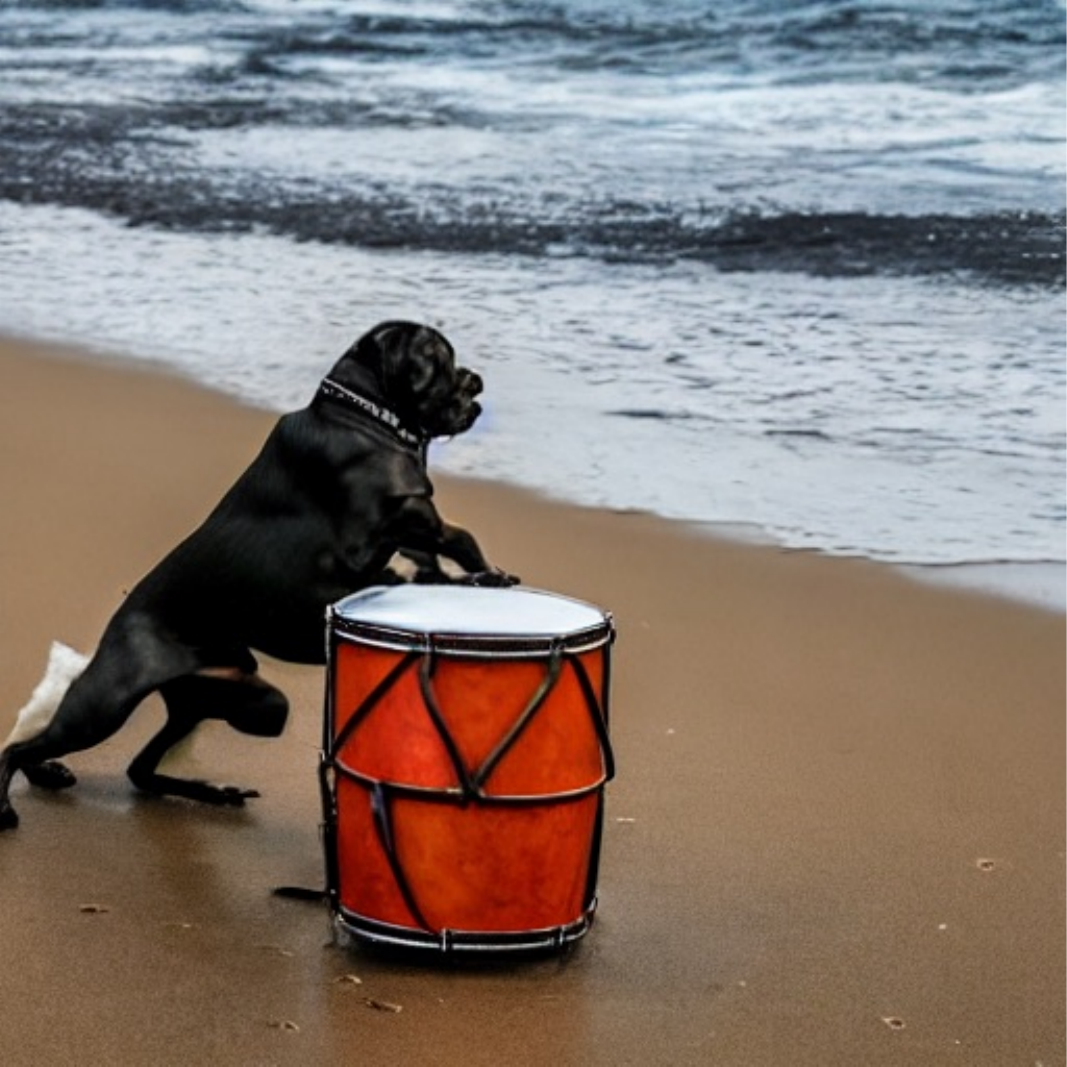} & \includegraphics[width = 0.11\textwidth]{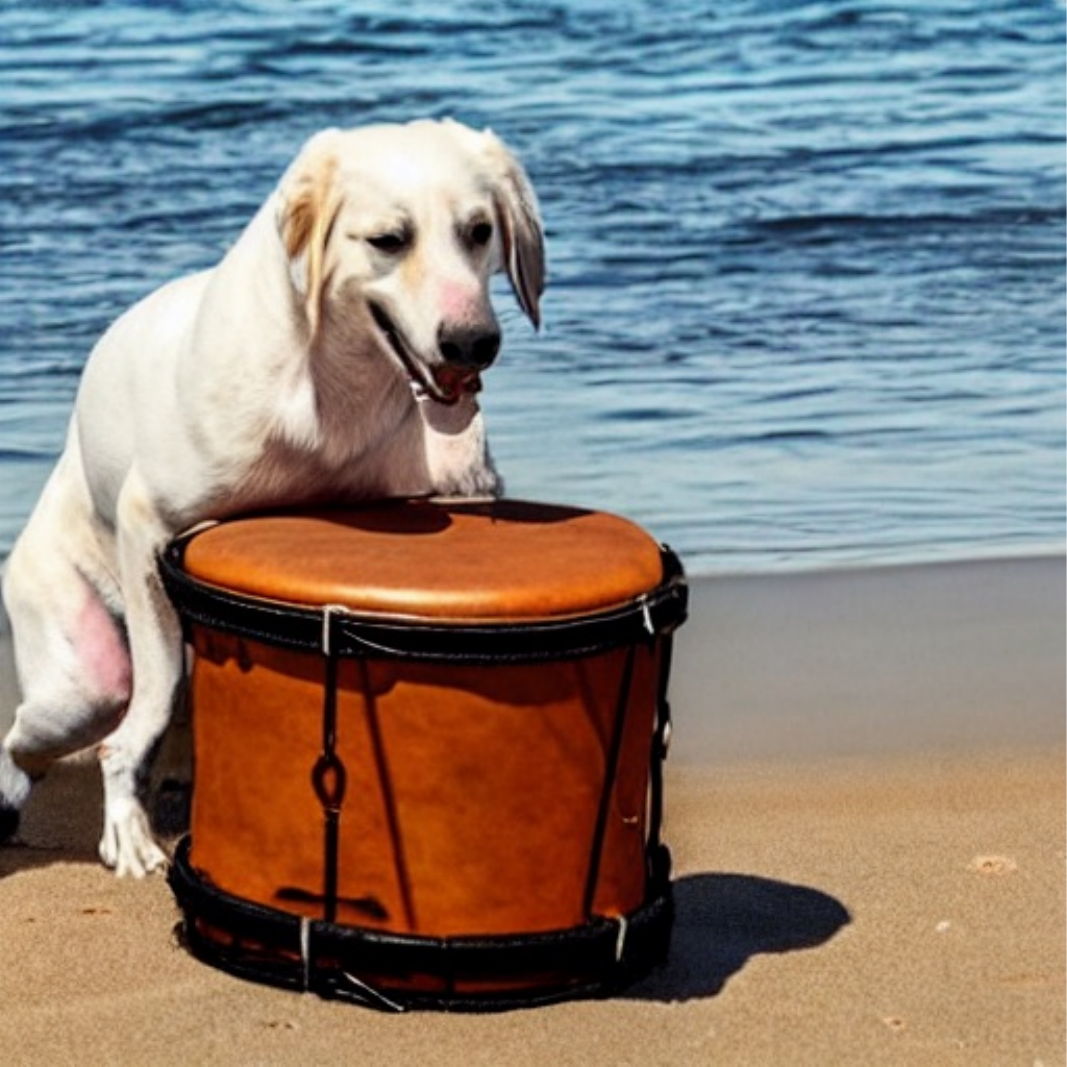} & \includegraphics[width = 0.11\textwidth]{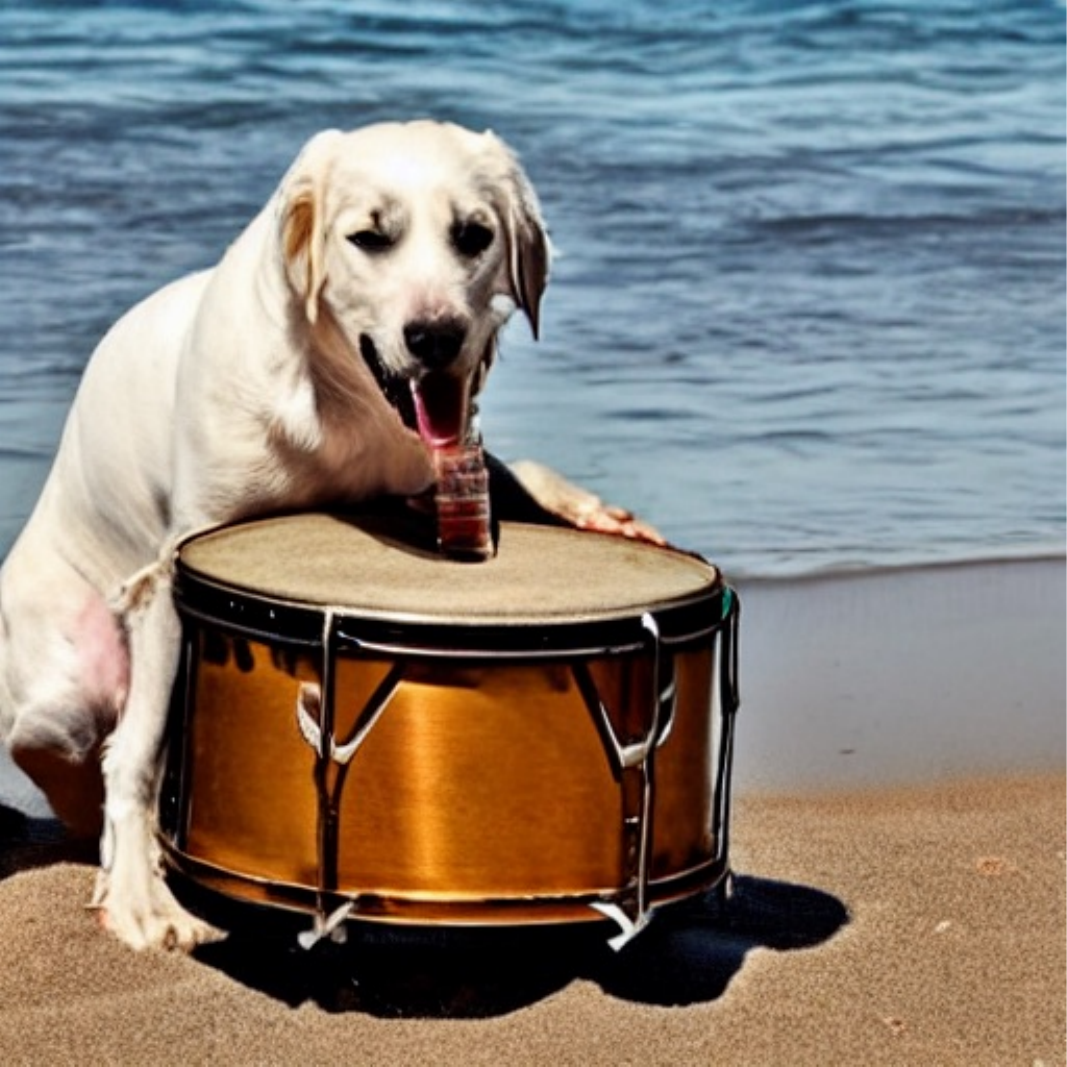} \\
              \multicolumn{4}{p{0.48\textwidth}}{  \small (d) a dog is playing a leather($\rightarrow$ metal) drum   on the beach }
    \end{tabular}
    \caption{\textbf{Augmented attribute editing with our method.} The left two columns demonstrates the attribute editing results from PtP, while the right two columns demonstrates the results from PtP w. ours.  
    }
    \label{fig:editing}
\end{wrapfigure}

PtP \cite{hertz2023prompttoprompt} allows local editing with word swap, adding new phrase, or re-weighting by replacing the attention weights of unchanged tokens, or re-weighting the attention maps of target tokens. 
This heavily relies on the semantic coupling between tokens and their attention maps. 
In Fig. \ref{fig:editing},  we categorize the failure cases in PtP shown in the left panel into four situations: (a) ineffective editing with aligned text-to-image generation; (b) ineffective editing with incorrect attribute binding, e.g. the semantic leakage of `pink'; (c) ineffective editing with object neglect, e.g. the `apple'; and (d) insignificant editing with aligned text-to-image generation, e.g. the property `metal' for the drum. 
In contrast, our method effectively enhances the semantic distribution of attention maps, allowing PtP with our approach to   apply effective and significant local attribute editing to the original images. 



\section{Conclusion}

We introduce an object-conditioned EBAMA framework to address the alignment issues in text-to-image diffusion models. We propose an object-centric attribute binding loss that maximizes the log-likelihood of the object-conditioned EBM in the attention feature space. An intensity regularizer is further designed to provide an extra degree of freedom balancing the trade-off between correct attribute binding and the necessary presence of objects. Extensive quantitative and qualitative comparisions demonstrate the superiority of our method in aligned text-to-image generation. This advancement promises great improvements in text-controlled attention-based image editing with semantically aligned attention maps.

\clearpage  

\section*{Acknowledgements}

The work was partially supported by NSF DMS-2015577 and a gift fund from Amazon. We truly thank the three anonymous reviewers for their valuable comments.

%
%
\bibliographystyle{splncs04}
\bibliography{main}

\clearpage
\setcounter{page}{1}

\appendix
\renewcommand{\thesection}{\Alph{section}}

 \section*{Supplementary Material}

 \section{Limitations}
We list the limitations of our model as follows: (1) The effectiveness of our method is limited by the expressive power of the standard Stable Diffusion model. (2) Our energy-based model  is conditioned on the object token $s \in S$. Typically, $|S| \neq 0$. However, when $|S| = 0$, it indicates that there are no objects in the prompt, and our method is degraded to regular diffusion model generation.

\section{Societal Impact and Ethical Concerns}

This paper introduces a novel method for aligning attention maps to improve the compositional generation of images from text prompts, using diffusion models. Our approach aims to advance AI-assisted visual content creation in the creative, media, and communication sectors. Although we foresee no significant ethical issues unique to our method, it inherits general concerns associated with generative models, such as privacy, copyright infringement, misuse for deceptive content, and content bias.

\section{Object-Conditioned Energy-Based Model}

In this section, we present the proof for Equ. (\ref{equ:ebm_grad}). Specifically, we elaborate on the derivation of the gradient of the log-likelihood for the EBM as defined in Equ. (\ref{equ:ebm}):
\begin{align*}
   &  \nabla_z \log p_{z}(l | s)
\\ = &\nabla_z \log (\exp ( f(A_l, A_s)) ) - \nabla_z \log ( \sum_l \exp(f (A_l, A_s) ))
\\ = &\nabla_z f(A_l, A_s) - \sum_l \frac{ \exp(f (A_l, A_s) )}{\sum_l \exp(f (A_l, A_s))} \nabla_z f (A_l, A_s)
\\ = & \nabla_z f(A_l, A_s) - \sum_l p_z(l|s) \nabla_z f (A_l, A_s)
\\ = &\nabla_z f(A_l, A_s) -
\mathbb{E}_{p_{z}(l | s)} \left[
\nabla_z f(A_l, A_s)
\right].
\end{align*}
  
\section{Algorithm}

Our workflow can be outlined in Algo. \ref{alg:cap}. Initially, for the first half of the denoising steps, the latent variable $z_t$
 is updated using the gradient of the loss function, i.e. Equ. (\ref{loss}). The latter half of the denoising steps follows the standard generation process of diffusion models.

\begin{algorithm}
\caption{Energy-Based Attention Map Alignment}\label{alg:cap}
   \textbf{Input:}   A text prompt $y$, a set of object tokens $S$, a set of modifier tokens $\{\mathcal{M}(s)\}_{s\in S}$, a pretrained Stable Diffusion model $SD$, total sampling steps $T$, an image decoder $\mathcal{D}$
   \\
    \textbf{Output:} An image $x$ aligned with the prompt $y$
\begin{algorithmic}[1]

\State \textbf{Initialize} $z_T \sim \mathcal{N}(0,1)$
\For{$t$ in $T:[T/2]+1$}
    \State $\_, A ,  \tilde A \gets  SD(z_t, t, y)$
    \State Compute attention loss  $L$  according to Equ. (\ref{loss})
    \State $z_t'  \gets  z_t - \nabla_{z_t} L$
    \State  $z_{t-1}, \_, \_ \gets SD(z_t', t, y)$
\EndFor
\For{$t$ in $[T/2]:1$}
 \State  $z_{t-1}, \_, \_ \gets SD(z_t , t, y)$
\EndFor
\State $x  \gets  \mathcal{D}(z_0)$
\State \Return $x$
\end{algorithmic}
\end{algorithm}

\section{Implementation Details}\label{implement}

Experiments were conducted on a Linux-based system equipped with 4 Nvidia R9000 GPUs, each of them has 48GB of memory. To ensure a fair comparison with previous methods, we utilized the official Stable Diffusion v1.4 text-to-image model with the CLIP ViT-L/14 text encoder.

\subsection{Hyperparameters}
In our approach, we utilize a default fixed guidance scale of 7.5. The update step size is selected as $\alpha=20$. We employ a DDIM sampler with a total of 50 steps. The update of the latent variable $z_t$   is confined to the first half of the denoising process, which, in this context, corresponds to the initial 25 steps. Further discussion regarding the step size and the updated timesteps is in Appendix \ref{add_abl}.

\subsection{Parser}
Following \cite{rassin2023linguistic}, we utilize the spaCy parser \cite{honnibal2017spacy}, specifically employing the transformer based \texttt{en\_core\_web\_trf} model. Initially, we identify tokens within the prompt that are tagged as either  NOUN  or PNOUN , thereby constituting our object set. Subsequently, we extract all modifiers within this set based on a predefined set of syntactic dependencies, which include \texttt{amod}, \texttt{nmod}, \texttt{compound}, \texttt{npadvmod}, and \texttt{conj}. Finally, any NOUN or PNOUN that functions as a modifier for other entity-nouns within the object set is excluded.

\subsection{Attention Map Extraction}
The aggregated attention features \(A_t\) comprises \(N\) spatial attention maps, each corresponding to a token of the input prompt \(y\). The CLIP text encoder appends a specialized \(\langle \texttt{SOT} \rangle\) token at the beginning of \(y\) to signify the start of the text. It has been observed that in Stable Diffusion, the \(\langle \texttt{SOT}\rangle\) token consistently receives the highest attention among all the tokens. Following \cite{chefer2023attend}, we exclude the attention allocated to \(\langle \texttt{SOT} \rangle\) and then apply a softmax operation to the remaining tokens to obtain attention scores $\tilde{A}_t$.

\subsection{Augmented Attribute Editing Setup} \label{sec:edit}
We utilize and slightly adapt the official repository from \cite{hertz2023prompttoprompt} for conducting attention editing. A cross-replace step of 0.8, i.e. replacing the first 80\% steps of cross-attention maps,  is employed for all editing tasks. Additionally, in line with the repository's provisions, self-attention maps also play a role in  preserving the image's shape. For this purpose, we set the self-replace step at 0.4, i.e. replacing the first 40\% steps of self-attention maps, for all the experiments.

\subsection{A-Star Comparison}  
As A-STAR has not released their official code of implementation, we display their reported numeric results of T-C Sim. in Tab. \ref{tab:a_star}. The table shows that our method demonstrates significantly superior performance when it comes to more complicated datasets. Note that A-A, A-O, and O-O include 0, 1, and 2 attributes in their prompts, respectively.

\begin{table}[!h]
\centering
\caption{\footnotesize Comparison with A-Star. * Values copied from the A-Star paper. }
\begin{tabular}{@{}lccc@{}}
\toprule
\footnotesize Method & \footnotesize A-A &\footnotesize  A-O & \footnotesize O-O \\ \midrule
\footnotesize AnE*  &\footnotesize 0.80 \textcolor{Red}{(-2.4\%)}   &\footnotesize 0.82 \textcolor{Red}{(-3.5\%)}  &\footnotesize 0.81 \textcolor{Red}{(-3.6\%)}   \\
\footnotesize SG & \footnotesize0.77 \textcolor{Red}{(-6.1\%)} &\footnotesize 0.83 \textcolor{Red}{(-2.4\%)} & \footnotesize0.81 \textcolor{Red}{(-3.6\%)} \\
\footnotesize A-STAR* &\footnotesize  {0.82} \textcolor{Red}{(-0.0\%)} & \footnotesize0.84 \textcolor{Red}{(-1.2\%)} & \footnotesize0.82 \textcolor{Red}{(-2.4\%)} \\ 
\midrule
\footnotesize Ours & \footnotesize \textbf{0.82} & \footnotesize \textbf{0.85} \footnotesize &\footnotesize \textbf{ 0.84} \\
\bottomrule
\end{tabular} \label{tab:a_star}
\end{table}

\section{Computational Efficiency}
We randomly sampled a total of 100 prompts from the ABC-6K dataset and compared the time required for each method. Since there are no updates in SD, it provides us with the lower bound of the time needed for generating the images. As shown in Tab. \ref{time}, we can observe that AnE takes the longest time, approximately 47.0 minutes, to complete the generation process, while our method and SG require less than half of that time. The reason for this discrepancy is that AnE employs a technique called iterative latent refinement, which may require multiple updates at certain steps when the loss objective does not meet specific thresholds. We emphasize that our method is computationally efficient while still achieving top performance.
\begin{table}[!h]
    \centering
    \caption{\textbf{Computational time comparison.} We compare the time required for generating images for 100 prompts in ABC-6K dataset for each method.}
    \begin{tabular}{lcccc} \toprule
         &  SD & AnE & SG & Ours\\ \midrule
     Time(min)&   13.2& 47.0     & 21.9    & 20.9 \\ \bottomrule
    \end{tabular}
    \label{time}
\end{table}

\section{External modifiers incorporation} 
 We modify our workflow shown in Fig. \ref{method} to include external modifiers without affecting the original prompt's guidance: (a) We first append the modifier(s) to the prompt to obtain attention maps for both original and external tokens. (b) We calculate the final loss using these attention maps. (c) After updating   $z_t$, we drop the external modifier(s) and proceed with the original prompt. Tab. \ref{tab:external} presents the results of incorporating arbitrary syntactically unrelated modifiers, with their number denoted as  $n$. The modifiers chosen are `red'($n=1$), and `red blue green'($n=3$). The performance decreases with $n$ increasing. This decline may be due to external tokens interfering with the repulsion of other non-modifier tokens, as the attribute binding loss has to balance the repulsion of external tokens and that of non-modifier tokens.  
\begin{table}[h]
    \centering
    \begin{tabular}{c|ccc}
     $n$ &   A-A &   A-O &   O-O 
    \\ \hline
 \rowcolor{gray}      0&    \textbf{0.340/0.256/0.817} &  \textbf{0.362/0.270/0.851} & \textbf{0.366/0.274/0.836}\\
        1  &     {0.337}/{0.254}/0.812   &    0.360/0.269/0.844    &    0.364/0.272/0.826\\
   3 &    0.331/0.248/0.822 &    0.357/0.266/0.846 &    0.362/0.270/0.823 
    \end{tabular}  \caption{\textbf{Results on external modifiers incorporation under the same setting as Tab. \ref{ane_table} }. $n$ denotes the number of external modifiers appended.} \label{tab:external}
\end{table}

\section{Additional Ablation Experiments} \label{add_abl}

\begin{figure}[t]
    \centering
    \includegraphics[width=0.48\textwidth]{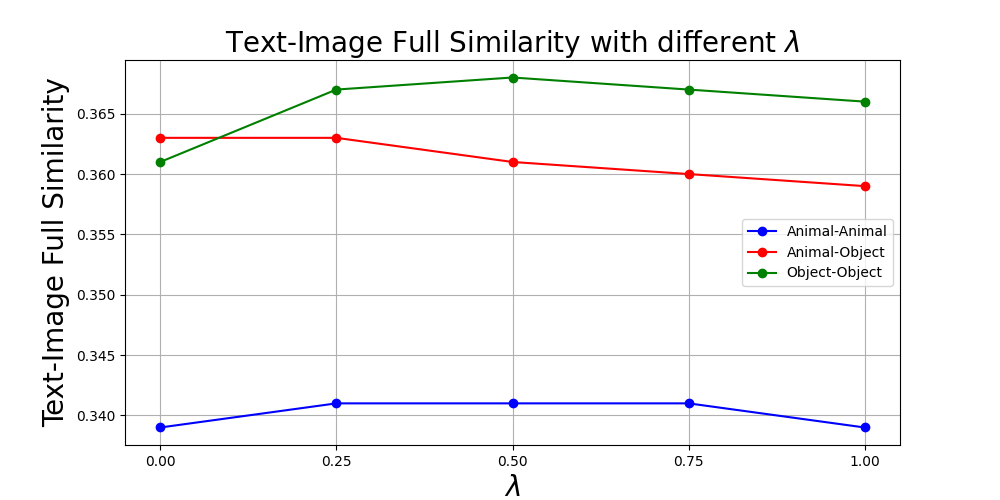} 
     \includegraphics[width=0.48\textwidth]{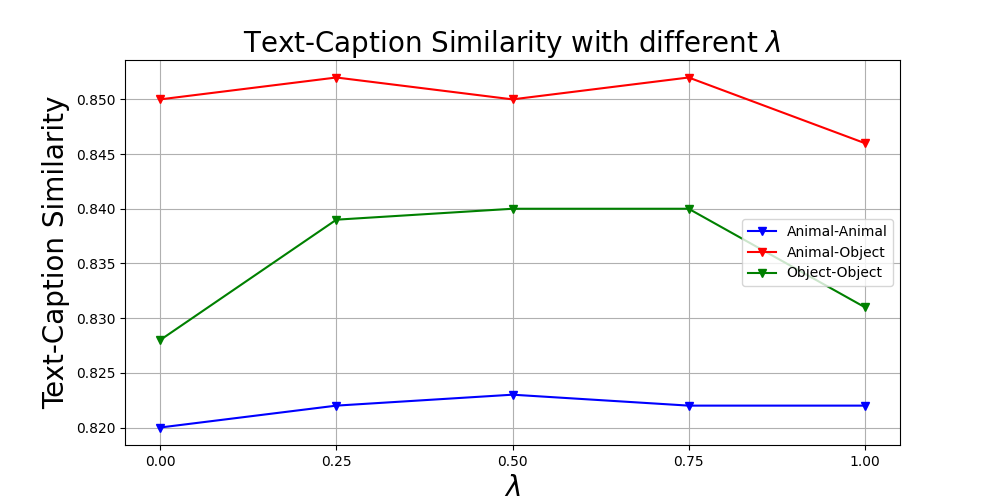}
    \caption{\textbf{Ablation study for $\lambda$.} We generated 10 images for each prompt with the same seed across all methods. The results indicate that for datasets with Animal-Animal and Object-Object pairings, a setting of \(\lambda = 0.5\) is optimal; whereas for the Animal-Object dataset, \(\lambda = 0.25\) yields the best performance.}
    \label{ablation}
\end{figure}

 \paragraph{Intensity Weight $\lambda$} We explore various settings of the intensity weight parameter \(\lambda\) as illustrated in Fig. \ref{ablation}, where the metrics are computed across 10 images for each prompt. The values of Text-Image Full Similarity (Full. Sim.) and Text-Caption Similarity (T-C Sim.) are presented as functions of varying \(\lambda\). At \(\lambda = 0\), the intensity level is disregarded by the method. Conversely, increasing \(\lambda\) shifts the focus more towards the intensity level, at the expense of  distribution alignment in attention maps.
 
  For Animal-Animal and Object-Object , both metrics peak at \(\lambda=0.5\). For the Animal-Object dataset, the Text-Image similarity attains its highest score at \(\lambda = 0\) or \(\lambda=0.25\). Given that Text-Caption Similarity is maximal at \(\lambda=0.25\), this value is selected for the Animal-Object dataset.

Our analysis indicates that \(\lambda\) effectively balances the trade-off between intensity level and attribute binding. Extremes of \(\lambda\) (e.g., \(1.0\) or \(0.0\)) yield suboptimal generation results. Based on these findings, and considering that the ABC-6K dataset is more akin to the Object-Object dataset -- notably, they both contain at least two attributes and two objects -- we select \(\lambda=0.5\) for the ABC-6K dataset in all our experiments. The situation with the DVMP dataset is more intricate due to its potential for containing one to three objects and an unlimited number of attributes. As we suggest in the main text, selecting different intensity weights based on the number of objects in a prompt is recommended to optimize our method's performance. 

 \begin{figure}[!h]
     \centering
     \renewcommand{\arraystretch}{0.7} 
\setlength{\tabcolsep}{0.5pt} 
    \begin{tabular}{cccc}
       \raisebox{0.05\textwidth}{ \hspace{-5pt} \rotatebox{90}{$T'=0$} }& \includegraphics[width=0.15\textwidth]{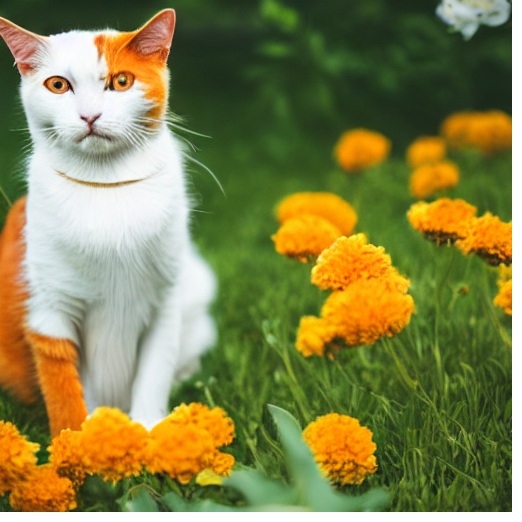} & \includegraphics[width=0.15\textwidth]{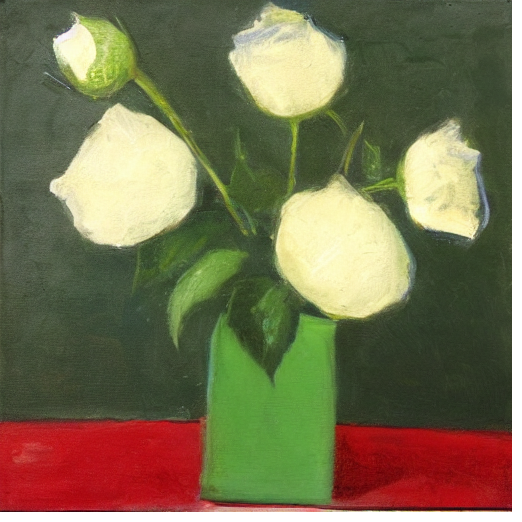} & \includegraphics[width=0.15\textwidth]{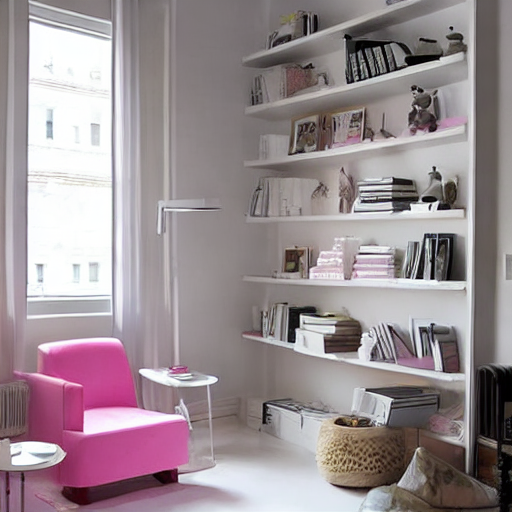}   \\
    \raisebox{0.05\textwidth}{ \hspace{-5pt}\rotatebox{90}{$T'=25$}} & \includegraphics[width=0.15\textwidth]{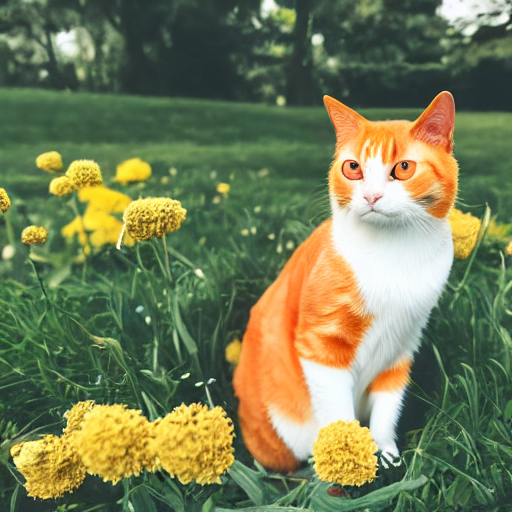} & \includegraphics[width=0.15\textwidth]{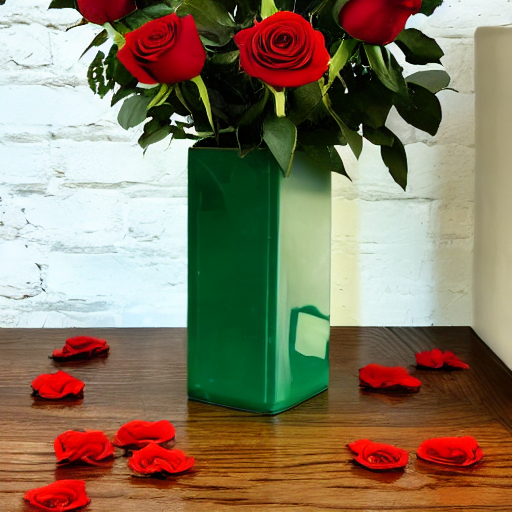} & \includegraphics[width=0.15\textwidth]{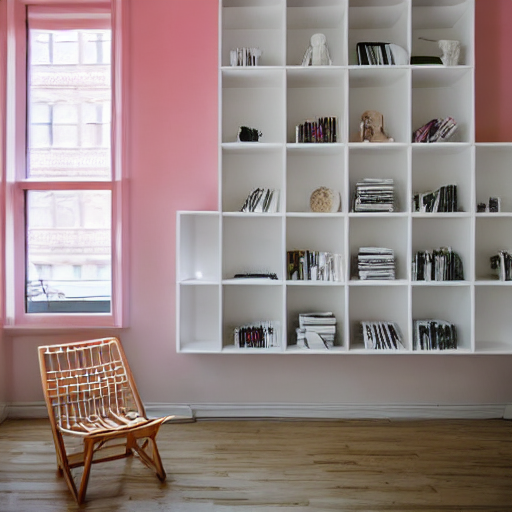}  \\
    \raisebox{0.05\textwidth}{ \hspace{-5pt}\rotatebox{90}{$T'=50$}} & \includegraphics[width=0.15\textwidth]{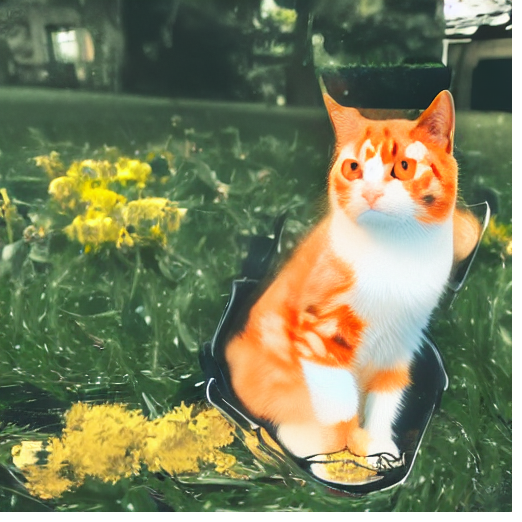} & \includegraphics[width=0.15\textwidth]{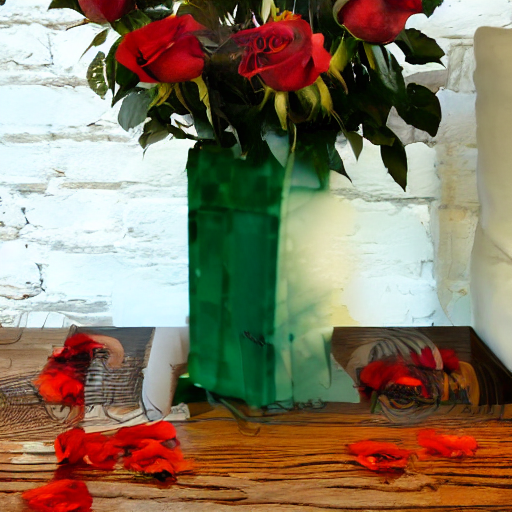} & \includegraphics[width=0.15\textwidth]{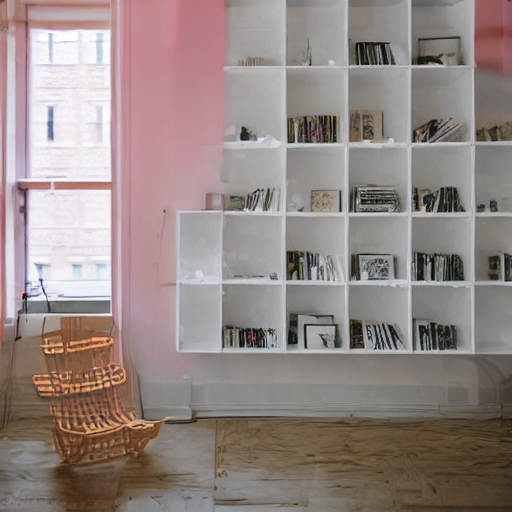}  \\
   & (a) & (b) & (c)
    \end{tabular}
     \caption{\textbf{Ablation demonstration for updated timesteps $T'$}. (a) an \textcolor{Orange}{orange} and \textcolor{tgray}{white} \underline{cat} sitting in the \underline{grass} near some \textcolor{Yellow}{yellow} \underline{flowers}; (b) \textcolor{red}{red} \underline{roses} in a square \textcolor{Green}{green} \underline{vase}; (c) A \underline{room} with \textcolor{pink}{pink} \underline{walls} and \textcolor{tgray}{white} display \underline{shelves} and \underline{chair}.   Each column shares the same random seed.}
     \label{Ablation_timesteps}
 \end{figure}

\paragraph{Number of updated timesteps $T'$}  
We explore different settings for the updated timesteps, denoted as $T'$, which refer to the timestep numbers of updating the latent variable $z_t$. This exploration is depicted in Fig.~\ref{Ablation_timesteps}. When $T'=0$, our method defaults to the standard stable diffusion generation, with no updates applied to the model. In this configuration, due to the lack of interventions during the generation process, the generated images often exhibit semantic misalignments. Examples include the yellow flowers in (a), the red roses in (b), and the pink walls in  (c). Conversely, setting $T'=25$ implements our proposed method, which produces images better aligned with the input text. However, increasing $T'$ to 50, where $z_t$ is updated throughout the generation process, can introduce artifacts. Notable instances of these artifacts are visible in the representations of the cat in  (a), the vase in   (b), and the chair in   (c).

 \begin{figure}[!h]
     \centering
     \renewcommand{\arraystretch}{0.7} 
\setlength{\tabcolsep}{0.5pt} 
    \begin{tabular}{cccc}
       \raisebox{0.05\textwidth}{ \hspace{-5pt} \rotatebox{90}{$\alpha=1$} }& \includegraphics[width=0.15\textwidth]{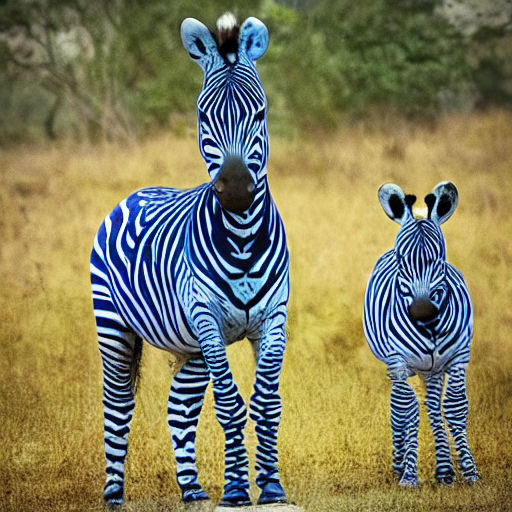} & \includegraphics[width=0.15\textwidth]{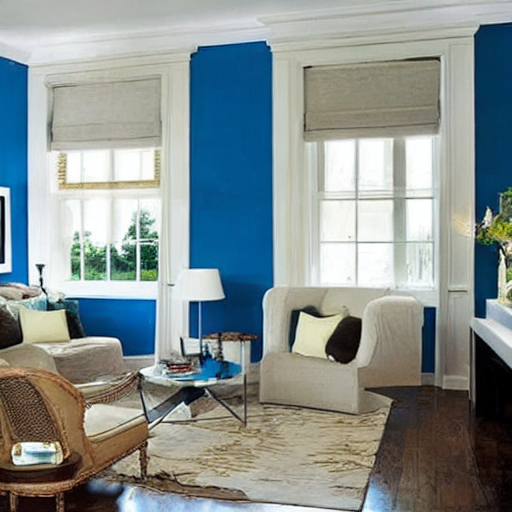} & \includegraphics[width=0.15\textwidth]{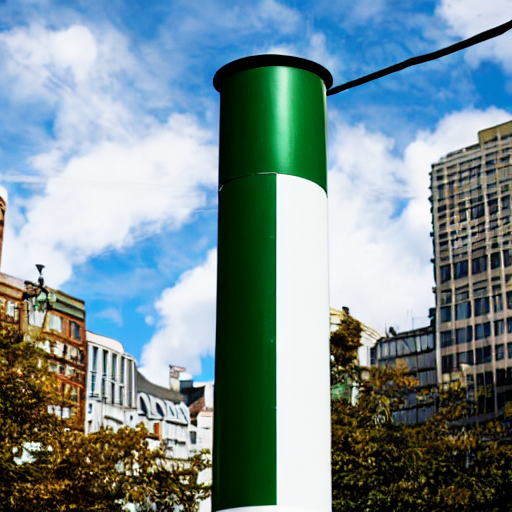}   \\
    \raisebox{0.05\textwidth}{ \hspace{-5pt}\rotatebox{90}{$\alpha=20$}} & \includegraphics[width=0.15\textwidth]{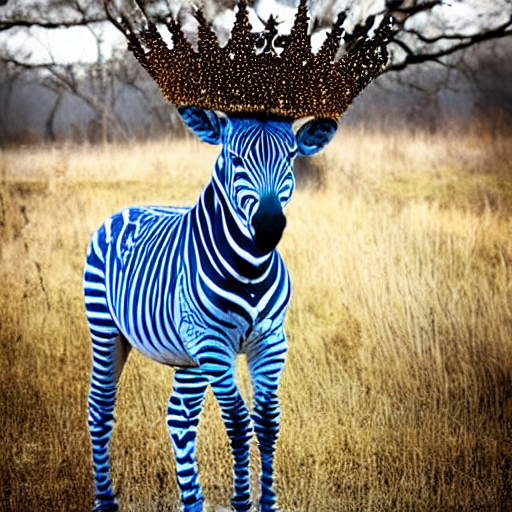} & \includegraphics[width=0.15\textwidth]{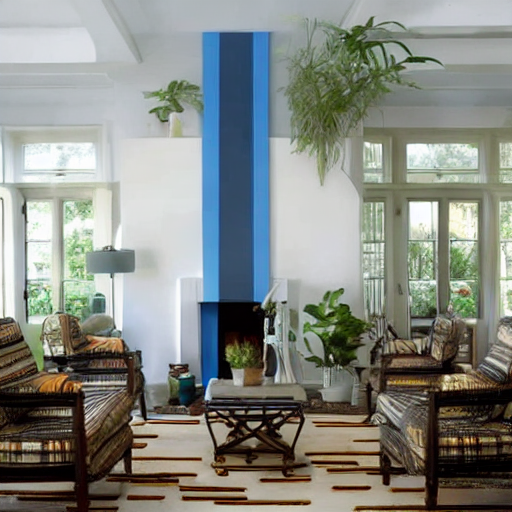} & \includegraphics[width=0.15\textwidth]{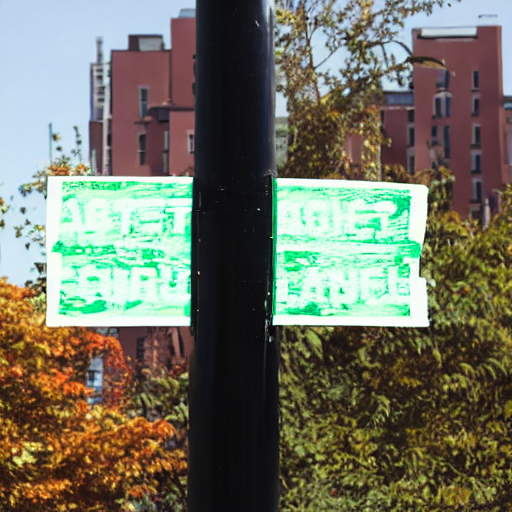}  \\
    \raisebox{0.05\textwidth}{ \hspace{-5pt}\rotatebox{90}{$\alpha=40$}} & \includegraphics[width=0.15\textwidth]{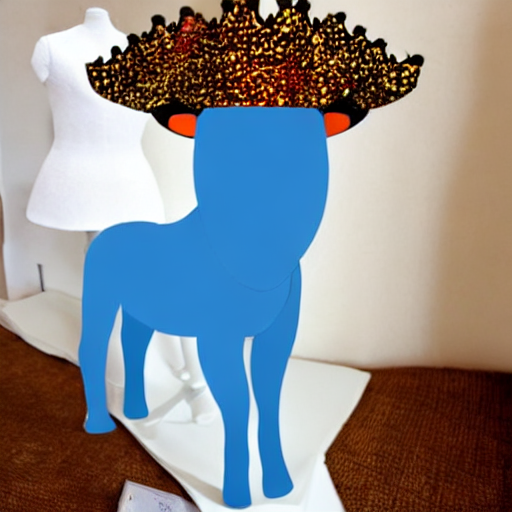} & \includegraphics[width=0.15\textwidth]{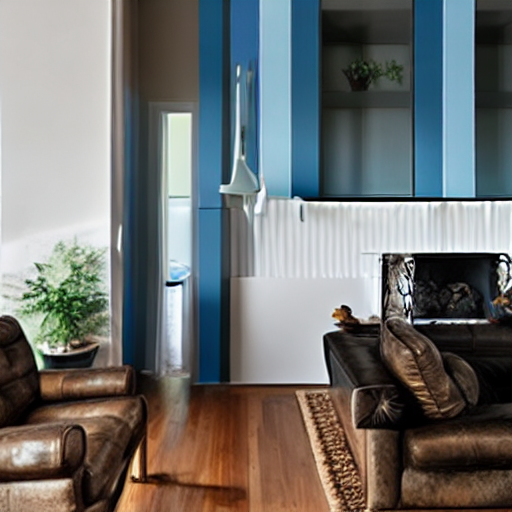} & \includegraphics[width=0.15\textwidth]{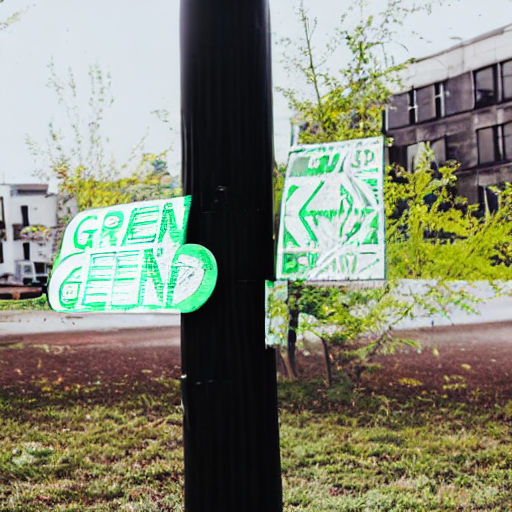}  \\
   & (a) & (b) & (c)
    \end{tabular}
     \caption{\textbf{Ablation demonstration for step size $\alpha$}. (a) a \textcolor{blue}{blue} \underline{zebra} and a spotted \underline{crown}; (b) a living \underline{room} with \textcolor{tgray}{white} \underline{walls} and \textcolor{blue}{blue} \underline{trim}; (c) a \textcolor{Green}{green} and \textcolor{tgray}{white} \underline{sign} on a \textbf{black} \underline{pole} and some \underline{buildings}.   Each column shares the same random seed.}
 
     \label{Ablation_stepsize}
 \end{figure}

\paragraph{Step Size $\alpha$} We investigate different settings for the step size  $\alpha$, as depicted in Fig.~\ref{Ablation_stepsize}. When $\alpha$ is set to 1, the step size is too small, leading to insufficient attribute binding and the inability to generate multiple objects effectively. This is evident from the examples of blue walls in   (b), a green pole in   (c), and the missing crown in   (a).  Conversely, with $\alpha$ set to 40, the step size becomes excessively large, causing an overemphasis on certain attributes, e.g. blue in (a), blue in (b), black in (c) (note that the building behind is also black).



\section{Additional Details on Human Evaluation} \label{add_user}

Raters were enlisted via an online platform under conditions of anonymity, with the requirement that each participant possessed an educational level of a bachelor's degree or higher. Additionally, they were assured of the protection of their privacy and the confidentiality of their identities throughout the process.

\begin{figure}[!h]
    \centering
    \includegraphics[width=1\textwidth]{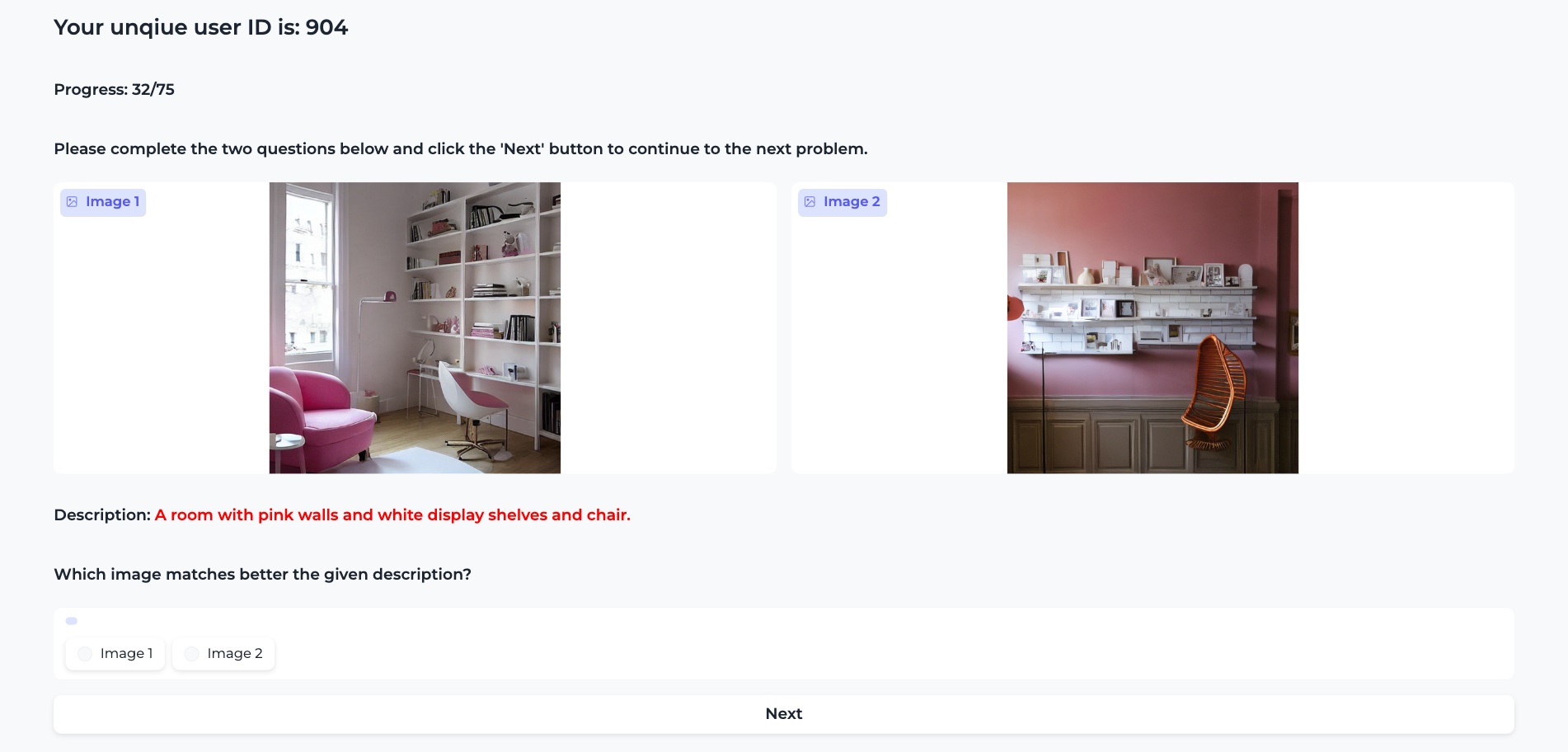}
    \caption{\textbf{A screenshot of the rating interface. }The order of images was randomized. }
    \label{fig:user-study}
\end{figure}

Fig. \ref{fig:user-study} provides a screenshot of the rating interface. The order of images was randomized. 

\end{document}